\documentclass[12pt]{article}
\usepackage{psfrag,epsf}
\usepackage{graphicx}
\usepackage{amssymb}
\usepackage{amsmath}
\usepackage{enumerate}
\usepackage{natbib}
\usepackage{url} 
\usepackage{program}
\usepackage{float}
\usepackage[normalem]{ulem}
\usepackage{framed}
\usepackage[hypcap=true]{caption}
\usepackage[plain]{algorithm}
\usepackage{rotating}
\usepackage{natbib}
\usepackage{multirow}
\usepackage{hyperref}
\usepackage{amsfonts}
\usepackage{mathrsfs}
\usepackage{subcaption}
\usepackage{mathtools}
\usepackage{cancel}
\usepackage{nicefrac}
\usepackage{mathtools}
\usepackage{ulem}
\usepackage{xr}

\usepackage[table,xcdraw]{xcolor}

\makeatletter

\newtheorem{theorem}{Theorem}
\newtheorem{lemma}{Lemma}

\newtheorem{assumption}{Assumption}

\usepackage{colortbl}
\usepackage{color}

\newcommand{\iid}{\stackrel{iid}{\sim}}
\newcommand{\ind}{\stackrel{ind}{\sim}}

\newcommand{\bm}[1]{\boldsymbol{#1}}

\newcommand{\E}{\mathbb{E}}
\renewcommand{\e}{\mathrm{e}}
\def\P{\mathbb{P}}
\newcommand{\mS}{\mathcal{S}}
\newcommand{\Y}{\bm{Y}}
\renewcommand{\C}{\origbar}
\renewcommand{\mid}{\origbar}
\renewcommand{\R}{\mathbb{R}}
\newcommand{\X}{\bm{X}}
\renewcommand{\Y}{\bm{Y}}
\newcommand{\wh}[1]{\smash{\widehat{#1}}}
\newcommand{\wt}[1]{\smash{\widetilde{#1}}}

\newcommand{\Prob}{\mathbb{P}}

\newcommand{\yw}[1]{{\color{black}#1}}

\def\spacingset#1{\renewcommand{\baselinestretch}%
{#1}\small\normalsize} \spacingset{1}

\begin{document}



{
  \title{\bf \sf Variable Selection via Thompson Sampling}
  \author{Yi Liu\footnote{Yi Liu is a \yw{$4^{th}$} year PhD student at the Department of Statistics at the University of Chicago}\, and Veronika Ro\v{c}kov\'{a}\footnote{Veronika Rockova is Associate Professor in Econometrics and Statistics at the Booth School of Business of the University of Chicago. The author gratefully acknowledges the support from the James S. Kemper Foundation Research Fund at the Booth School of Business. } }
  \maketitle
}

\bigskip
\begin{abstract}


Thompson sampling is a heuristic algorithm for the multi-armed bandit problem which has a long tradition in machine learning.  The algorithm has a Bayesian spirit in the sense that it selects arms based on posterior samples of  reward probabilities of each arm.  By forging a connection between combinatorial binary bandits and spike-and-slab variable selection, we propose a stochastic optimization approach to subset selection called Thompson Variable Selection (TVS). TVS is a framework for interpretable machine learning  which does not rely on the underlying model to be linear. TVS brings together Bayesian reinforcement and machine learning in order to extend the reach of Bayesian subset selection to non-parametric models and  large datasets  with very many predictors  and/or very many observations.  Depending on the choice of a reward, TVS can be deployed in  offline as well as  online setups with streaming data batches.  Tailoring multiplay bandits to variable selection, we provide regret bounds  without necessarily assuming that the arm mean rewards be unrelated. We show a very strong empirical performance on both simulated and real data. Unlike  deterministic optimization methods for spike-and-slab variable selection, the stochastic nature  makes TVS less prone to local convergence and thereby more robust.  

\end{abstract}

\noindent%
{\bf Keywords: \it BART, Combinatorial  Bandits, Interpretable Machine Learning, Spike-and-Slab, Thompson Sampling, Variable Selection}  

\spacingset{1.45} 

\clearpage

\section{Interpretable Machine Learning}\label{sec:setting} 


A fundamental challenge in statistics that goes beyond mere prediction is to glean interpretable insights into the nature of real-world processes by identifying important correlates of variation.
Many today's most powerful prediction tools, however, 
lack an intuitive algebraic form which renders their interpretability (i.e. insight into the black box  decision process)   far from straightforward.
Substantial effort has been recently devoted to enhancing the explainability of machine  learning  through the identification 
of  key variables  that drive predictions (\cite{garson1991comparison,olden2002illuminating,zhang2000adaptive,lu2018deeppink,burns,horel2019towards}).
 While these procedures may possess nice theoretical guarantees, they may not yet be feasible for large-scale applications.
This work develops a new computational platform  for understanding  black-box predictions which is based on reinforcement learning and which  can be applied to very large datasets.

A variable can be important because its change has a causal impact or because leaving it out reduces overall prediction capacity (\cite{mase2019explaining}). 
  Such leave-one-covariate-out type inference has a long tradition,  going back to at least \cite{breiman2001random}.
  In random forests, for example,
variable importance is assessed by the difference between prediction errors in the out-of-bag sample before and after noising the covariate through a permutation.
\cite{lei2018distribution}  propose the LOCO method which gauges  local effects of removing each covariate on the overall prediction capability and derives an asymptotic distribution for this measure to conduct proper statistical tests. There is a wealth of literature on  variable importance measures, see \cite{fisher2019all} for a recent overview.
In Bayesian forests, such as BART (\cite{chipman_practical_2001}), one keeps track of predictor inclusion frequencies and outputs an average proportion of all splitting rules inside a tree ensemble that split on a given variable. In deep learning, one can construct variable importance measures using network weights (\cite{garson1991comparison,ye2018variable}). 
\cite{owen2017shapley} introduce a variable importance based on a Shapley value and \cite{hooker2007generalized}  investigates diagnostics of black box functions using functional ANOVA decompositions with dependent covariates. 
While useful for ranking variables,  importance measures are less intuitive for model selection and are often not well-understood  theoretically (with a few exceptions including \cite{ishwaran2007variable,kazemitabar2017variable}).


This work focuses on  high-dimensional applications (either very many predictors or very many observations, or both), where computing importance measures and performing tests for predictor effects quickly becomes infeasible. 
We consider the  non-parametric regression model which provides a natural statistical  framework for supervised machine learning. The data setup consists of a continuous response vector ${\bf Y}^{(n)} = (Y_1, \cdots, Y_n)'$ that is  linked stochastically to a fixed set of predictors ${\bf x}_i = (x_{i1}, \cdots, x_{ip})'$ for $1\leq i\leq n$ through   
\begin{equation}
\label{eq:basic} 
 Y_i = f_0({\bf x}_i) + \epsilon_i  \quad \text{where }\epsilon_i \iid \text{N}(0, \sigma^2),
\end{equation}
and where $f_0$ is an unknown regression function. The variable selection problem occurs when there is a subset $\mathcal{S}_0 \subset \{ 1,\cdots, p \}$ of   $q_0 =\C{\mathcal{S}_0}\C$ predictors which exert   influence on the mixing function $f_0$ and we do not know which subset it is. In other words, $f_0$ is constant in directions outside $\mathcal{S}_0$ and the goal is to identify active  directions (regressors) in $\mS_0$  while, at the same time, permitting  nonlinearities and interactions. 
The traditional Bayesian approach to this problem starts with a prior distribution over the $2^p$ sets of active variables. This is typically done in a hierarchical fashion by first assigning a prior distribution $\pi(q)$ on the subset size $q=\C{\mS}\C$ and then a conditionally uniform prior on $\mS$, given $q$, i.e. 
$\pi(\mathcal{S}\mid q) =\frac{1}{{p\choose q}}.$ 
This prior can be translated into the {\sl spike-and-slab} prior where, for each coordinate $1\leq i\leq p$, one assumes a binary indicator $\gamma_i$  for whether or not the variable $x_i$ is active and assigns a prior
 \begin{equation}\label{eq:definition}
 \P(\gamma_i  \,\C\,  \theta) =\theta, \quad \theta \sim \texttt{Beta}(a,b) \,\,\text{for some}\,\, a,b>0.
 \end{equation}
 The active subset $\mS$ is then constructed as  $\mS=\{j : \gamma_j =1\}$. There is no shortage of literature on spike-and-slab variable selection  in the linear model,  
addressing prior choices (\cite{mitchell1988bayesian,rockova_spike-and-slab_2018,rossell2017nonlocal,bernardo_bayesian_2011,vannucci1}), computational aspects (\cite{carbonetto2012scalable,rockova_emvs:_2014,Bottolo}, \cite{george1993variable},\cite{george_approaches_1997}) and/or variable selection consistency results (\cite{castillo2015bayesian,johnson2012bayesian,narisetty2014bayesian}).  Traditionally, spike-and-slab methodology relies on the underlying model to be linear, which may be woefully inaccurate, and  can be computationally slow.  In this work, we leave behind the linear model framework and focus on interpretable machine learning linking spike-and-slab methods with binary bandits.
The two major methodological benefits are (a)  ability to capture non-linear effects and (b) scalability to very large datasets. 
Existing non-linear variable selection approaches include grouped shrinkage/selection of basis-expansion coefficients  \citep{lin2006component, radchenko2010variable, ravikumar2009sparse, scheipl2011spikeslabgam},
regularization of the derivative expectation operator  \citep{lafferty2008rodeo} or  model-free knockoffs  \citep{candes_panning_2018}. 
The main distinguishing feature of our approach is the development of a new computational platform via a spike-and-slab wrapper that extends the reach of machine learning to large-scale data.

This paper introduces Thompson Variable Selection (TVS), a stochastic optimization approach to subset selection based on reinforcement learning.
The key idea behind TVS  is  that  variable selection can be regarded as a combinatorial  bandit problem  
where each variable is treated as an arm.
TVS sequentially learns promising combinations of arms (variables) that are most likely to provide a reward.
Depending on the learning  tool for modeling $f_0$ (not necessarily a linear model), TVS accommodates a wide range of rewards  for both offline and online (streaming batches) setups.  The fundamental appeal of active learning for subset selection (as opposed to MCMC sampling) is that  those variables which provided a small reward in the past are less
 likely to be pulled again in the future. This exploitation aspect steers model exploration towards more promising combinations and offers dramatic computational dividends.
Indeed, similarly as with backward elimination TVS narrows down the inputs contributing to $f_0$ but does so in a stochastic way by learning from past mistakes.
TVS aggregates evidence for variable inclusion and quickly separates signal from noise by minimizing regret motivated by the median probability model rule \citep{barbieri2004optimal}.
We provide regret bounds which do not necessarily assume that the arm outcomes be unrelated. In addition, we show strong empirical performance and demonstrate the potential of TVS to meet demands of very large datasets. 

This paper is structured as follows. Section  \ref{sec:mabs} revisits known facts about multi-armed bandits. Section \ref{sec:mab_vs} develops the bandits framework for variable selection and Section \ref{sec:TVS} proposes Thompson Variable Selection and presents a regret analysis. Section \ref{sec:Implementation} presents two implementations (offline and online) on two benchmark simulated data. Section \ref{sec:simu} presents a thorough simulation study and Section \ref{sec:real_data} showcases TVS performance  on  real data. We conclude with a discussion in Section \ref{sec:discussion}.

\vspace{-0.5cm}
\section{Multi-Armed Bandits Revisited}\label{sec:mabs}
Before introducing Thompson Variable Selection,  it might be useful to review several known facts about multi-armed bandits.
 The multi-armed bandit (MAB) problem can be motivated by the following gambling metaphor. A slot-machine player needs to decide between multiple arms. When pulled at time $t$, the $i^{th}$ arm gives a random payout $\gamma_i(t)$. In the Bernoulli bandit problem, the rewards $\gamma_i(t)\in\{0,1\}$ are binary and  $\P(\gamma_i(t)=1)=\theta_i$. The distributions of  rewards are unknown and the player can only learn about them through playing. In doing so, the player faces a dilemma: {\sl exploiting} arms that  have provided high yields in the past and {\sl exploring} alternatives that may give higher rewards in the future.

More formally, an algorithm for MAB must decide which of the $p$ arms  to play at time $t$, given the outcome of the previous $t-1$ plays.   A natural goal in the MAB game is to minimize {\sl regret}, i.e. the amount of money one loses by not playing the optimal arm at each step.
Denote with $i(t)$ the arm played at time $t$, with $\theta^\star=\max\limits_{1\leq i\leq p}\theta_i$ the best average reward and with $\Delta_i=\theta^\star-\theta_i$ the gap between the rewards of an optimal action and a chosen action.  The expected regret after $T$ plays can be then written as 
$
\E [\mathcal{R}(T)]=\sum_{i=1}^p\Delta_i\E[k_i(T)],
$
where $k_j(T)=\sum_{t=1}^T\mathbb{I}[i(t)=j]$ is the number of times an arm $j$ has been played up to step $T$.  There have been two main types of algorithms designed to  minimize regret in the MAB problem:
Upper Confidence Bound (UCB)  of \cite{lai_asymptotically_1985} and Thompson Sampling (TS) of  \cite{thompson_likelihood_1933}. 
Thompson Sampling  is a   Bayesian-inspired heuristic algorithm  that  achieves a logarithmic expected regret \citep{agrawal_analysis_2012} in the Bernoulli bandit problem.
Starting with a non-informative prior  $\theta_i\iid \texttt{Beta}(1,1)$ for $1\leq i\leq p$,  this algorithm: (a) updates the distribution of $\theta_i$ as $\texttt{Beta}(a_i(t)+1,b_i(t)+1)$, 
where $a_i(t)$  and $b_i(t)$ are the number of successes and  failures of the arm $i$ up to time $t$,  (b) samples $\theta_i(t)$ from these posterior distributions, and (c) plays the arm with the highest  $\theta_i(t)$. 
 \cite{agrawal_analysis_2012} extended this algorithm to the general case where  rewards are not necessarily Bernoulli but general random variables on the interval $[0,1]$. 


The MAB problem is most often formulated as a single-play problem, where only one arm can be selected at each round. \cite{komiyama_optimal_2015} extended Thompson sampling to a {\em multi-play scenario}, where  at each round $t$ the player selects a subset $\mS_t$ of $L<p$ arms  and receives binary rewards of all  selected arms. For each $1\leq i\leq p$, these rewards $r_i(t)$ are iid  Bernoulli with unknown success probabilities $\theta_i$ where $\gamma_i(t)$ and $\gamma_j(t)$ are  independent for $i\neq j$ and where, without loss of generality, $\theta_1>\theta_2>\dots> \theta_p$.  The player is interested in maximizing the sum of expected rewards over drawn arms, where the optimal action is playing the top $L$ arms $\mS_0=\{1,\dots, L\}$. The regret depends on the combinatorial structure of arms drawn and, similarly as before, is  defined as the gap between an expected cumulative reward and the optimal drawing policy, i.e.
$
\E [\mathcal{R}(T)]=\E\sum_{t=1}^T\left(\sum_{i\in  \mS_0}\theta_i-\sum_{i\in \mS_t}\theta_i\right)
$
Fixing $L$, the number of arms played, \cite{komiyama_optimal_2015} propose a Thompson sampling  algorithm for this problem and show that it has a logarithmic expected regret with respect to time  and a linear regret with respect to the number of arms. Our  metamorphosis  of multi-armed bandits  into a  variable selection algorithm will ultimately require that the number $L$ of arms played  is random and that the rewards at each time $t$ can be dependent.

Finally, we complete the review of MAB techniques with {\em combinatorial  bandits} (\cite{chen2013combinatorial,gai2012combinatorial,cesa2012combinatorial}) which are the closest relative to our proposed method here. Combinatorial bandits can be seen as a generalization of multi-play bandits, where any arbitrary combination of arms $\mS$ (called super-arms) is played at each round and where the reward $r(\mS)$  can be revealed for the entire collective    $\mS$ (a full-bandit feedback) or for each contributing arm $i\in\mS$ (a semi-bandit feedback), see e.g. \cite{wang2018thompson,combes2014unimodal,kveton2015combinatorial,combes2014unimodal,kveton2015combinatorial}.  We will draw upon connections between combinatorial bandits and variable selection multiple times throughout Section 3 and 4.

\vspace{-0.5cm}
\section{Variable Selection as a Bandit Problem}\label{sec:mab_vs}
 The purpose of this section is to link spike-and-slab model selection with multi-armed bandits. Before formalizing the ideas, we discuss two possibilities inspired by the search for the MAP (maximum-a-posteriori) model and the MPM (median probability) model.

 Bayesian model selection {\color{blue}with spike-and-slab priors} has often been  synonymous to finding the MAP  model $\wh\mS=\arg\max_{\mS}\pi(\mS\,\mid\,\Y^{(n)})$.  Even when the marginal likelihood is available, this model  can computationally unattainable for $p$ as small as $20$. In order to accelerate Bayesian variable selection using multi-armed bandits techniques one idea immediately comes to mind.
One could treat each of the $2^p$ {\em models}   as a base arm. Assigning  prior model probabilities  according to 
$\theta_i\sim \texttt{Beta}(a_i,b_i)$  for $1\leq i\leq 2^p$ for some\footnote{chosen to correspond to marginals of a Dirichlet distribution} $a_i>0$ and $b_i>0$, one  could  play a game by sequentially trying out various arms (variable subsets) and collect  rewards to prioritize  subsets that were suitably ``good". 
Identifying the arm with the highest mean reward could then serve as a proxy for the best model.
 This naive strategy, however, would  not  be operational due to the exponential number of arms to explore. 

Instead of the MAP model, it has now been standard practice to report  the {\sl median probability model} (MPM)  (\cite{barbieri2004optimal}) consisting of those variables  whose  posterior inclusion probability 
$
\pi_i\equiv \P(\gamma_i=1\,\mid\, \Y^{(n)})
$
is at least $0.5$. More formally,  MPM is defined,   for  $\bm \pi=(\pi_1,\dots,\pi_p)'$,  as
\begin{equation}\label{eq:MPM}
\wh\mS_{MPM}=\arg\max_{\mS}r_{\bm\pi}(\mS)=\{i:\pi_i\geq 0.5\}\quad\text{where}\quad r_{\bm\pi}(\mS)=\left\{ \prod_{i\in\mS}\pi_i\prod_{i\notin\mS}(1-\pi_i)\right\}.
\end{equation}
 This is now the default model selection rule with spike-and-slab priors \eqref{eq:definition}. 
The MPM  model is the optimal predictive model in linear regression under some assumptions \citep{barbieri_median_2018}. 
Obtaining $\pi_i$'s, albeit easier than finding the MAP model,
 requires posterior sampling over variable subsets. While this can be done using standard MCMC sampling techniques in linear regression (\cite{george_approaches_1997,narisetty2014bayesian,bhattacharya2016fast}),
here we explore new curious connections to bandits in order to develop a much faster stochastic optimization routine for finding MPM-alike models when the true model is not necessarily linear.

Having reviewed the two traditional Bayesian model choice reporting methods (MAP and MPM), we can now forge connections to multi-armed bandits.
 While the MAP model suggests treating {\em each model} $\mS$ as a bandit arm, the MPM model suggests treating {\em each variable} $\gamma_i$ as a bandit arm. Under the MAP framework, the player  would be required to play  a single arm (i.e. a model) at each step. The MPM framework, on the other hand,  requires playing a random subset of arms (i.e. a model) at each play opportunity.
This is appealing for at least two reasons: (1) there are fewer arms to explore more efficiently, (2)   the quantity $r_{\bm\pi}(\mS)$ in \eqref{eq:MPM} can be regarded as a mean regret of a combinatorial arm (more below) which, given $\bm\pi$, has MPM as its computational oracle.  The computational oracle is defined as the regret minimizer when an oracle furnishes yield probabilities $\theta_i$ (see forthcoming Lemma \ref{lemma:oracle}). Based on the discussion above, we regard the MPM framework as more intuitively appealing for bandit techniques. 
We thereby reframe spike-and-slab selection with priors \eqref{eq:definition} as a bandit problem treating each variable as an arm. This idea is formalized below. 

We view Bayesian spike-and-slab selection  through the lens of {\em combinatorial bandit problems} {\color{blue} (reviewed earlier in Section  \ref{sec:mabs})} by treating variable selection indicators $\gamma_i$'s in \eqref{eq:definition}  as  Bernoulli rewards. From now on, we will refer to each $\theta_i$ as an unknown mean reward, i.e. a probability that the $i^{th}$ variable exerts influence on the outcome.
In sharp contrast to \eqref{eq:definition} which deploys one $\theta$ for all arms,  each arm  $i\in\{1,\dots,p\}$ now  has {\sl its own} prior inclusion probability $\theta_i$, i.e.
\begin{equation}\label{eq:thetas}
\P(\gamma_i=1 \,\origbar\, \theta_i) =\theta_i, \,\, \theta_i \ind \texttt{Beta}(a_i,b_i) \,\,\text{for some}\,\, a_i,b_i>0.
\end{equation}
 In the original spike-and-slab setup \eqref{eq:definition}, the mixing weight $\theta$ served as a global shrinkage parameter determining the level of sparsity and 
linking  coordinates to borrow strength (\cite{rockova_spike-and-slab_2018}). In our new bandit formulation \eqref{eq:thetas}, on the other hand,  the reward probabilities $\theta_i$ serve as a proxy for  posterior inclusion probabilities $\pi_i$ whose distribution we want to learn by playing the bandits game. 
Recasting  the spike-and-slab prior in this way allows one to approach Bayesian variable selection from a more algorithmic (machine learning) perspective.

\subsection{The Global Reward}
Before proceeding, we need to  define the reward  in the context of variable selection. 
One conceptually appealing strategy would be to collect a joint reward $R(\mS_t)$  (e.g. goodness of model fit) reflecting the collective effort of all contributing arms and then redistribute it among arms inside the super-arm $\mS_t$ played at time $t$.  One example would be the Shapley value (\cite{shapley1953value,owen2017shapley}), a construct from cooperative game theory for the attribution problem that distributes the value created by a team to its individual members. 

We try a different route.Rather than distributing, we will aggregate. Namely, instead of collecting a global reward first and then redistributing it, we first collect individual rewards $\gamma_i^t\in\{0,1\}$ for each played arm  $i\in \mS_t$ and then weave them into a global reward $R(\mS_t)$.
We assume that  $\gamma_{i}^t$'s  are  \texttt{iid} from \eqref{eq:thetas}  for each $i\in\{1,\dots,p\}$.  Unlike traditional combinatorial bandits  that define the global reward $R(\mS_t)=\sum_{i\in\mS}\gamma_i^t$ as a sum of individual outcomes \citep{gai2012combinatorial}, we consider a global reward for variable selection motivated by the median probability model.

One natural choice  would be a {\em binary} global reward $R(\mS_t)=\prod_{i\in\mS_t}\gamma_i^t\prod_{i\notin\mS_t}(1-\gamma_i^t)\in\{0,1\}$  for whether or not {\em all} arms inside $\mS_t$ yielded a reward and, at the same time, {\em none} of the arms outside $\mS_t$ did.   Assuming independent arms, the expected reward equals  $\E [R(\mS_t)]=\prod_{i\in\mS_t}\theta_i\prod_{i\notin\mS_t}(1-\theta_i)=r_{\bm\theta}(\mS_t)$  and has  the ``median probability model" as its computational oracle, as can be seen from \eqref{eq:MPM}.
However, this expected reward is not monotone in $\theta_i$'s (a requirement  needed for  regret analysis) and, due to its dichotomous nature, it penalizes all mistakes (false positives and negatives) equally. 

We consider an alternative reward function which also admits a computational oracle but treats mistakes {\em differentially}.
For some $0<C<1$, we define the {\em global reward} $R_C(\mS_t)$ for a subset $\mS_t$ at time $t$   as 
\begin{equation}\label{eq:global_reward}
R_C(\mS_t)=\sum_{i\in \mS_t} \log\left(C+\gamma_i^t\right).
\end{equation}
Similarly as $R(\mS_t)$ (defined above)  the reward is maximized  for the model which includes all the positive arms and none of the negative arms, i.e. $\arg\max_\mS R_C(\mS)=\{i:\gamma_i^t=1\}$.
Unlike $R(\mS_t)$, however, the reward will penalize subsets with false positives, a penalty $\log (C)$ for each, and there is an opportunity cost of $\log(1+C)$ for each false negative.
The expected global reward depends on the subset $\mS_t$ and the vector of yield probabilities $\bm\theta=(\theta_1,\dots,\theta_p)'$, i.e.
\begin{equation}\label{eq:expected_reward}
r_{\bm\theta}^C(\mS_t)=\E \left[R_C(\mS_t)\right]= \sum_{i\in \mS_t}\left[\theta_i \log\left(\frac{C+1}{C}\right)-\log\left(\frac{1}{C}\right)\right].
\end{equation}
 Note that this expected reward {\em is monotone} in $\theta_i$'s and  is Lipschitz continuous. Moreover,  it {\em also has} the median probability model as its computational oracle. 
 \begin{lemma}\label{lemma:oracle}
 Denote with $\mS_{O}=\arg\max_{\mS} r_{\bm\theta}^C(\mS)$ the computational oracle. Then we have  
 \begin{equation}\label{eq:oracle}
 \mS_{O}=\left\{i: \theta_i\geq \frac{\log(1/C)}{\log[(C+1)/C]}\right\}.
 \end{equation}
 With $C=(\sqrt{5}-1)/2$, the oracle is the median probability model  $\{i:\theta_i\geq 0.5\}$.
  \end{lemma}
 \proof
It follows immediately from the definition of $R_C(\mS_t)$ and the fact that $\log(1/C)=0.5 \log[(1+C)/C]$  for  $C=(\sqrt{5}-1)/2$. 
 
Note that the choice of $C=(\sqrt{5}-1)/2$ incurs {\em the same} penalty/opportunity cost for false positives and negatives since $\log(1+C)=-\log(C)$. 
In streaming feature selection, for example, \cite{zhou2006streamwise} accommodate measurement cost and place cheaper variables earlier in the stream. 
In our framework, we can allow for different cost $C_i$ (e.g. a measurement cost) for each variable $1\leq i\leq p$.
The existence of the computational oracle for the expected reward $r_{\bm\theta}^C(\mS)$  is very comforting and will be exploited in our Thompson sampling algorithm introduced in Section \ref{sec:TVS}

\subsection{The Local Rewards}\label{sec:local_reward}
The global reward \eqref{eq:global_reward} is a deterministic functional of the local rewards.  We have opted for the reward functional \eqref{eq:global_reward} because the regret minimizer is the median probability model when the yield probabilities are provided (see Lemma \ref{lemma:oracle}). We now clarify the definition of local rewards $\gamma_i^t$.
We regard $\mS_t$ as a smaller pool of candidate variables, which can contain false positives and false negatives. The goal is to play a game by sequentially trying out different subsets and reward true signals so that they are selected in the next round and to discourage  false positives  from being included again in the future. 
Denote with  $\mathbb{S}$ the set of all subsets of $\{1,\dots, p\}$ and with $\mathcal{D}$ the ``data" at hand consisting of $\mid\mathcal D\mid$ observations $(Y_i,\bm x_{i})$ from \eqref{eq:basic}.
We introduce a  {\em feedback rule}
\begin{equation}\label{eq:variable_screening}
r(\mS_t,\mathcal D):\mathbb{S}\times \R^{\mid\mathcal{D}\mid}\rightarrow \{0,1\}^{\mid\mS_t\mid},
\end{equation}
which, when presented with data $\mathcal D$ and a subset $\mS_t$, outputs a  vector of {\em binary rewards} $r(\mS_t,\mathcal{D})=(\gamma_i^t: i\in\mS_t)'$ for whether or not a variable $x_i$  for $i\in\mS_t$ is relevant for predicting or explaining the outcome. This feedback is only revealed if $i\in\mS_t$.
We consider two sources of randomness that implicitly define the reward distribution $r(\mS_t,\mathcal D)$:  (1) a  {\em stochastic feedback rule} $r(\cdot)$ assuming  that data $\mathcal D$ is given, and  (2)  a {\em deterministic feedback rule} $r(\cdot)$ assuming that  data $\mathcal D$ is stochastic.

The first reward type has a Bayesian flavor in the sense that it is {\em conditional} on the observed data $\mathcal D_{n}=\{(Y_i,\bm x_i): 1\leq i\leq n\}$, where rewards can be sampled using Bayesian stochastic computation (i.e. MCMC sampling). Such rewards are  natural in  {\em offline settings} with  Bayesian feedback rules, as we explore in Section \ref{sec:offline}. 
As a lead example of this strategy in this paper,  we consider a stochastic reward based on BART (Chipman et al. (2001)). 
We refer to \cite{hill_review} and references therein for a nice recent overview of BART.
In particular, we use the following binary local reward $r(\mS_t,\mathcal D_n)=(\gamma_i^t:i\in\mS_t)'$ where
\begin{equation}\label{eq:reward_binary}
\gamma_i^t=\mathbb{I}(\text{$M^{th}$ sample from the BART posterior splits on the variable $x_i$}).
\end{equation}
The mean reward $\theta_i=\P(\gamma_i^t=1)=\P[i\in \mathcal F\,\C\,\mathcal D_n]$  can be interpreted as the posterior probability that a variable $x_i$ is split on in a Bayesian forest $\mathcal F$ given the entire data $\mathcal D_n$. 
The stochastic nature of the BART computation  allows one to regard the reward \eqref{eq:reward_binary} as an actual random variable, whose values can be sampled from using standard software. Since BART  is run  {\em only} with variables inside $\mS_t$ (where $\mid\mS_t\mid<<p$) and   {\em only} for $M$ burn-in MCMC iterations, computational gains are dramatic (as we will see in Section \ref{sec:offline}).

The second  reward type has a frequentist flavor in the sense that rewards are sampled by applying 
deterministic feedback rules on new streams (or bootstrap replicates) $\mathcal D_t$ of data. 
 Such rewards are  natural in  {\em online settings}, as we explore in Section \ref{sec:online}. As a lead example of this strategy in this paper,  we assume that the dataset $\mathcal D_n$ consist of $n=sT$ observations and is partitioned into  minibatches $\mathcal D_t=\{(Y_i,\bm x_i): (t-1)s+1\leq i\leq ts\}$  for $t=1,\dots, T$. One could think of these batches as new  independent observations  arriving in an online fashion or as manageable snippets of big data. 
The `deterministic' screening rule we consider here is  running BART  for a large number $M$ of MCMC iterations and collecting an aggregated importance measure $IM(i;\mathcal D_t,\mS_t)$ for each variable.\footnote{
This rule is deterministic in the sense that computing it again on the same data should in principle provide the same answer. One could, in fact, deploy any other machine learning method that outputs some measure of variable importance.} 
We define  $IM(i;\mathcal D_t,\mS_t)$ as the average number of times a variable $ x_i$ was used in a forest where the average is taken over the $M$ iterations and  we then reward those arms which were used at least once on average,
\begin{equation}\label{eq:reward_binary2}
\gamma_i^t=\mathbb{I}[IM(i;\mathcal D_t,\mS_t)\geq1].
\end{equation}
The mean reward $\theta_i=\P(\gamma_i^t=1)$  can be then interpreted as the (frequentist) probability that BART, when run on $s=n/T$ observations arising from   \eqref{eq:basic}, uses a variable $x_i$ at least once on average over $M$ iterations. We illustrate this online variant in Section \ref{sec:online}.

\vspace{-0.5cm}
\subsection{Other Feedback Rules} 
 TVS is not confined to a BART reward function. Deploying any variable selection method using only a subset $\mS_t$ will yield a binary feedback rule.
For example,  the LASSO method yields  $\hat\beta_i$ for $i\in\mS_t$ which can be turned into the following feedback  
 $\gamma_i^t=\mathbb I(\hat\beta_i\neq 0)$ for $i\in\mS_t$. In offline setups, randomness of $\gamma_i^t$'s  can be induced by taking a bootstrap replicate of $\mathcal D_n$ at each play time $1\leq t\leq T$. Another possibility for a binary reward is  dichotomizing $p$-values, similarly as in alpha-investing  for streaming variable selection by \cite{zhou2006streamwise}.
Instead of binary local rewards \eqref{eq:variable_screening},  one can also consider continuous rewards $\gamma_i^t\in[0,1]^{\mid\mS_t\mid}$ by rescaling  variable importance measures obtained by a machine learning method (e.g. random forests (\cite{louppe2013understanding}),  deep learning (\cite{horel2019towards}), and  BART (\cite{chipman_bart:_2010})). 
Our Thompson sampling algorithm can be  then modified  by dichotomizing these rewards through independent Bernoulli trials with  probabilities equal to the continuous rewards  \citep{agrawal_analysis_2012}.

\vspace{-0.5cm}
\section{Introducing Thompson Variable Selection (TVS)}\label{sec:TVS}
This section introduces Thompson Variable Selection (TVS), a reinforcement learning algorithm for subset selection in non-parametric regression environments. The computation alternates between {\em Choose, Reward} and {\em Update} steps that we  describe in more detail below.

The  unknown mean rewards will be denoted with $\theta_i^\star$ and the ultimate goal of TVS is to learn their distribution once we have seen the `data'\footnote{The `data' here refers to the sequence of observed rewards.}.  To this end, we  take  the  {\em combinatorial bandits}  perspective (\cite{chen2013combinatorial}, Gai et al. (2010))  where, instead of playing one arm at each play opportunity $t$, we  play a random subset  $\mS_t\subseteq \{1,\dots,p\}$ of multiple arms. Each such {\em super-arm} $\mS_t$ corresponds to a {\em model} configuration and the goal is to discover promising models by playing more often the more promising variables.

Similarly as with traditional Thompson Sampling, the $t^{th}$ iteration of TVS starts off by sampling mean rewards $\theta_i(t)\sim\mathrm{Beta}(a_i(t),b_i(t))$  from a posterior distribution  that incorporates past reward experiences up to time $t$ (as we discussed in Section \ref{sec:mabs}). 
The {\em Choose Step} then decides which arms will be played in the next round.  While the single-play Thompson sampling policy dictates playing the arm with the highest sampled expected reward, the combinatorial Thompson sampling policy (\cite{wang2018thompson}) dictates playing the {\em subset} that maximizes the expected global reward, given the vector of  sampled probabilities $\bm\theta(t)=(\theta_1(t),\dots,\theta_p(t))'$.
 The availability of the computational oracle (from Lemma \ref{lemma:oracle}) makes this step awkwardly simple as it boils down to computing $\mS_{O}$ in \eqref{eq:oracle}. 
 Unlike multi-play bandits where the number of played arms is predetermined (Komiyama (2015)), this strategy allows one to adapt to the size of the model.
 We do,  however,   consider a variant of the computational oracle (see \eqref{eq:oracle_q} below) for when the size $q^\star=\mid\mS^\star\mid$ of the `true' model $\mS^\star=\arg\max_\mS r_{\bm\theta^\star}^C(\mS)$ is known.
 The  {\em Choose Step}  is then followed by the {\em Reward Step} (step R in Table \ref{alg:vanillathompson}) which assigns a prize to the chosen subset $\mS_t$ by collecting individual rewards $\gamma_{i}^t$ (for the offline setup in \eqref{eq:reward_binary} or for the online setup \eqref{eq:reward_binary2}). Finally, each TVS iteration concludes with an  {\em Update Step} which updates the beta posterior distribution (step U in Table \ref{alg:vanillathompson}).

{\color{blue}
\begin{algorithm}[!t]
{ \centering
	\spacingset{1.1}
	\small
	\scalebox{0.95}{
		\begin{tabular}{l l}
			\hline\hline
			\multicolumn{2}{c}{{\bf Algorithm 1:}\sl \hspace{0.2cm}  \cellcolor[gray]{0.8}  Thompson Variable Selection with  BART}\\
			\hline
			\multicolumn{2}{c}{\bf INPUT \cellcolor[gray]{0.6} }\\			
			\multicolumn{2}{c}{Define $\wt C=\frac{\log(1/C)}{\log[(1+C)/C}$ for some $0<C<1$ and pick $M,a,b>0$ }\\
			\multicolumn{2}{c}{ Initialize $a_i(0) := a$ and $ b_i(0) := b $ for each arm $1\leq i\leq p$.} \\
			\multicolumn{2}{c}{\bf LOOP \cellcolor[gray]{0.6}}\\			
			\multicolumn{2}{c}{For $t=1,\dots, T$ repeat steps C(1)-(3), R and U.} \\
			\multicolumn{2}{c}{\bf  \cellcolor[gray]{0.9}Choose Step}\\
			C(1):  Set $\mS_t=\emptyset$ and  for $ i= 1 \cdots p$ do  &\\
			C(2): Sample ${\theta}_{i}(t)\sim \texttt{Beta}(a_i(t),b_i(t))$ &\\
			C(3): (Unknown $q^\star$) Compute $
			\mS_t=\{i:\theta_i(t)\geq \wt C\}$  from \eqref{eq:oracle} &\\
			C(3)$^*$: (Known $q^\star$) Compute $\mS_t$ from \eqref{eq:oracle_q} &\\
			\multicolumn{2}{c}{\bf \cellcolor[gray]{0.9} Reward Step}\\
			R: Collect local rewards $\gamma_i^t$ for each $1\leq i\leq p$ from \eqref{eq:reward_binary} (offline) or  \eqref{eq:reward_binary2} (online)\\
			\multicolumn{2}{c}{\bf \cellcolor[gray]{0.9} Update Step}\\
			U:  If $\gamma_i^t=1$ then set $a_i(t+1) = a_i(t) +1$, else $b_i(t+1) = b_i(t) + 1$&\\
						\multicolumn{2}{c}{\bf OUTPUT \cellcolor[gray]{0.6}}\\			
		\multicolumn{2}{c}{Evidence probabilities  $\pi_i(t)=a_i(t)/[a_i(t)+b_i(t)]$ for $1\leq i\leq p$ and $1\leq t\leq T$. }\\
			\hline\hline
	\end{tabular}}
	
}
\caption{\small  Thompson Variable Selection with  BART ($^\star$ is an alternative with known $q^\star$)} 
\label{alg:vanillathompson}
\end{algorithm}
\label{sec:Vanilla}
}

The fundamental goal of Thompson Variable Selection is to learn the distribution of mean rewards $\theta_i$'s by playing a game, i.e. sequentially creating a dataset of rewards  by sampling  from  beta posterior\footnote{This posterior treats the past rewards as the actual data.}  distributions that incorporate past rewards and the observed data $\mathcal D$. One natural way to distill evidence for variable selection is through the means $\bm\pi(t)=(\pi_1(t),\dots,\pi_p(t))'$ of these beta distributions 
\begin{equation}\label{eq:pi}
\pi_i(t)=\frac{a_i(t)}{a_i(t)+b_i(t)},\quad 1\leq i\leq p,
\end{equation}
which serve as a proxy for posterior inclusion probabilities. Similarly as with the classical median probability model \citep{barbieri2004optimal},
one can deem important those variables with $\pi_i(t)$  above $0.5$  (this corresponds to one specific choice of $C$ in Lemma \ref{lemma:oracle}). More generally, at each iteration $t$ TVS outputs a model $\wh\mS_t$, which satisfies
$\wh\mS_t=\arg\max_{\mS}r^C_{\bm \pi(t)}$. From Lemma \ref{lemma:oracle},  this model can be simply computed by truncating individual $\pi_i(t)$'s.
Upon convergence, i.e. when trajectories $\pi_i(t)$  stabilize over time, TVS will output the same model $\wh\mS_t$. We will see from our empirical demonstrations in Section \ref{sec:Implementation} that the separation between signal and noise (based on $\pi_i(t)$'s) and the model stabilization occurs fast.
Before our empirical results, however, we will dive into the regret analysis of TVS.

\subsection{Regret Analysis} 
Thompson sampling (TS) is a policy that uses Bayesian ideas to solve a fundamentally frequentist problem of regret minimization.
In this section, we explore  regret properties of TVS  and expand current theoretical understanding of combinatorial TS  by allowing for correlation between arms. Theory for TS was essentially unavailable until the path-breaking paper by
 \cite{agrawal_analysis_2012}  where the first finite-time analysis was presented for single-play bandits. 
 Later, \cite{leike_thompson_2016} proved that TS converges to the optimal policy in probability and almost surely under some  assumptions.  
Several theoretical and empirical studies for TS in {\em multi-play} bandits are also available.
In particular, \cite{komiyama_optimal_2015} extended TS to multi-play problems with a fixed number of played arms and showed that it achieves the optimal regret bound. Recently, \cite{wang2018thompson}  introduced TS for combinatorial bandits and derived regret bounds for Lipschitz-continuous rewards under an offline oracle.
We build on their development and extend their results  to the case of related arms.

Recall that the goal of the player is to minimize the total (expected) regret under time horizon $T$ defined below
\begin{equation}\label{eq:regret}
Reg(T)=\E\left[\sum_{t=1}^T\left(r_{\bm\theta^\star}^C(\mS^\star)- r_{\bm\theta^\star}^C(\mS_t)\right) \right],
\end{equation}
where $\mS^\star=\arg\max_{\mS}r_{\bm\theta^\star}^C(\mS_t)$ with $q^\star=\mid \mS^\star \mid$, $\theta_i^\star=\E[\gamma_i^t]$ and where the expectation is taken over the unknown drawing policy. Choosing $C$ as in Lemma \ref{lemma:oracle}, one has $\log(1+C)=-\log(C)=D$ and thereby
$$
Reg(T)=D\,\E\left[\sum_{t=1}^T\sum_{i=1}^p (2\theta_i^\star-1)[\mathbb{I}(i\in\mS^\star\backslash \mS_t)- \mathbb{I}(i\in\mS_t\backslash \mS^\star)]\right].
$$
Note that $(2\theta_i^\star-1)$ is positive iff $i\in\mS^\star$.
Upper bounds for the regret \eqref{eq:regret} under the drawing policy of our TVS Algorithm 1 can be obtained under various assumptions.  Below, we first review two available regret bounds following from
\cite{wang2018thompson} for when (a) $q^\star$ is known and arms are independent (Lemma \ref{lemma:known}) and (b) arms are related and $q^\star$ is unknown (Lemma \ref{lemma:indep}). Later in  our Theorem \ref{the:corr}, we relax these assumptions and provide a regret bound assuming that $q^\star$ unknown and, at the same time,  arms are related.

Assuming that the   size $q^\star$ of the optimal model $\mS^\star$  is {\em known}, one can modify Algorithm 1 to confine the search to models of size up to $q^\star$. Denoting $\mathcal I=\{\mS\subset\{1,\dots,p\}:\mid\mS\mid\leq q^\star\}$, one plays the optimal set of arms {\em within the set $\mathcal I$}, i.e. replacing the computational oracle in  \eqref{eq:oracle} with 
$\mS_O^{q^\star}=\arg\max_{\mS\in\mathcal I}r_{\bm\theta}(\mS)$. We denote this modification with $C2^\star$ in Table \ref{alg:vanillathompson}. It turns out that this  oracle can also be easily computed, where the solution  consists of   (up to) the top $q^\star$ arms that pass the selection threshold, i.e.
\begin{equation}\label{eq:oracle_q}
\mS_O^{q^\star}=\left\{i:\theta_i\geq  \frac{\log(1/C)}{\log[(1+C)/C]}\right\}\cap J(\bm\theta)=\{(i_1,\dots, i_{q^\star})'\in\mathbb N^{q^\star}:\theta_{i_1}\geq\theta_{i_2}\geq\dots \geq \theta_{i_q^\star}\}.
\end{equation} 
We have the following regret bound which, unlike the majority of existing results for Thompson sampling, {\em does not} require the arms to have independent outcomes $\gamma_i^t$. The regret bound depends on the amount of separation between signal and noise.
\begin{lemma}\label{lemma:known}
Define the identifiability gap  $\Delta_i=\min \left\{\theta^\star_j: \theta^\star_j>\theta^\star_i\,\,\text{for}\,\,  j\in\mS^\star\right\}$ for each arm $i\notin\mS^\star$.
The Algorithm 1 with a computational oracle $\mS_O^{q^\star}$ in C2$^\star$ achieves the following regret bound
$$
Reg(T)\leq \sum_{i\notin \mS^\star }\frac{(\Delta_i-\varepsilon)\log T }{(\Delta_i-2\varepsilon)^2}+C\left(\frac{p}{\varepsilon^4}\right)+p^2
$$
for any $\varepsilon>0$ such that $\Delta_i>2\varepsilon$ for each $i\notin\mS^\star$ and for some constant $C>0$.
\end{lemma}
\proof  Since $\mathcal I$ is a matroid (\cite{kveton2014matroid}) and our mean regret function is Lipschitz continuous and it depends only on expected rewards of revealed arms, one can apply Theorem 4 of \cite{wang2018thompson}.

Assuming that the  size $q^\star$ of the optimal model  is {\em unknown} and the rewards $\gamma_i^t$ are {\em independent}, one can derive the following bound for the original Algorithm 1 (without restricting the solution to up to $q^\star$ variables).
\begin{lemma}\label{lemma:indep}
Define the maximal reward gap   $\Delta_{max}=\max_\mS\Delta_\mS$ where $\Delta_\mS= [r_{\bm\theta}(\mS^\star)-r_{\bm\theta}(\mS)]$ and for each arm $i\in\{1,\dots,p\}$ define 
$
\eta_i\equiv \max_{\mS: i\in\mS}\frac{8B^2\mid\mS\mid}{\Delta_\mS-2B(q^{\star2}+2)\varepsilon}$ for $B=\log[(C+1)/C]$.
Assume that $\gamma_i^t$'s are independent for each $t$. Then the Algorithm 1  achieves the following regret bound
$$
Reg(T)\leq \log(T)\sum_{i=1}^p \eta_i+p\left(\frac{p^2}{\varepsilon^2}+3\right)\Delta_{max}+C\frac{8\Delta_{max}}{\varepsilon^2}\left(\frac{4}{\varepsilon^2}+1\right)^{q^\star}\log(q^\star/\varepsilon^2)
$$
 for some constant $C>0$ and for any $\varepsilon>0$ such that $\Delta_\mS>2B(q^{\star2}+2)\varepsilon$ for each $\mS$.
\end{lemma}
\proof Follows from Theorem 1 of \cite{wang2018thompson}.

The bandit literature has largely focused on studying the regret  in terms of time $T$ rather than the number of arms $p$. Note that the dependence on $p$ in Lemma \ref{lemma:indep} is cubic, which  (albeit relevant for large $n$ setups) makes the bound less useful when $p$ is very large.  \cite{lai_asymptotically_1985} showed a lower regret bound (in terms of $p$) that   is  $\mathcal{O}(p)$ for any bandit algorithm. \cite{agrawal_analysis_2012} (Remark 3) further showed that Thompson Sampling does achieve this lower bound in a single-play bandit problem. Our Theorem \ref{the:corr} below shows a linear dependence on $p$ for combinatorial bandits with correlated arms when $q^\star$ is unknown.

We now  extend Lemma \ref{lemma:indep} to the case  when the rewards obtained from pulling different arms are related to one another.  \cite{gupta2018correlated} introduced a correlated single-play bandit version of Thompson sampling using pseudo-rewards (upper bounds on the conditional mean reward of each arm). 
{Similarly as with structured bandits (\cite{pandey2007multi}) we instead interweave  the arms by allowing their mean rewards to depend on $\mS_t$, i.e. instead of a single success probability $\theta_i$ we now have
\begin{equation}\label{eq:probs_S}
\theta_i(\mS) = \Prob(\gamma_i^t = 1 \mid \mS_t = \mS).
\end{equation}  
We are interested in obtaining  a regret bound for the Algorithm \ref{alg:vanillathompson}  assuming \eqref{eq:probs_S} in which case the expected global regret \eqref{eq:expected_reward}  writes as  
	\begin{equation}\label{eq:expected_reward_corr}
	r_{\bm\theta}^C(\mS_t)=\E \left[R_C(\mS_t)\right]= \sum_{i\in \mS_t}\left[\theta_i(\mS_t) \log\left(\frac{C+1}{C}\right)-\log\left(\frac{1}{C}\right)\right].
	\end{equation}
To this end we impose an identifiability assumption, which requires a separation gap between the reward probabilities of signal and noise arms.
\begin{assumption}	\label{assumption_identifiability}
Denote with $\mS^\star=\arg\max_{\mS}r_{\bm\theta^\star}^C(\mS_t)$ the optimal set of arms.
We say that  $\mS^\star$ is strongly identifiable if there exists $0<\alpha<1/2$ such that
	\begin{align*}
		\forall i \in \mS^*\quad\text{we have}\quad \theta_i(\mS^\star)\geq \theta_i(\mS) &> 0.5 + \alpha \quad \forall \mS \quad \text{such that} \quad i\in \mS,\\
		\forall i \notin \mS^*\quad\text{we have} \quad\theta_i(\mS) &<0.5 - \alpha  \quad \forall \mS  \quad \text{such that}\quad i\in \mS.
		\end{align*}
\end{assumption}
Under this assumption we provide the following regret bound.
\begin{theorem}
	\label{the:corr}
	 Suppose  that  $\mS^\star$ is strongly identifiable with $\alpha>0$. Choosing $C$ as in Lemma \ref{lemma:oracle}, the regret of Algorithm \ref{alg:vanillathompson} satisfies
	\begin{equation}
	Reg(T)\leq \Delta_{\max} \Bigg[\frac{8 p\log(T)}{\alpha^2} +c(\alpha) q^*  + 
	\left(2 + \frac{4}{\alpha^2}\right)p\Bigg],
\end{equation}
	where $c(\alpha) =\wt C\, \left[\frac{\e^{-4\alpha}}{1 - \e^{-\alpha^2 /2}} \right]+ \frac{8}{\alpha^2} \frac{1}{\e^{2\alpha}-1} + \frac{\e^{-1}}{1 - \e^{-\alpha/8}}+\left\lceil\frac{8}{\alpha}\right\rceil\left(3+\frac{1}{\alpha}\right)$ and $\wt C= \left(C_1+C_2\frac{1-2\alpha}{32\,\alpha}\right)$ for some $C_1,C_2>0$ not related to the Algorithm \ref{alg:vanillathompson}, and $\Delta_{\max} = \max_{\mS} [r_{\bm\theta}^C(\mS^*) -  r_{\bm\theta}^C(\mS)]$.
\end{theorem}
\proof Appendix (Section \ref{sec:proofs})} 

Two of the most common problems studied in reinforcement learning with bandits are (1) regret minimization, and (b) best arm identification \citep{even2006action,bubeck2009pure}. 
  Variable selection could be loosely regarded as the `top $q^\star$-arms' identification problem when $q^\star$ is unknown.  While Thompson sampling is devised to minimize regret,  not necessarily to select the best arms (see e.g.  \cite{russo2016simple}), we nevertheless show that our sampling policy satisfies a version of variable selection consistency in the sense that the event $\{\mS_t=\mS^*\}$ occurs all but finitely many times as $t\rightarrow\infty$. This result is summarized in the following theorem.  
  
  \begin{theorem}
			\label{thm:almostsureconvergence}
				Under the Assumption 1 with $p$ fixed, the TVS sampling policy in Table \ref{alg:vanillathompson} with   $C=(\sqrt{5}-1)/2$  satisfies
				$\P\left(\lim\inf\limits_{t\rightarrow\infty}\{\mS_t=\mS^*\}\right)=1$.
		\end{theorem}

\proof Appendix (Section \ref{sec:proof_thm:almostsureconvergence})

\section{TVS  in Action} \label{sec:Implementation}
This section serves to illustrate Thompson Variable Selection  on benchmark simulated datasets and to document its performance. 
While various implementations are possible (by choosing different rewards $r(\mS_t,\mathcal D)$  in \eqref{eq:variable_screening}), 
we will focus on two specific choices that we broadly categorize into {\em offline} variants  for when $p>>n$ (Section \ref{sec:offline}) and  {\em streaming/online} variants for when $n>>p$ (Section \ref{sec:online}).

\begin{figure}[!t]

\begin{subfigure}{0.245\textwidth}\centering\includegraphics[width=4.2cm, height=5cm]{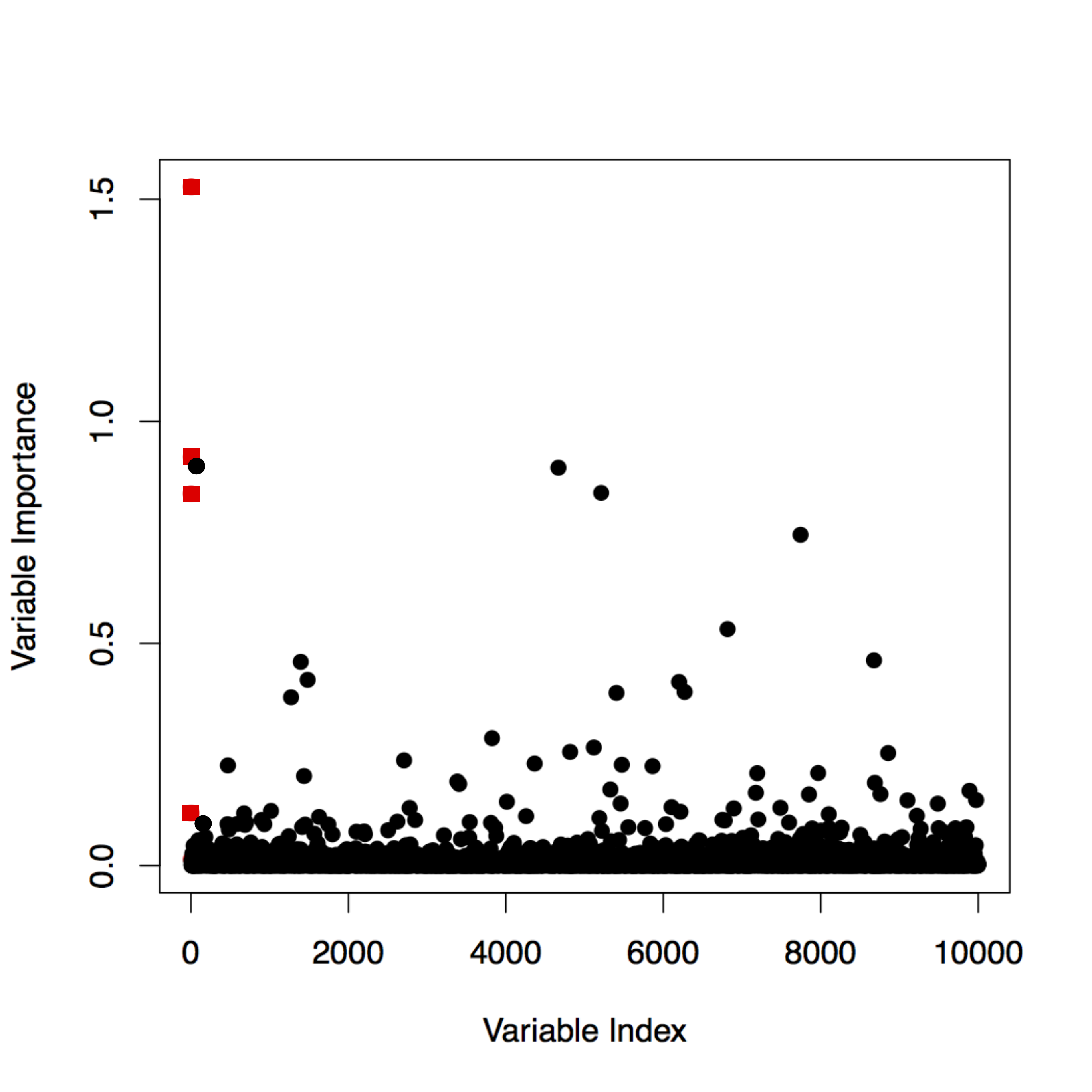}\caption{\scriptsize$D=50,M=5\,000$}\label{fig:vanillafriedman}\end{subfigure}
\begin{subfigure}{0.245\textwidth}\centering\includegraphics[width=4.2cm, height=5cm]{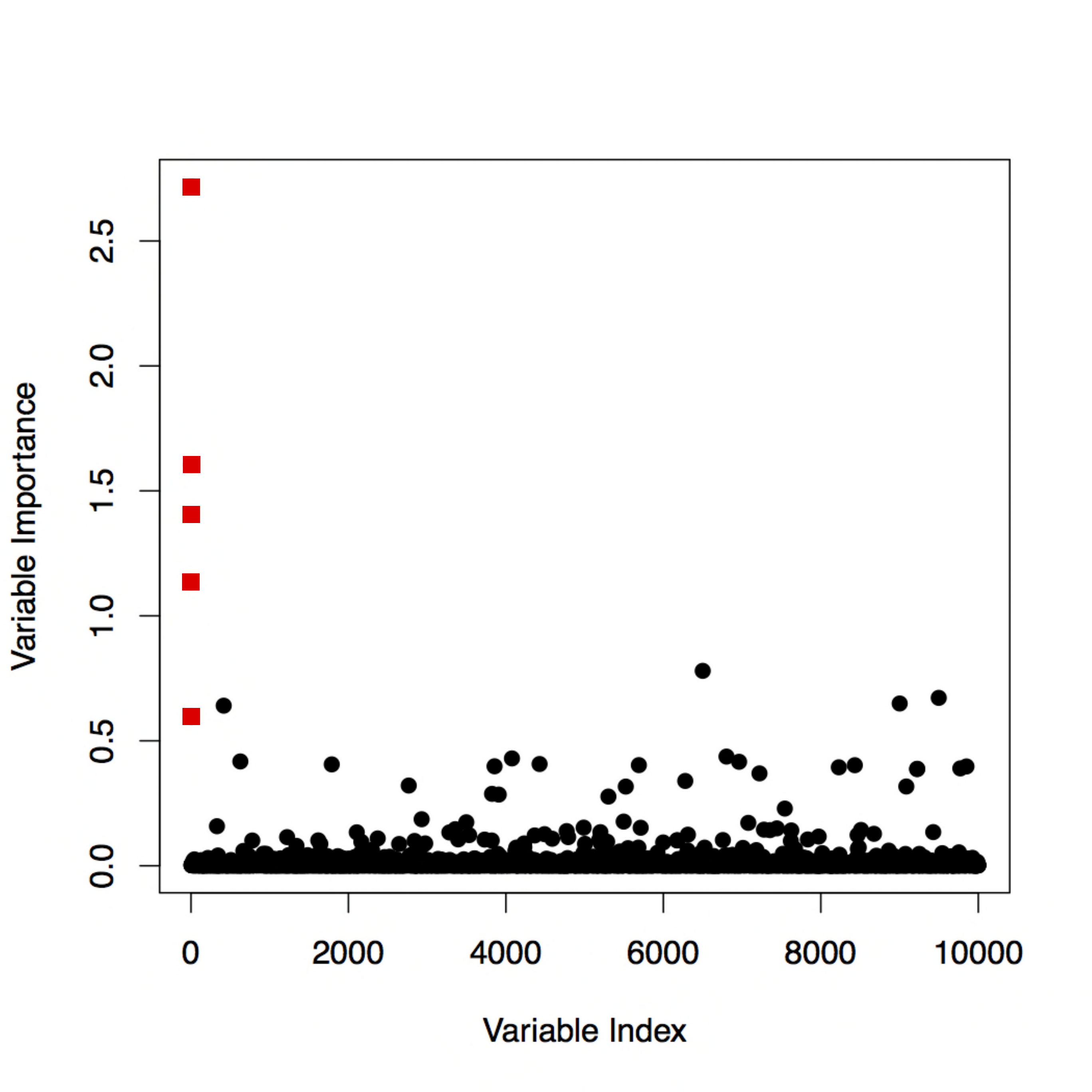}\caption{\scriptsize$D=50,M=50\,000$}\label{fig:vanillafriedman}\end{subfigure}
\begin{subfigure}{0.245\textwidth}\centering\includegraphics[width=4.2cm, height=5cm]{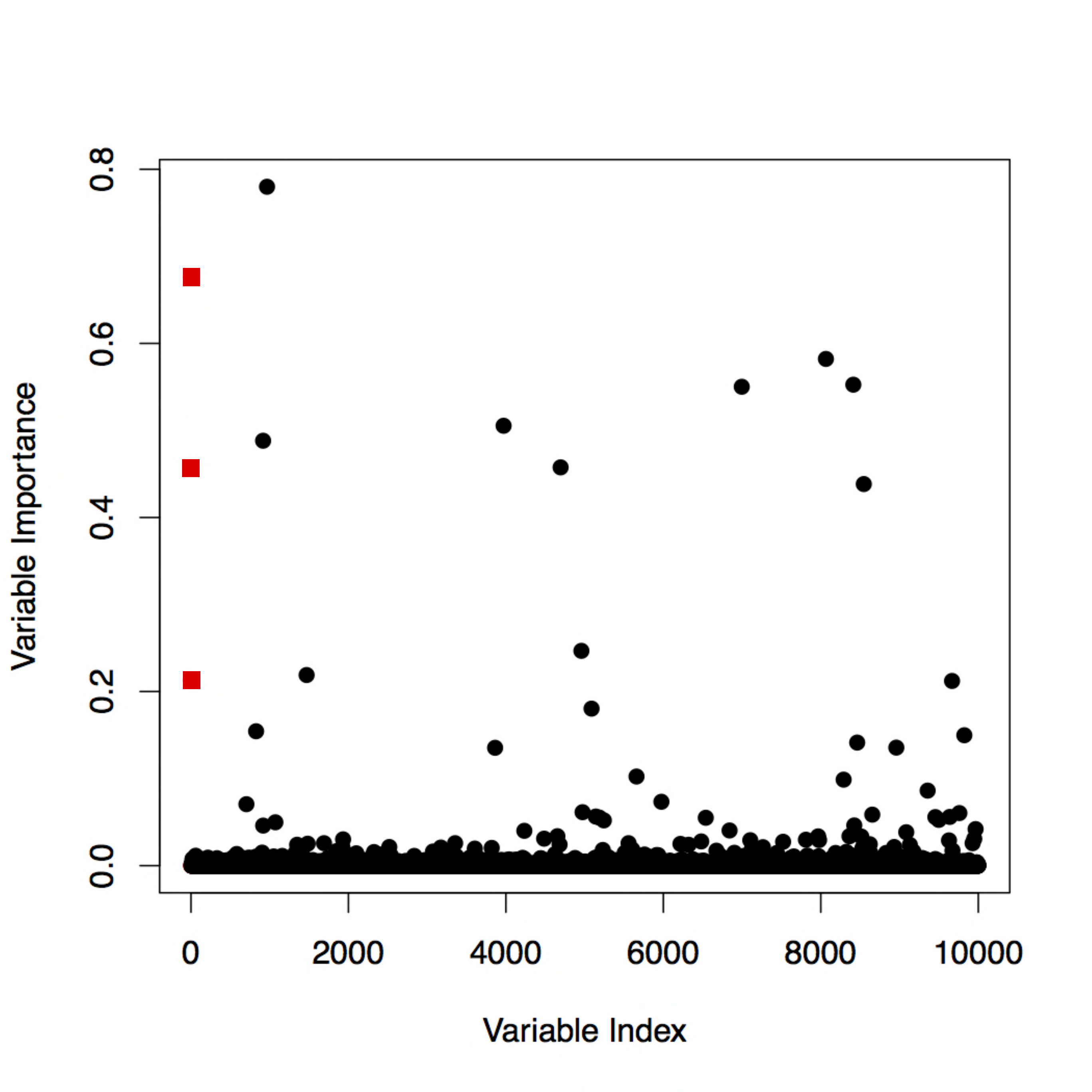}\caption{\scriptsize$D=10,M=5\,000$}\label{fig:vanillafriedman}\end{subfigure}
\begin{subfigure}{0.245\textwidth}\centering\includegraphics[width=4.2cm, height=5cm]{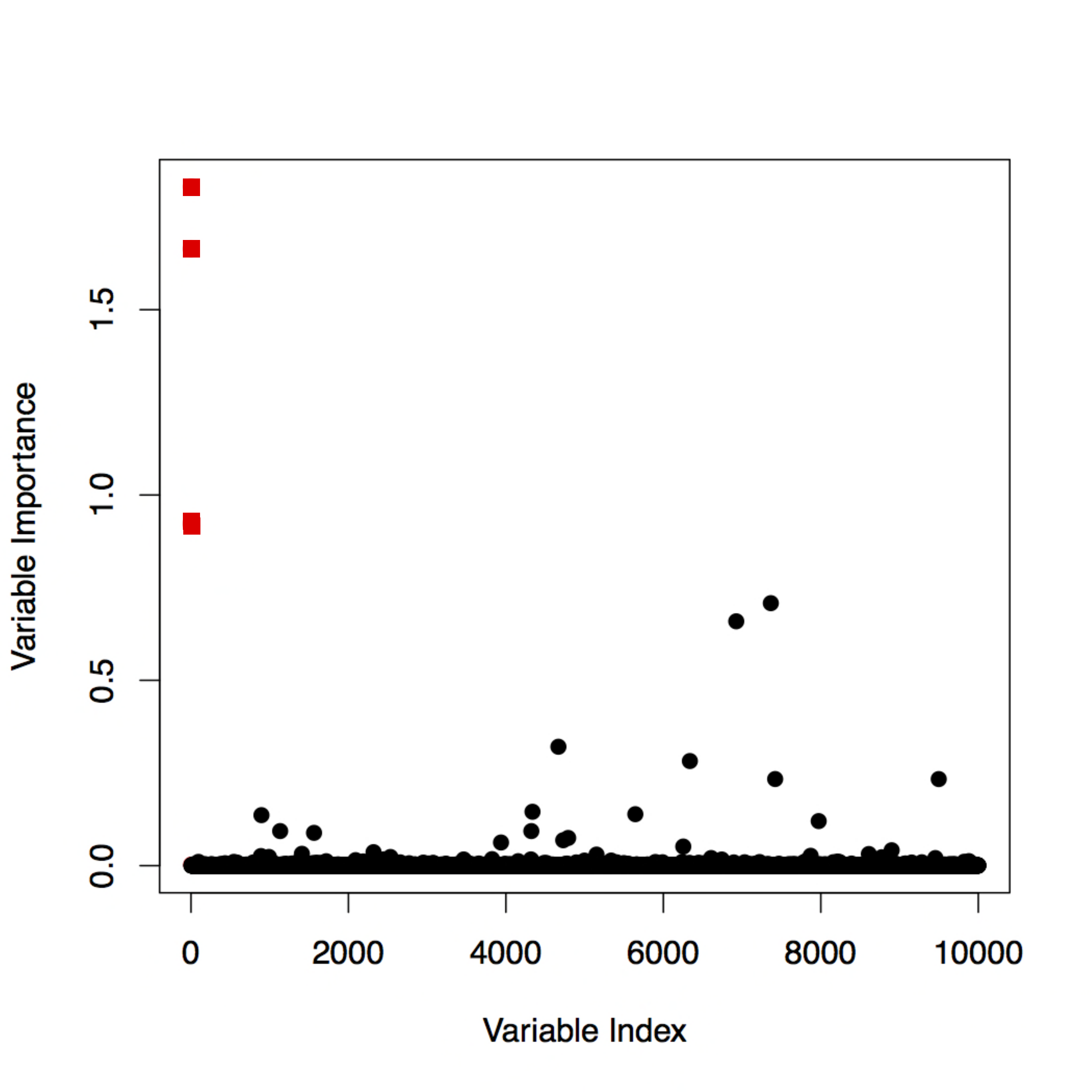}\caption{\scriptsize$D=10,M=50\,000$}\label{fig:vanillafriedman}\end{subfigure}
\caption{\small BART variable importance using  \texttt{sparse=TRUE} and various number of trees $D$ and MCMC iterations $M$. Red squares (the first five covariates) are signals and black dots are noise variables.}\label{fig:BART0}
\end{figure}

\subsection{Offline TVS} \label{sec:offline}
As a lead example in this section, we consider the benchmark Friedman data set (\cite{jerome_h._friedman_multivariate_1991})  with a vastly larger number of $p=10\,000$ predictors $ {\bm x}_i\in[0,1]^p $ obtained by iid sampling from $\texttt{Uniform}(0,1)$ and responses ${\bm Y} = (Y_1, \cdots Y_n)'$   obtained from  \eqref{eq:basic} with $\sigma^2 =1$ and
$$
f_0({\bm x}_i) = 10 \sin(\pi x_{i1}x_{i2}) + 20 (x_{i3} - 0.5)^2 + 10 x_{i4} + 5 x_{i5}\quad\text{for $i=1,\dots, 300$}.
$$
Due to the considerable number of covariates, feeding {\em all $10\,000$} predictors into a black box to obtain variable importance may not be computationally feasible and/or reliable. However, TVS can overcome this limitation by deploying subsets of predictors.
For instance,  we considered variable importance using the BART method (using the option \texttt{sparse=TRUE} for variable selection) with $D\in\{10,50\}$ trees and $M\in\{5\,000,50\,000\}$  MCMC iterations  are 
plotted them in Figure \ref{fig:BART0}. While increasing the number of iterations certainly helps in separating signal from noise,   it is not necessarily obvious where to set the cutoff for selection. 
One natural rule would be selecting those variables which have been used at least once  on average over the $M$  iterations. With $D=50$ and $M=50\,000$, this rule would identify  $4$ true signals, leaving out the quadratic signal variable $x_3$. The computation took around $8.5$ minutes.

The premise of  TVS is that one can  deploy a weaker learner (such as a forest with fewer trees) which generates a random reward that roughly captures signal and  is allowed to make mistakes.
With reinforcement learning, one hopes that each round will be wrong in a different way so that  mistakes will not be propagated over time. The expectation is that  (a) feeding  {\em only a small subset} $\mS_t$ in a black box  and (b) reinforcing positive outcomes, one obtains a more principled way of selecting variables and speeds up the computation. We illustrate the effectiveness of this mechanism below.

We use the offline local binary reward defined in \eqref{eq:reward_binary}.
We start with a non-informative prior $a_i(0)=b_i(0)=1$ for $1\leq i\leq p$ and choose $T=10$ trees in BART
so that variables are discouraged from entering the model too wildly.   This is a {\em weak learner} which does not seem to do perfectly well for variable selection even after very many MCMC iterations (see Figure \ref{fig:BART0}(d)). 
We use the TVS  implementation in Table \ref{alg:vanillathompson} with a dramatically smaller number $M\in\{100,500,1\,000\}$ of  MCMC  burn-in iterations  for BART inside TVS.   We will see below that large $M$ is {\em not needed} for TVS to unravel signal  even with as few as $10$ trees.

{\centering
\begin{figure}[t]
\begin{subfigure}{0.32\textwidth}\centering\includegraphics[width=5.5cm,height=6cm]{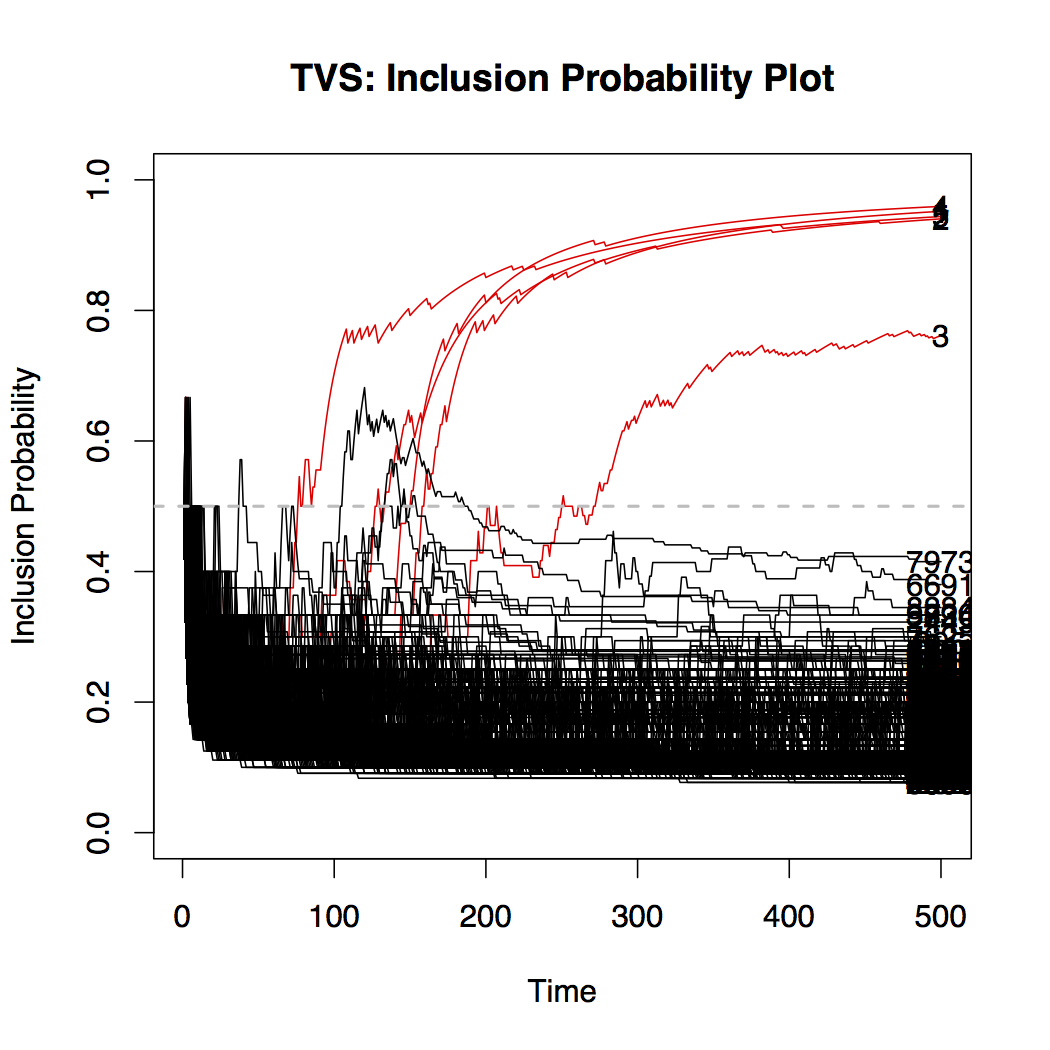}\caption{$M=100$}\label{fig1a}
\end{subfigure}
\begin{subfigure}{0.32\textwidth}\centering\includegraphics[width=5.5cm,height=6cm]{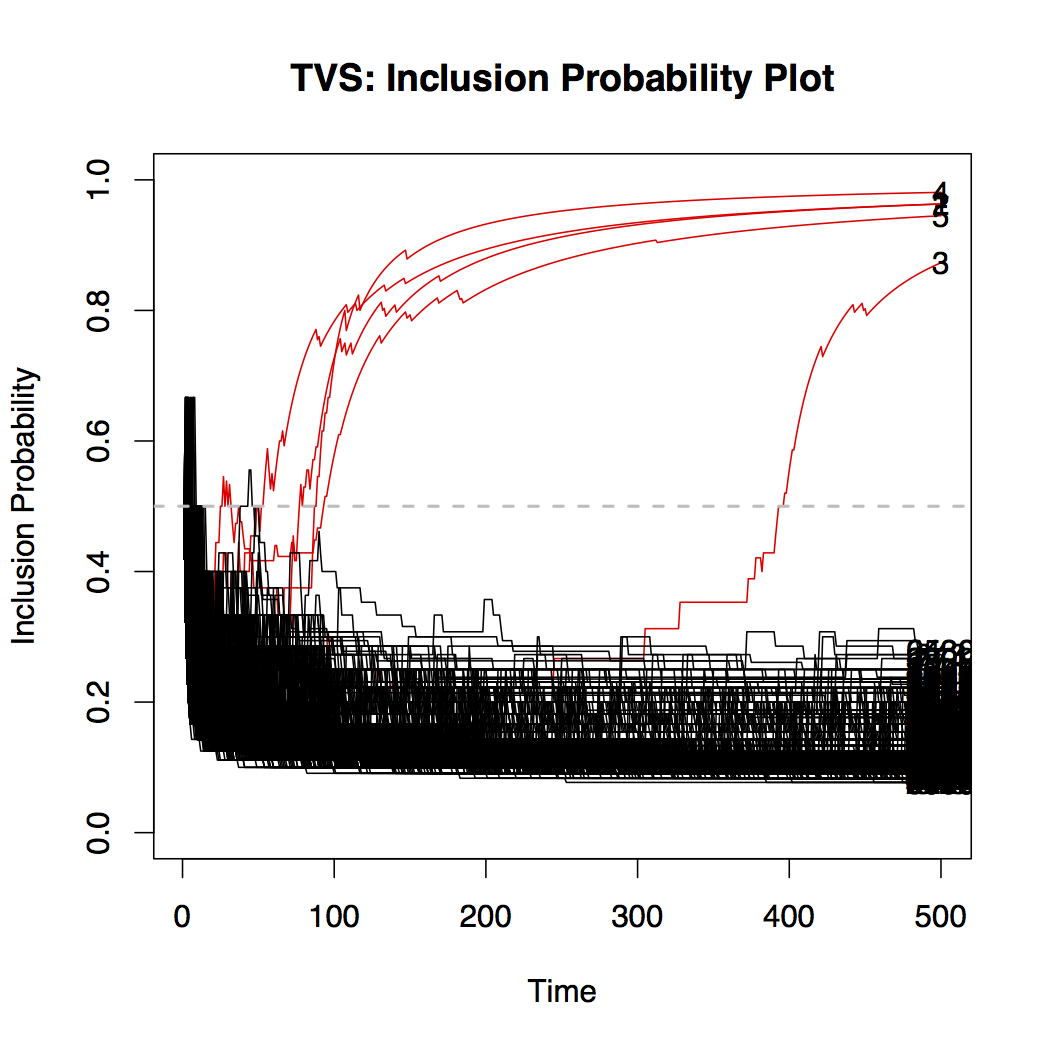}\caption{$M=500$}\label{fig1b}
\end{subfigure}
\begin{subfigure}{0.32\textwidth}\centering\includegraphics[width=5.5cm,height=6cm]{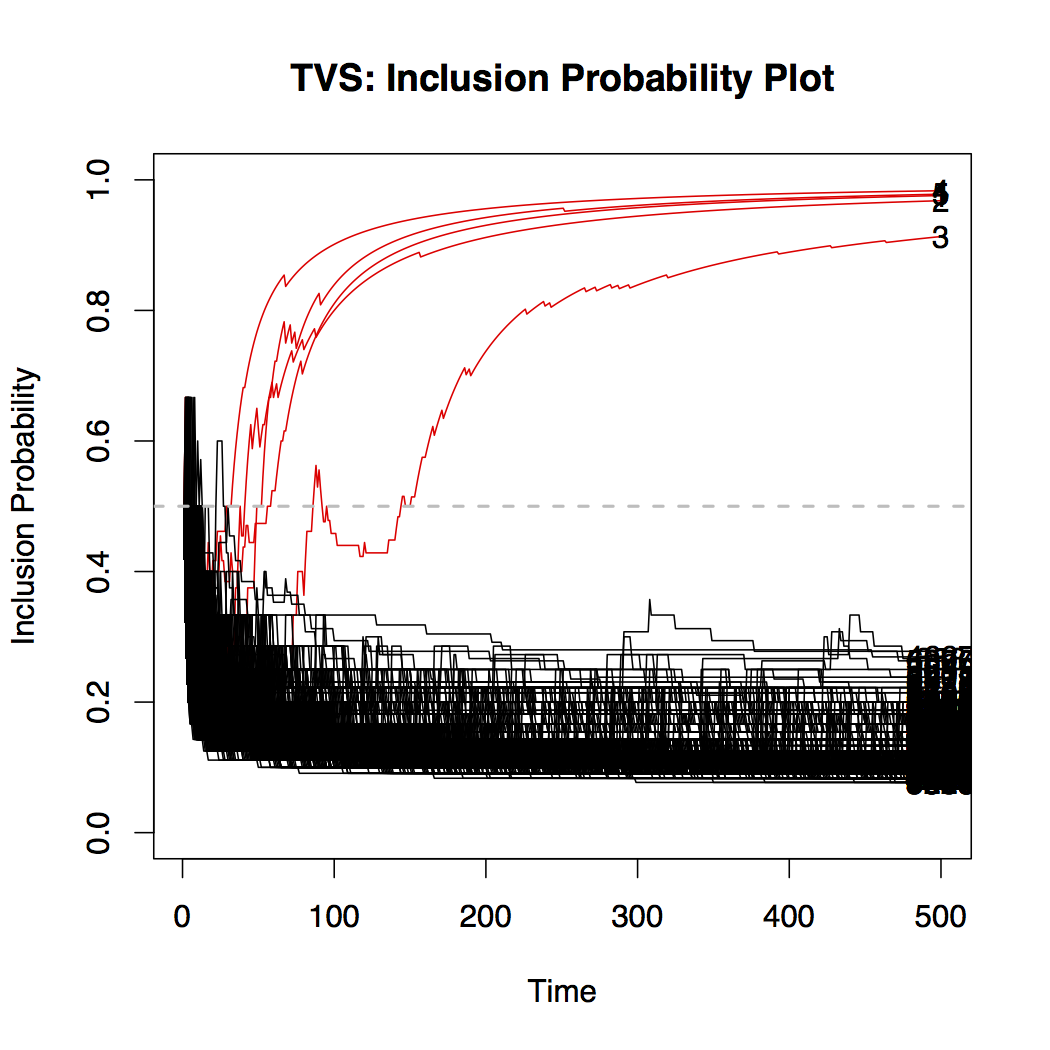}\caption{$M=1\,000$}\label{fig1c}
\end{subfigure}
\caption{{\small The evolution of inclusion probabilities \eqref{eq:pi} over ``time" $t$ for  the Friedman data set. The plot depicts posterior inclusion probabilities $\pi_i(t)$ in \eqref{eq:pi}  over time (number of TVS iterations).
 Red lines indicate the $5$ signal variables  and black lines indicate the noise variables.}}
\label{tab:outcomes}
\end{figure}
}

TVS results are summarized in  Figure \ref{tab:outcomes}, which depicts  `posterior inclusion probabilities' $\pi_i(t)$ defined in \eqref{eq:pi}
over time $t$ (the number of plays), one line for each of the $p=10\,000$ variables. 
To better appreciate  informativeness of  $\pi_i(t)$'s,  true variables $x_1,\dots, x_5$ are depicted in red while the noise variables are black. 
Figure  \ref{tab:outcomes} shows a very successful demonstration for several reasons. The first panel (Figure \ref{fig1a}) shows a very weak learner (as was seen from Figure \ref{fig:BART0}) obtained by sampling rewards only after $M=100$ burnin iterations. 
Despite the fact that learning at each step is weak, it took only around $T=300$  iterations (obtained in less than $40$ seconds!) for  the 
$\pi_i(t)$ trajectories of the $5$ signals to cross the $0.5$ decision boundary. After $T=300$ iterations, the noise covariates are safely suppressed below the decision boundary and the trajectories $\pi_i(t)$ stabilize towards the end of the plot. Using more MCMC iterations $M$, fewer TVS iterations are needed to obtain a cleaner separation between signal and noise (noise $\pi_t$'s are closer to zero while signal $\pi_t$'s are closer to one). 
With enough internal MCMC iterations ($M=1\,000$ in the right panel),   TVS is able to effectively separate signal from noise in around $200$ iterations (obtained in less than $2$ minutes). We have not seen such conclusive separation with plain BART  (Figure \ref{fig:BART0})  even after very many MCMC iterations which took considerably longer.   Note that each iteration of TVS uses {\em only a small subset of predictors} and much fewer iterations $M$  and TVS is thereby destined to be faster than  BART on the entire dataset (compare $8.5$ minutes for $20\,000$ iterations with $40$ seconds in Figure \ref{fig1a}). Applying the more traditional variable selection techniques was also not as successful. For example,  the Spike-and-Slab LASSO  (SSL) method ($\lambda_1=0.1$ and $\lambda_0\in\{0.1+k\times 10:k=1,\dots, 10\}$)  which relies on a linear model  missed the quadratic predictor but identified all $4$ remaining signals with no false positives.

\subsection{Online TVS}\label{sec:online}
As the second TVS example,  we focus on the case with many more observations than covariates, i.e. $n>>p$. As we already pointed out in Section \ref{sec:local_reward}, we assume that the dataset $\mathcal D_n=\{(Y_i,\bm x_i):1\leq i\leq n\}$ has been  partitioned into  mini-batches $\mathcal D_t$ of size $s=n/T$.
We deploy our online TVS method (Table \ref{alg:vanillathompson} with C2$^\star$) to sequentially screen each batch and transmit the posterior information onto the next mini-batch through a prior. This should be distinguished from streaming variable selection, where new features arrive at different time points (\cite{foster2008alpha}).
Using the notation $r_t\equiv r(\mS_t,\mathcal D_t)$ in \eqref{eq:variable_screening} with $\mathcal D_t=\{(Y_i,\bm x_i):(t-1)s+1\leq i\leq ts\}$ and having processed $t-1$ mini-batches, one can treat the beta posterior as a new prior for the incoming data points, where 
$$
\pi(\bm\theta\,\C\, r_1,\dots, r_t)\propto \pi(\bm\theta\,\C\,  r_1,\dots, r_{t-1})\prod_{i\in\mS_t}\theta_i^{\gamma_{i}^t}(1-\theta_i)^{1-\gamma_{i}^t}.
$$
Parsing the observations in batches  will be particularly beneficial when processing the entire dataset (with overwhelmingly many rows) is not feasible for the learning algorithm. TVS leverages the fact that applying a machine learning method $T$ times using only a subset of $s$ observations and a subset $\mS_t$ of variables is a lot faster than processing the entire data. While the posterior distribution\footnote{Treating the rewards as data.} of $\theta_i$'s after one pass through the entire dataset will have seen all the data $\mathcal D_n$, $\theta_i$'s can be interpreted as the frequentist probability that the screening rule picks a variable $x_i$ having access to only $s$ measurements.

\begin{figure}[!t]
\begin{subfigure}{0.245\textwidth}\centering\includegraphics[width=4.2cm, height=6cm]{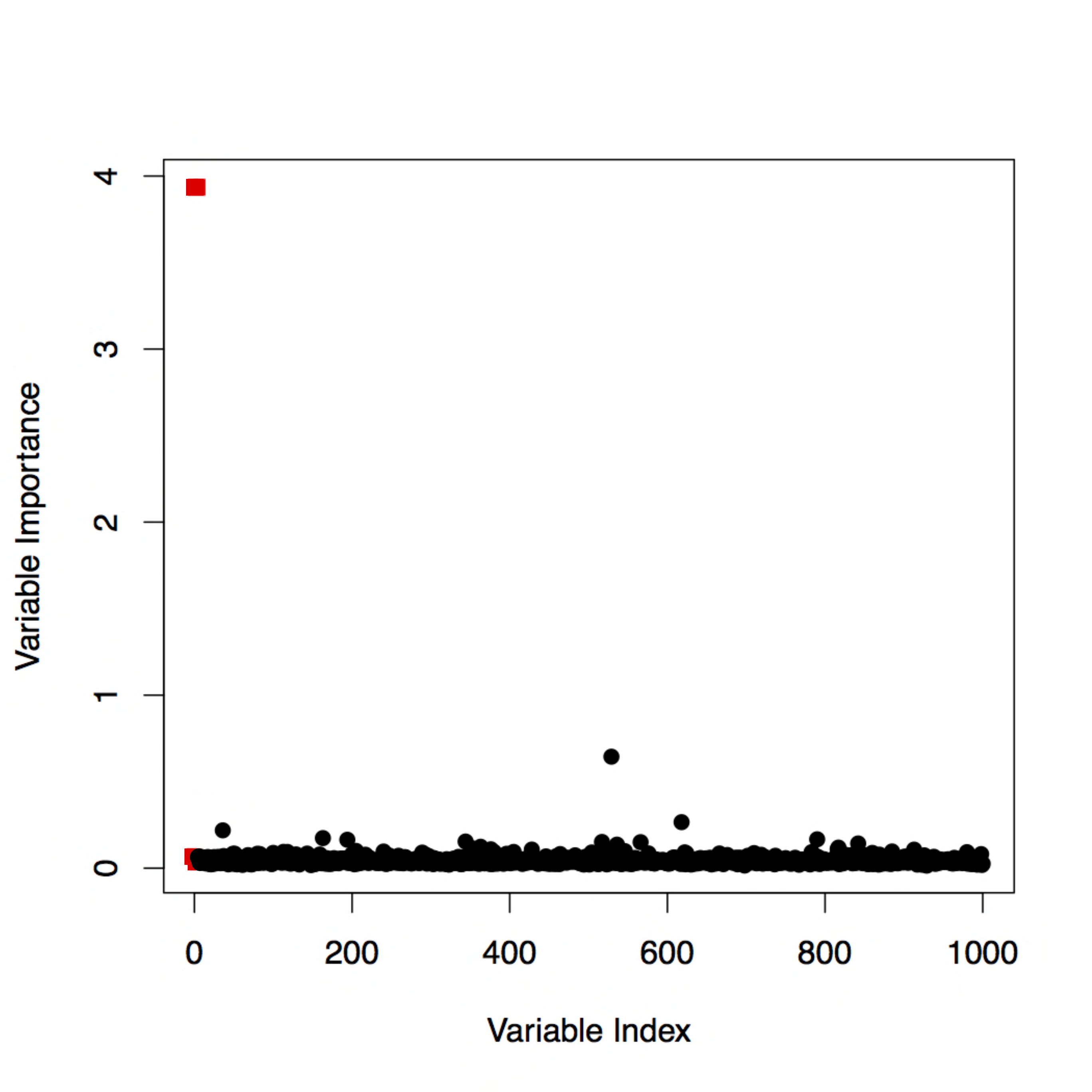}\caption{$s=100$}\label{fig:2a}\end{subfigure}
\begin{subfigure}{0.245\textwidth}\centering\includegraphics[width=4.2cm, height=6cm]{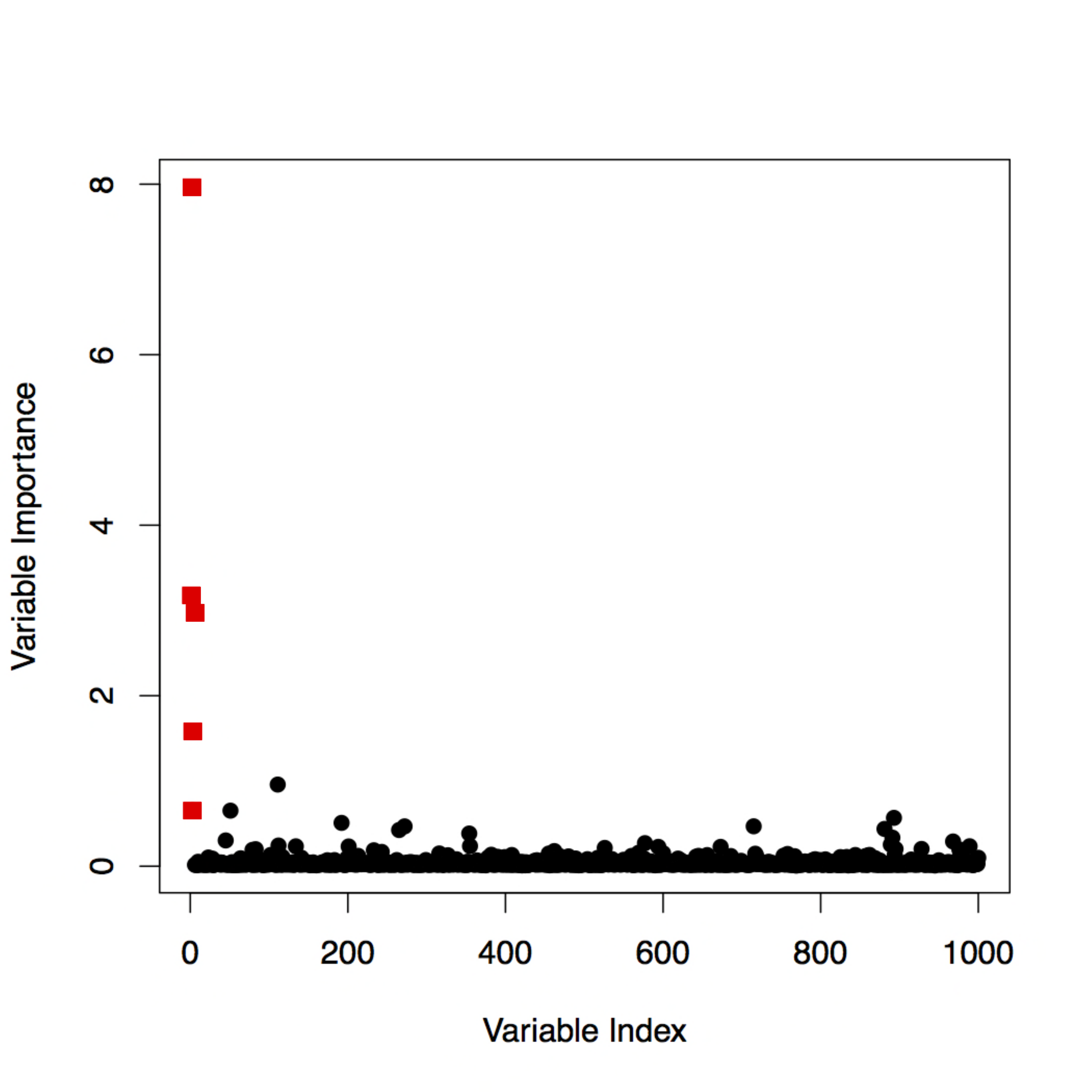}\caption{$s=500$}\label{fig:2b}\end{subfigure}
\begin{subfigure}{0.245\textwidth}\centering\includegraphics[width=4.2cm, height=6cm]{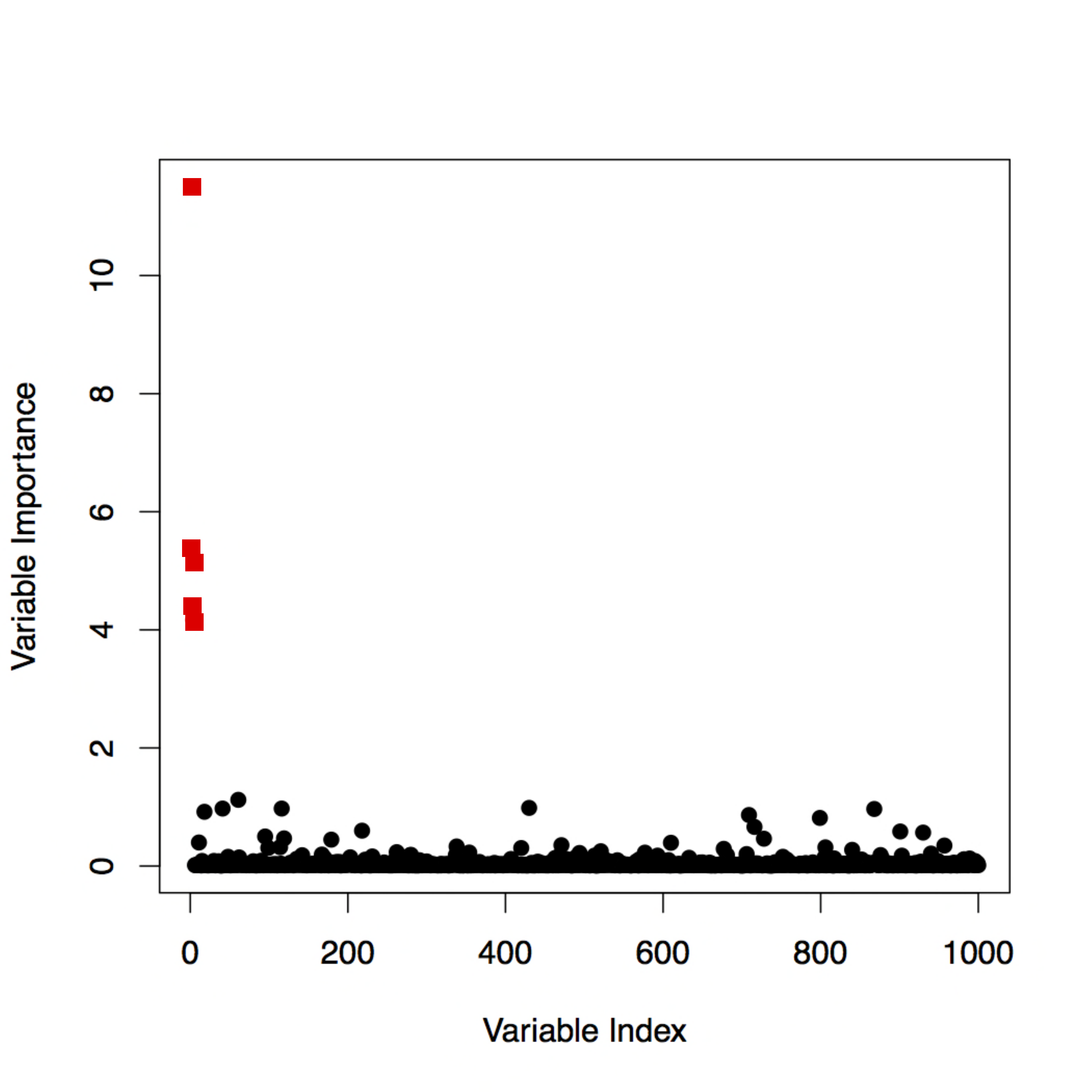}\caption{$s=1\,000$}\label{fig:2c}\end{subfigure}
\begin{subfigure}{0.245\textwidth}\centering\includegraphics[width=4.2cm, height=6cm]{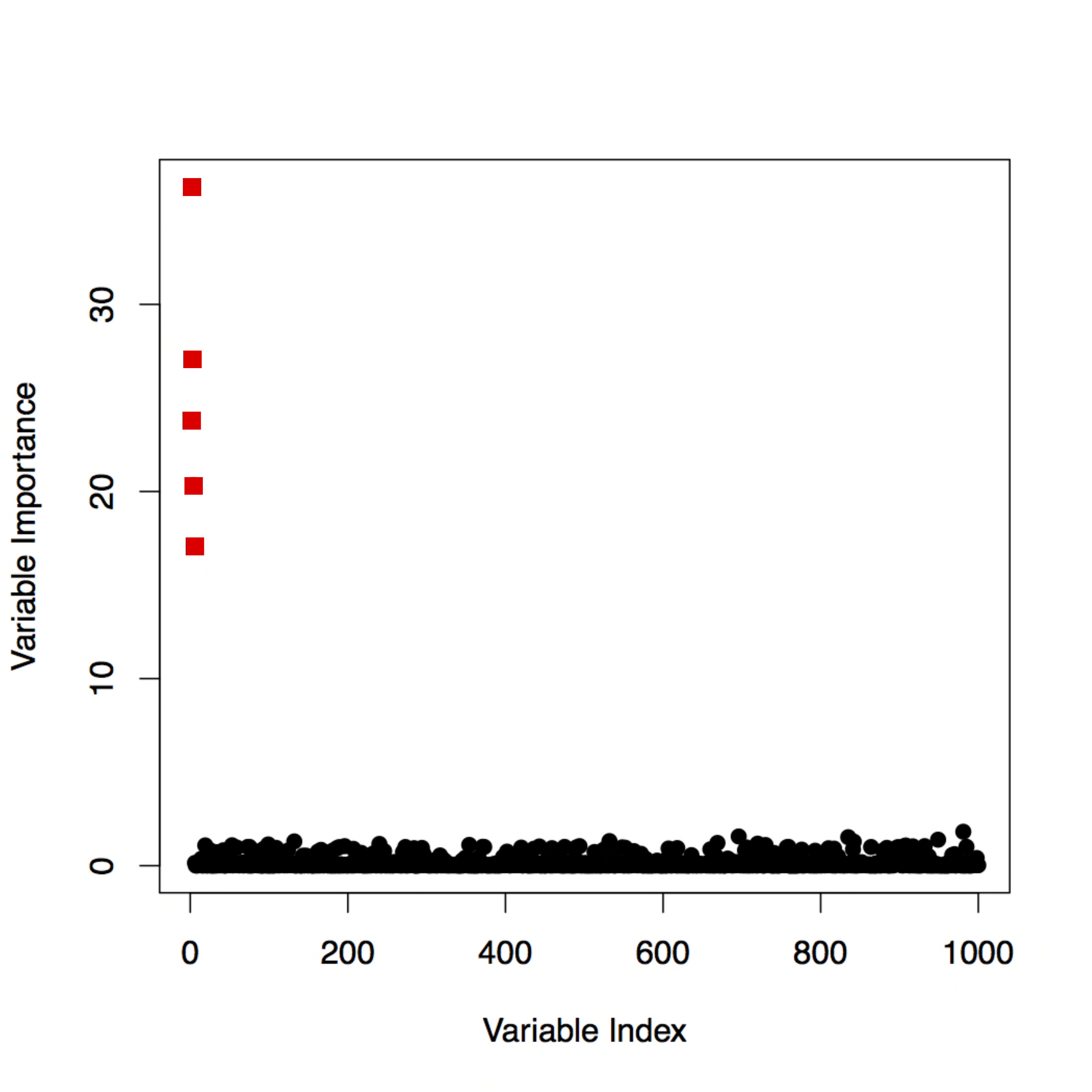}\caption{Entire Data}\label{fig:2d}\end{subfigure}
\caption{\small BART variable importance using $T=20\,000$ (using $50$ trees and \texttt{sparse=TRUE}) and various data subsets. Red squares (the first five covariates) are signals and black dots are noise variables.}\label{fig:BART1}
\end{figure}

We illustrate this sequential learning method on a challenging simulated example from \cite[Section 5.1]{liang2018bayesian} .
We assume that the explanatory variables $ {\bf x}_i\in[0,1]^p $ have been obtained from $ x_{ij}=(e_i + z_{ij})/2$ for $1\leq i\leq n$ and $1\leq j\leq p$, where $e,z_{ij}\iid\mathcal{N}(0,1)$.
This creates  a sample correlation of about $0.5$ between all variables. The responses ${\bf Y} = (Y_1, \cdots, Y_n)'$ are then obtained from  \eqref{eq:basic}, where
$$
f_0({\bf x}_i) = \frac{10x_{i2}}{1+x_{i1}^2}+ 5\sin(x_{i3}x_{i4}) +2x_{i5}
$$
with $\sigma^2=0.5$. This is a challenging scenario due to (a) the non-negligible correlation between signal and noise, and (b) the non-linear contributions of $x_{1}-x_4$. 
Unlike Liang et al. (2008) who set $n=p=500$, we make the problem considerably more difficult by choosing $n=20\,000$ and  $p=1\,000$.
We would expect a linear model selection method to miss these two nonlinear signals. Indeed, the Spike-and-Slab LASSO method   (using $\lambda_1=0.1$ and $\lambda_0\in\{0.1+k\times 10:k=1,\dots, 10\}$ only identifies variables   $x_1,x_2$ and $x_5$.
Next, we deploy the BART method with variable selection (Linero 2016) by setting   \texttt{sparse=TRUE} (\cite{linero_bayesian_2017}) and $50$ trees\footnote{Results with $10$ were not nearly as satisfactory.} in the BART software (\cite{chipman_bart:_2010}).
The choice of $50$ trees for variable selection was recommended in \cite{bleich} and was seen to work very well. 
 Due to the large size of the dataset, it might be reasonable to first inquire about variable selection from smaller fractions of data. We consider random subsets of different sizes $s\in\{100,500,1\,000\}$ as well as the entire dataset and we run BART for  $M=20\,000$ iterations.
 Figure  \ref{fig:BART1} depicts BART  importance measures (average number of times each variable was used in the forest). We have seen BART separating the signal from noise rather well on batches of size $s\geq 1\,000$ and with MCMC iterations $M\geq 10\, 000$. 
The scale of the importance measure depends on $s$ and it is not necessarily obvious where to make the cut for selection. A natural (but perhaps ad hoc) criterion would be to pick variables which were on average used at least once.
This would produce false negatives for smaller $s$ and many false positives ($29$ in this example) for $s=20\,000$. The Hamming distance between the true and estimated model as well as the computing times are reported in Table \ref{table:time_BART}.  This illustrates how selection based on the importance measure is difficult to automate. While visually inspecting the importance measure for $M=20\,000$  and $s=20\,000$ (the entire dataset) in Figure \ref{fig:2d} is very instructive for selection, it took more than $70$ minutes on this dataset. To enhance the scalability, we deploy our reinforcement learning TVS method for streaming batches of data.

\begin{table}
\begin{center}
\scalebox{0.7}{
\begin{tabular}{c | c c| c c| c c| c c| c c| c c| c c}
\hline\hline
		  &\multicolumn{2}{c}{ $s=100$} &\multicolumn{2}{c}{ $s=200$} &\multicolumn{2}{c}{ $s=500$} &\multicolumn{2}{c}{ $s=1\,000$} 
		  &\multicolumn{2}{c}{ $s=5\,000$} &\multicolumn{2}{c}{ $s=10\,000$} &\multicolumn{2}{c}{ $s=20\,000$} \\
\hline
T& Time & HAM & Time & HAM& Time & HAM& Time & HAM& Time & HAM& Time & HAM& Time & HAM\\
\hline
$5\,000$   & 6.7 &4&7.7&5&11.4&3&21.5&2&103.1&3&264.2&8&794.1 &16  \\ 
$10\,000$ & 16.2&4&16&4&23.6&2&37.8&3&213&8&549.9&18&2368.4  &25   \\
$20\,000$ &  27.3&4&31.1&4&47.7&1&74.7&1&418.4&10&1090.6&21&4207.4&29\\
\hline\hline
 \end{tabular}}
 \end{center}
 \caption{\small Computing times (in seconds) and Hamming distance of BART on subsets of observations using all  $p$ covariates.
 Hamming distance compares the true model with a model obtained by truncating the BART importance measure at 1.}\label{table:time_BART}
\end{table}

\begin{figure}[!t]
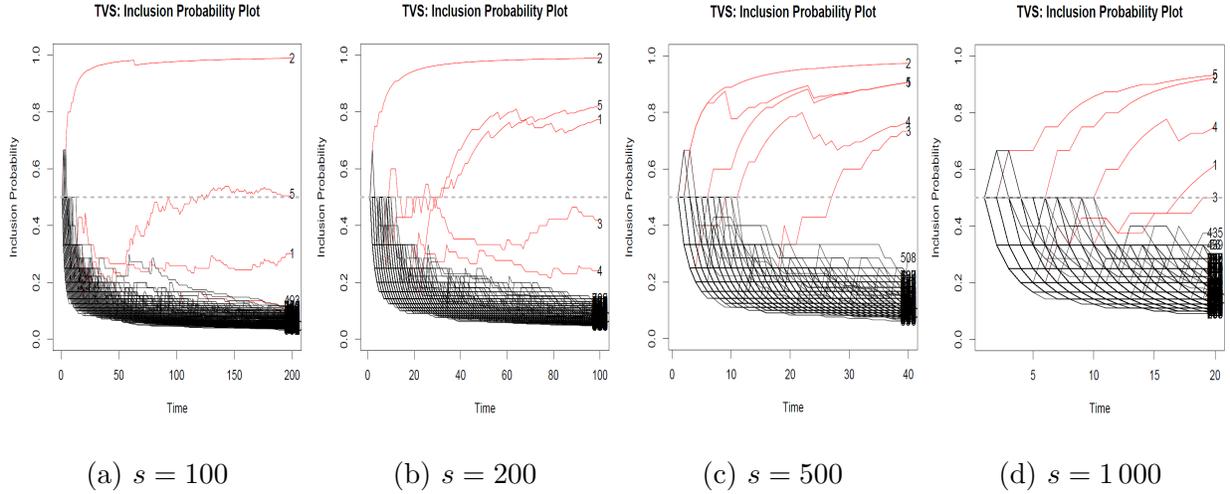

\begin{subfigure}{0.245\textwidth}\centering\includegraphics[width=4.2cm, height=6cm]{Pics/plot_10000_100_1}\caption{$s=100$}\label{fig:vanillafriedman}\end{subfigure}
\begin{subfigure}{0.245\textwidth}\centering\includegraphics[width=4.2cm, height=6cm]{Pics/plot_10000_200_1}\caption{$s=200$}\label{fig:vanillafriedman}\end{subfigure}
\begin{subfigure}{0.245\textwidth}\centering\includegraphics[width=4.2cm, height=6cm]{Pics/plot_10000_500_1}\caption{$s=500$}\label{fig:vanillafriedman}\end{subfigure}
\begin{subfigure}{0.245\textwidth}\centering\includegraphics[width=4.2cm, height=6cm]{Pics/plot_10000_1000_1}\caption{$s=1\,000$}\label{fig:vanillafriedman}\end{subfigure}
\caption{\small TVS inclusion probabilities $T=10\,000$ (using $10$ trees and \texttt{sparse=TRUE}) and various batch sizes after single pass through the data.}\label{fig:TVS1}
\end{figure}

Using the online local reward \eqref{eq:reward_binary2} with  BART (with $10$ trees and \texttt{sparse=TRUE} and $M$ iterations) on batches of data $\mathcal D_n$ of size $s$. 
This is a weaker learning rule than the one considered   in Figure \ref{fig:BART1} ($50$ trees). 
Choosing   $s=100$ and  $M=10\,000$, BART may not be able to obtain perfect signal isolation on a single data batch (see Figure \ref{fig:BART1}(a) which identifies only one signal variable).
However, by propagating information from one batch onto the next,  TVS  is able to tease out more signal (Figure \ref{fig:TVS1}(a)).
Comparing Figure \ref{fig:BART1}(b)) and Figure \ref{fig:TVS1}(c) is even more interesting, where TVS inclusion probabilities for all signals eventually cross the decision boundary after merely $40$ TVS iterations. There is ultimately a tradeoff between the batch size $s$ and the number of iterations needed for the TVS to stabilize. For example, with $s=1\,000$ one obtains a far stronger learner (Figure \ref{fig:BART1}(c)), but the separation may not be as clear after only $T=n/s=20$ TVS iterations (Figure \ref{fig:TVS1}(d)). One can increase the number of TVS iterations by performing multiple passes through the data after bootstrapping the entire dataset and chopping it into new batches which are a proxy for   future data streams. Plots of TVS inclusion probabilities after $5$ such passes through the data are in Figure \ref{fig:TVS2}. Curiously,  one obtains much better separation even for $s=200$ and with larger batches ($s=500$) the signal is perfectly separated. Note that TVS is a random algorithm and thereby the trajectories in Figure \ref{fig:TVS2}  at the beginning are slightly different from Figure \ref{fig:TVS1}. Despite the random nature, however, we have seen the separation apparent from Figure \ref{fig:TVS2} occur consistently across multiple runs of the method.

Several observations can be made from the timing and performance comparisons presented in Table \ref{table:time_TVS}.  When the batch size is not large enough, repeated runs will not help.  The Hamming distance in all cases only consists of false negatives and can be decreased by increasing the batch size or increasing the number of iterations and rounds. Computationally, it seems beneficial to increase the batch size $s$ and supply enough MCMC iterations. Variable selection accuracy can also be increased with multiple rounds.

\begin{figure}[!t]
\begin{subfigure}{0.245\textwidth}\centering\includegraphics[width=4.2cm, height=6cm]{Pics/plot_10000_100_5}\caption{$s=100$}\label{fig:vanillafriedman}\end{subfigure}
\begin{subfigure}{0.245\textwidth}\centering\includegraphics[width=4.2cm, height=6cm]{Pics/plot_10000_200_5}\caption{$s=200$}\label{fig:vanillafriedman}\end{subfigure}
\begin{subfigure}{0.245\textwidth}\centering\includegraphics[width=4.2cm, height=6cm]{Pics/plot_10000_500_5}\caption{$s=500$}\label{fig:vanillafriedman}\end{subfigure}
\begin{subfigure}{0.245\textwidth}\centering\includegraphics[width=4.2cm, height=6cm]{Pics/plot_10000_1000_5}\caption{$s=1\,000$}\label{fig:vanillafriedman}\end{subfigure}
\caption{\small TVS inclusion probabilities $T=10\,000$ (using $10$ trees and \texttt{sparse=TRUE}) and various batch sizes after  $5$ passes through the data.}\label{fig:TVS2}
\end{figure}

\begin{table}
\begin{center}
\scalebox{0.6}{
\begin{tabular}{|c | c c |c c|  c c| c c|}
\hline\hline
		  &\multicolumn{2}{c}{ $s=100$} &\multicolumn{2}{c}{ $s=200$} &\multicolumn{2}{c}{ $s=500$} &\multicolumn{2}{c}{ $s=1\,000$} \\
\hline
 & Time & HAM & Time & HAM  & Time & HAM & Time & HAM \\
\hline
\multicolumn{9}{c}{$T=500$ }\\
\hline
\quad\sl\small $1$ round	                                    & 58.9&4&34.1&3&19.3&3&15.7&4\\
\quad\sl\small $5$ rounds		   &251.5&4  &165&3 &91.2 &3&68.7&2  \\ 
\quad\sl\small $10$ rounds		   &467.5&4&348.8&3&187.8&3&137.2&1\\
\hline
\multicolumn{9}{c}{$T=1\,000$ }\\
\hline
\quad\sl\small $1$ round	  				   & 84&4&49.8&3&29.5&3&23.6& 2  \\
\quad\sl\small $5$ rounds		   &411.8&4&241.5&3&140.8&1&111.4&0\\
\quad\sl\small $10$ rounds	   &870.6&3&507&2&288.2&1&224.2&0\\
\hline
\multicolumn{9}{c}{$T=1\,0000$ }\\
\hline
\quad\sl\small $1$ round					  &  541.8&3&330.9&2&220.4&0&182.8&0\\
\quad\sl\small $5$ rounds		    &2421.2&3&1501.9&2&1060.2&0&972.2&0\\
\quad\sl\small $10$ rounds		    &4841.9&3&3027.9&0&2248.3&0&2087.6&0\\
\hline\hline
 \end{tabular}}
 \end{center}
 \caption{\small Computing times (in seconds) and Hamming distance for TVS using  different batch sizes $s$ and BART iterations $T$ and multiple passes through the data.
 The Hamming distance compares the true model with a model truncating the last TVS inclusion probability at 0.5. 
 }\label{table:time_TVS}
\end{table}

\section{Simulation Study}
\label{sec:simu}
We further evaluate the performance of TVS in a more comprehensive simulation study. We compare TVS with several related non-parametric variable selection methods and with classical parametric ones. We assess these methods based on the following performance criteria: False Discovery Proportion (FDP) (i.e. the proportion of discoveries that are false), Power  (i.e. the proportion of true signals discovered as such), Hamming Distance (between the true and estimated model) and Time.

	\subsection{Offline Cases}	
	\label{sec:simu_offline}
 For a more comprehensive performance evaluation, we consider the following $4$ mean functions $f_0(\cdot)$  to generate outcomes using  \eqref{eq:basic}. 
 	For each setup, we summarize results over $50$ datasets of a dimensionality $p\in\{1\,000, 10\,000\}$ and a sample size $n=300$.
	\begin{itemize}
	\item {\bf Linear Setup}: The regressors ${\bm x}_i$ are drawn  independently from $N(0,\Sigma)$, where $\Sigma = (\sigma_{jk})_{j,k=1,1}^{p,p}$ with $\sigma_{jj} = 1$ and $\sigma_{jk} = 0.9^{\origbar j-k \origbar}$ for $j \neq k$. Only the first $5$ variables are related to the outcome (which is generated from \eqref{eq:basic} with $\sigma^2 = 5$) via the mean function $f_0(\bm  x_i) = x_{i1} + 2x_{i2} + 3x_{i3} -2x_{i4} -x_{i5}$. 
	 \item {\bf Friedman Setup}: The Friedman setup was  described earlier in Section \ref{sec:Vanilla}.  In addition, we now introduce correlations of roughly $0.3$ between the explanatory variables.  
	 
	\item {\bf Forest Setup}: We generate ${\bm x_i}$  from  $N(0,\Sigma)$, where $\Sigma = (\sigma_{jk})_{j,k=(11)}^{p,p}$ with $\sigma_{jj} = 1$ and $\sigma_{jk} = 0.3^{\origbar j-k \origbar} $ for $j \neq k$. We then draw the mean function $f_0(\cdot)$ from a BART prior with $200$ trees, using only first $5$ covariates for splits. The outcome is generated from   \eqref{eq:basic} with $\sigma^2 =0.5$.
	\item {\bf Liang et al (2016) Setup}: This setup was described earlier in Section \ref{sec:online}. We now use $\sigma^2 =0.5$.
\end{itemize}

We run TVS with $M=500$ and $M=1\,000$ internal BART MCMC iterations and with $T=500$ TVS iterations. 
As two benchmarks for comparison, we consider the original BART method (in the R-package \texttt{BART}) and a newer variant called DART (\cite{linero_bayesian_2017}) which is tailored to high-dimensional data and which can be obtained in \texttt{BART} by setting \texttt{sparse=TRUE} (\texttt{ a=1, b=1}). 
We ran BART and DART  for $M=50\,000$  MCMC iterations using the default prior settings  with  $D=20$, $D=50$ and $D=200$ trees for BART  and $D=10$, $D=50$ and $D=200$ trees for DART. 
We considered two variable selection criteria: (1)  posterior inclusion probability (calculated  as the proportion of sampled forests that split on a given variable) at least $0.5$ (see Linero (2018) and \cite{bleich} for more discussion on variable selection using BART), (2)  the average number of splits in the forest (where the average is taken over the $M$ iterations) at least $1$.  We report the settings with the best performance, i.e. BART with $D =20$ trees and DART with $D=50$ trees using the second inclusion criterion. The third benchmark method we use for comparisons is the  Spike-and-Slab LASSO  (\cite{rockova_spike-and-slab_2018}) implemented in the R-package \texttt{SSLASSO} with \texttt{lambda1=0.1}  and the spike penalty ranging from $\lambda_1$ to the number of variables $p$ (i.e. \texttt{lambda0 = seq(1, p, length=p)}). We choose the same set of variable chosen by \texttt{SSLASSO} function after the regularization path has stabilized using the \texttt{model} output.
 We have also implemented  Sure Independence Screen (SIS) (\cite{fan2008sure}) as a  variable filter before applying BART variable selection\footnote{SIS on its own yielded too many false positives.}. Next, we applied one of the benchmark Bayesian variable selection methods, the horseshoe prior (\cite{van2017uncertainty,carvalho2010horseshoe}) implementation  from the \texttt{R-package} \texttt{horseshoe}. We run the Markov chain for  $55\,000$ iterations, discarding the first $5\,000$  as a burnin period. Variable selection is performed by checking whether $0$ is contained in the credible set (see \cite{horse}). Finally, we implement the  LASSO (\cite{tibshirani2011regression}) 
and  report the model chosen according to the $1$-s.e. rule (\cite{friedman2001elements})).
	
	\begin{figure}[t!]
	\begin{subfigure}[b]{0.50\columnwidth}
		\centering
		\includegraphics[width=1.0\columnwidth]{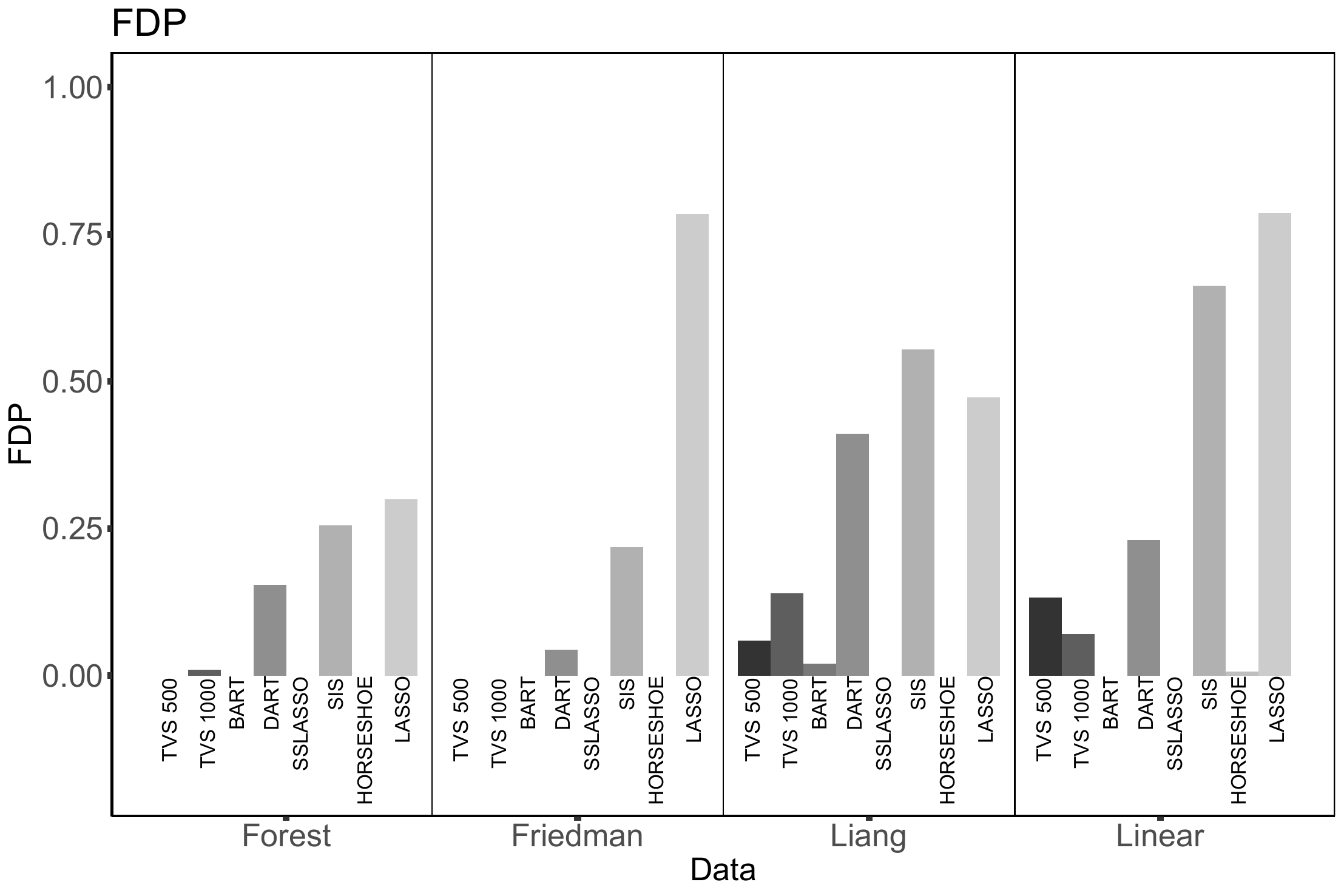}
		\caption{FDP}\label{fig:fdr2}
	\end{subfigure}
	~
	\begin{subfigure}[b]{0.50\columnwidth}
		\centering
		\includegraphics[width=1.0\columnwidth]{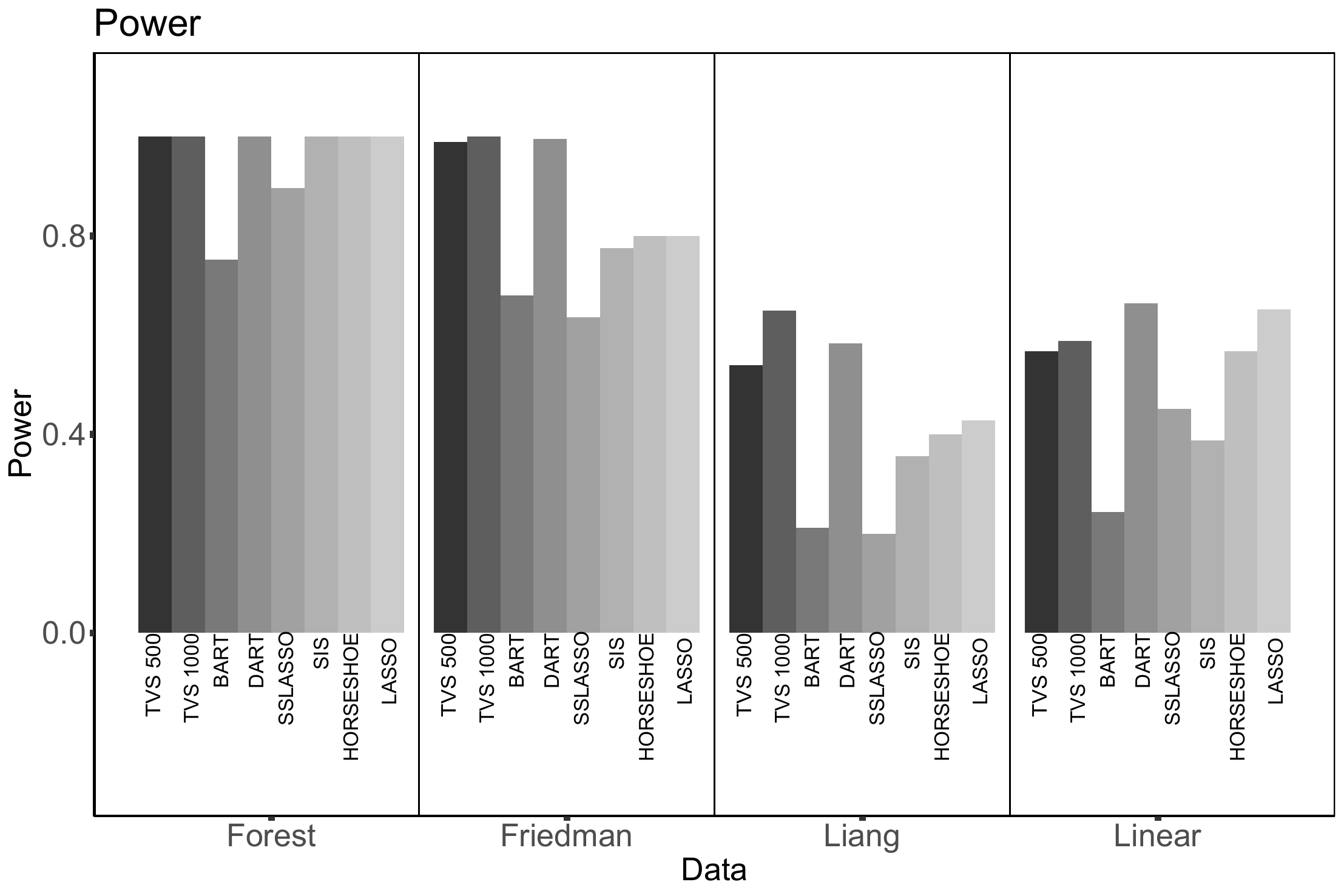}
		\caption{Power}\label{fig:power2}
	\end{subfigure}
	\\
	\begin{subfigure}[b]{0.50\columnwidth}
		\centering
		\includegraphics[width=1.0\columnwidth]{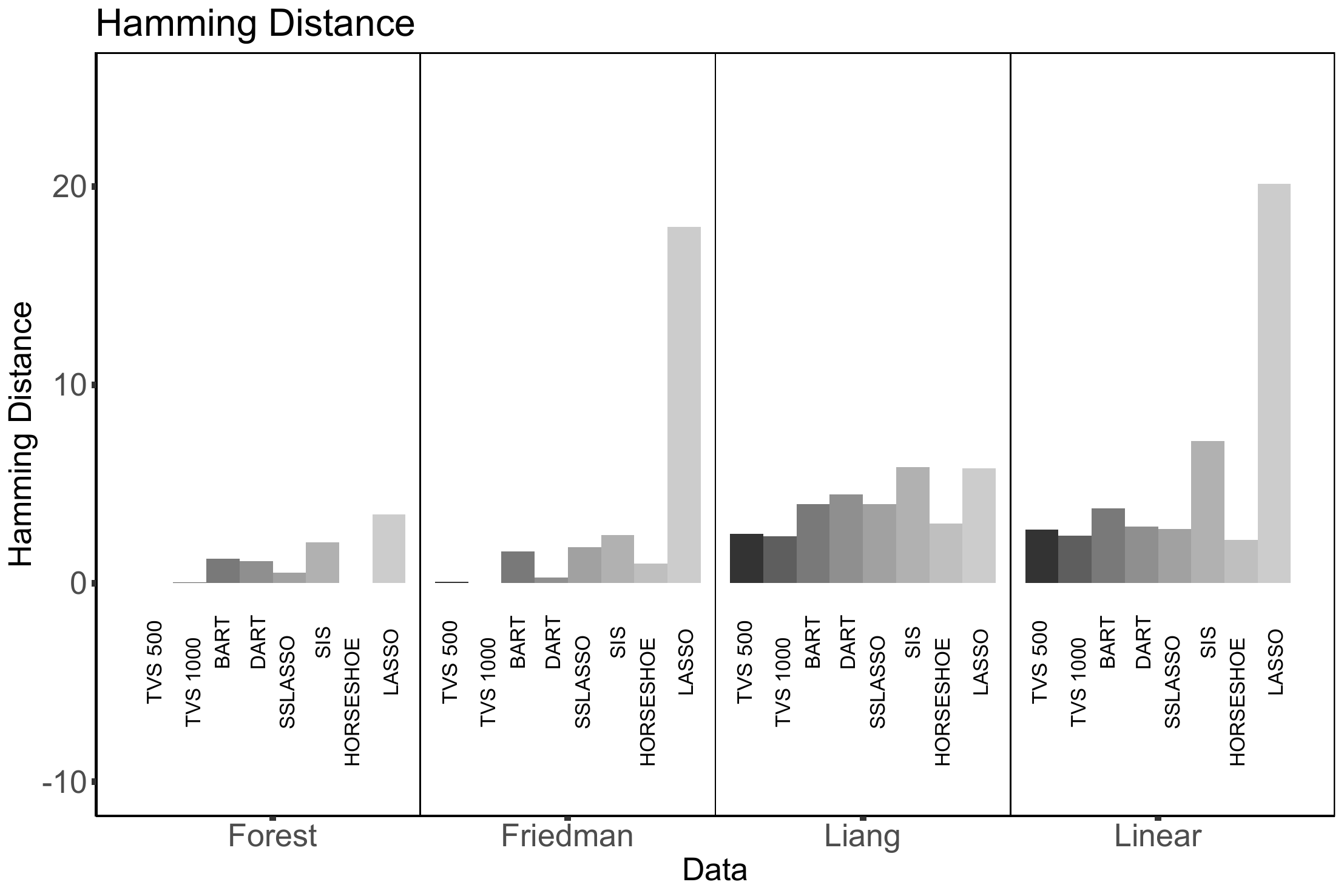}
		\caption{Hamming Distance}\label{fig:ham2}
	\end{subfigure}
	~
	\begin{subfigure}[b]{0.50\columnwidth}
		\centering
		\includegraphics[width=1.0\columnwidth]{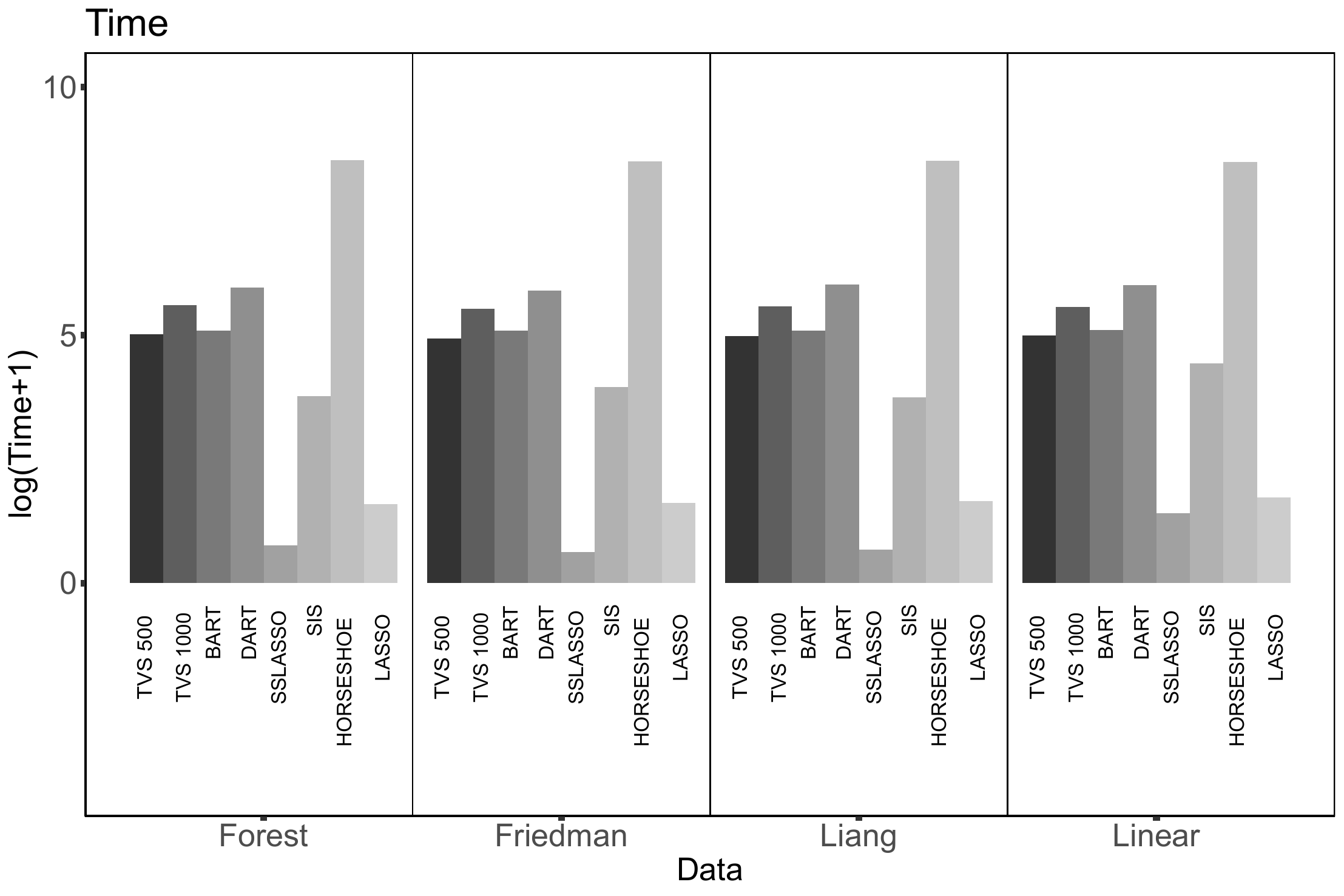}
		\caption{Time}\label{fig:time2}
	\end{subfigure}
	\caption{\label{tab:image}{\small Graphs denoting FDP (\ref{fig:fdr2}), Power (\ref{fig:power2}), Hamming Distance (\ref{fig:ham2}), and Time (\ref{fig:time2}) for the $4$ choices of $f_0$ assuming $p=10\,000$ and $n =300$. The $x$-axis denotes the choice of $f_0$ and the various methods are marked with various shades of gray. For TVS, we have two choices $M=500$ and $M=1\,000$.}}
	
\end{figure}

	
	
	We report the average performance (over $50$ datasets) for $p=10\,000$ in Figure \ref{tab:image} and  the rest (for $p=1\,000$) in the Appendix.  Recall that the model estimated by TVS is obtained by truncating $\pi_i(500)$'s at $0.5$.  In terms of the Hamming distance,  we notice that TVS performs best across-the-board. DART (with $D=50$) performs consistently well in terms of variable selection, but the timing comparisons are less encouraging. BART (with $D=20$) takes a relatively comparable amount of time as TVS with $M=1\,000$, but suffers from less power. SS-LASSO's performance is strong, in particular for the less non-linear data setups.  The performance of TVS is seen to improve with $M$.  SIS is  a screening method  based on a  linear model assumption.  Albeit faster than TVS,  we observe  that SIS generally over-selects.  In addition, SIS screens out key signals with non-linear effects (Liang and Friedman datasets).  The Horseshoe prior performs very well but is much slower. LASSO seems to overfit, in spite of the 1-se rule.

{	
We also implement a stopping criterion for TVS based on the stabilization of the inclusion probabilities
$\pi_i(t)=\frac{a_i(t)}{a_i(t) +b_i(t)}$.
One possibility is to stop TVS when the estimated model $\wh\mS_t$ obtained by truncating $\pi_i(t)$'s at $0.5$ has {\em not} changed for over, say, $100$ consecutive TVS iterations.  With this convergence criterion,  the convergence times differs across the different data set-ups.
Generally, TVS is able to converge in $\sim200$ iterations for $p=1\,000$ and  $\sim 300$ iterations for $p=10\,000$. While the computing times are faster, TVS may be more conservative (lower FDP but also lower Power). The Hamming distance is hence a bit larger, but comparable to TVS with $500$ iterations (Appendix C.1)

	}

	\begin{figure}[t!]
	\begin{subfigure}[b]{0.50\columnwidth}
		\centering
		\includegraphics[width=1.0\columnwidth]{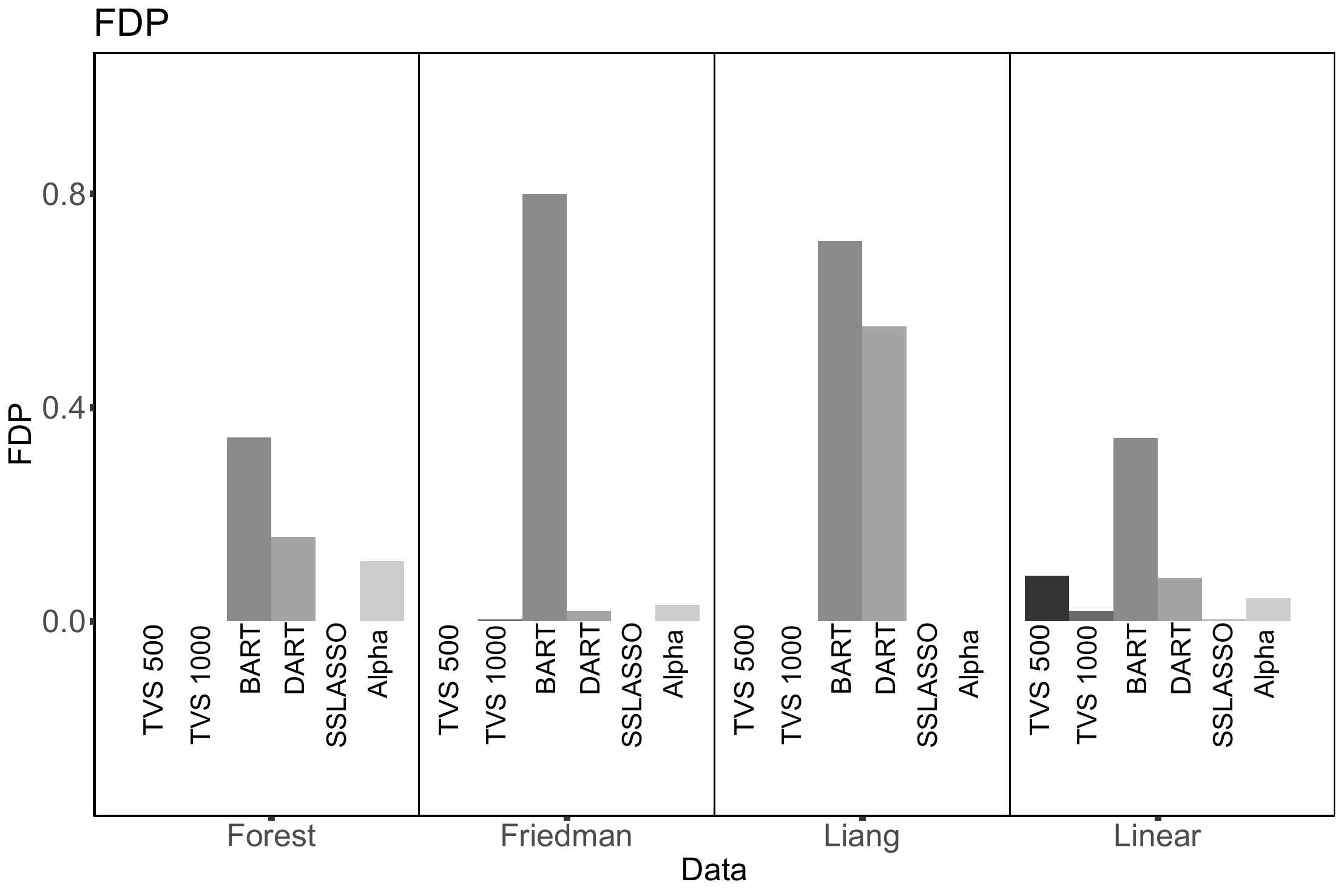}
		\caption{FDP}\label{fig:fdr_online}
	\end{subfigure}
	~
	\begin{subfigure}[b]{0.50\columnwidth}
		\centering
		\includegraphics[width=1.0\columnwidth]{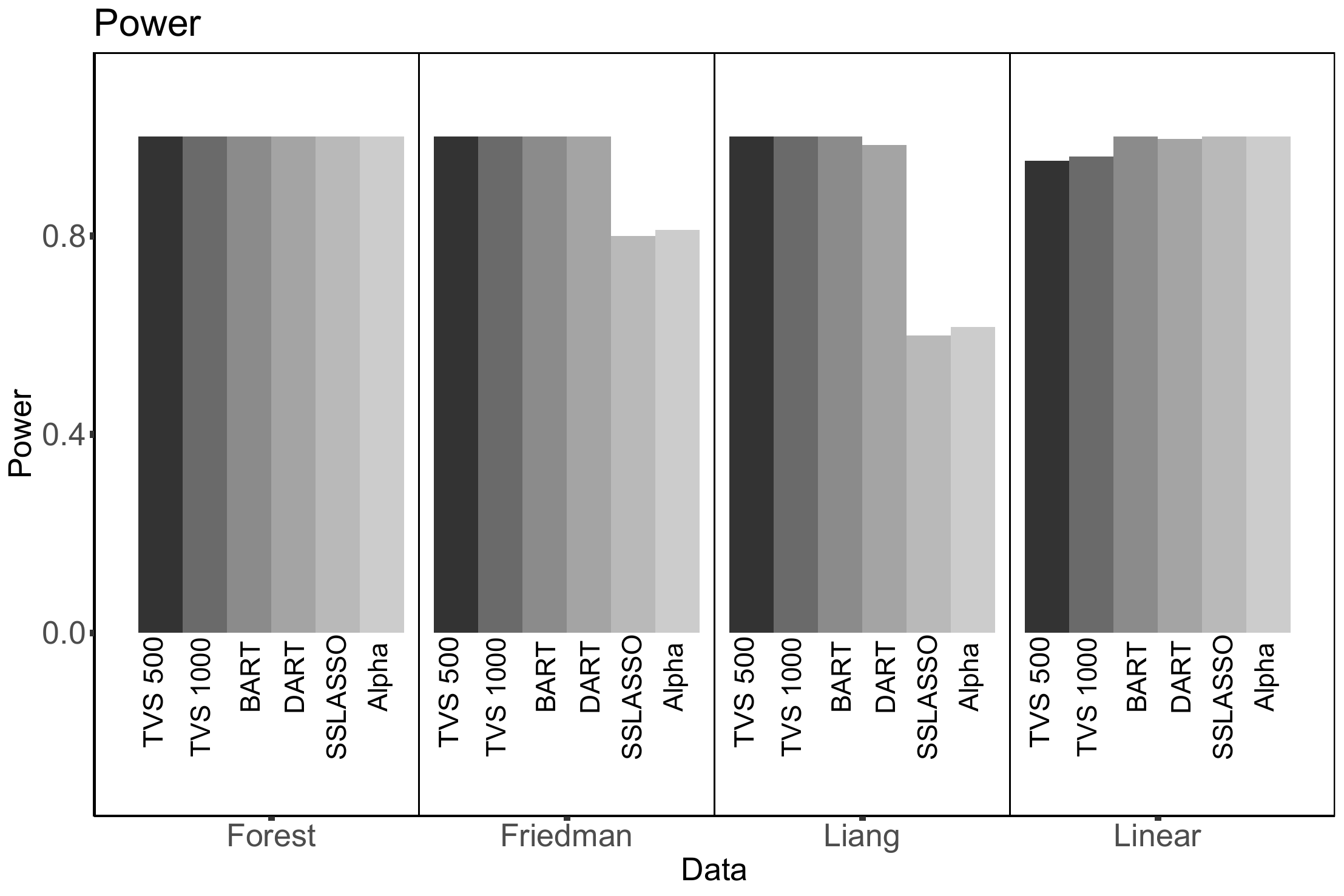}
		\caption{Power}\label{fig:power_online}
	\end{subfigure}
	\\
	\begin{subfigure}[b]{0.50\columnwidth}
		\centering
		\includegraphics[width=1.0\columnwidth]{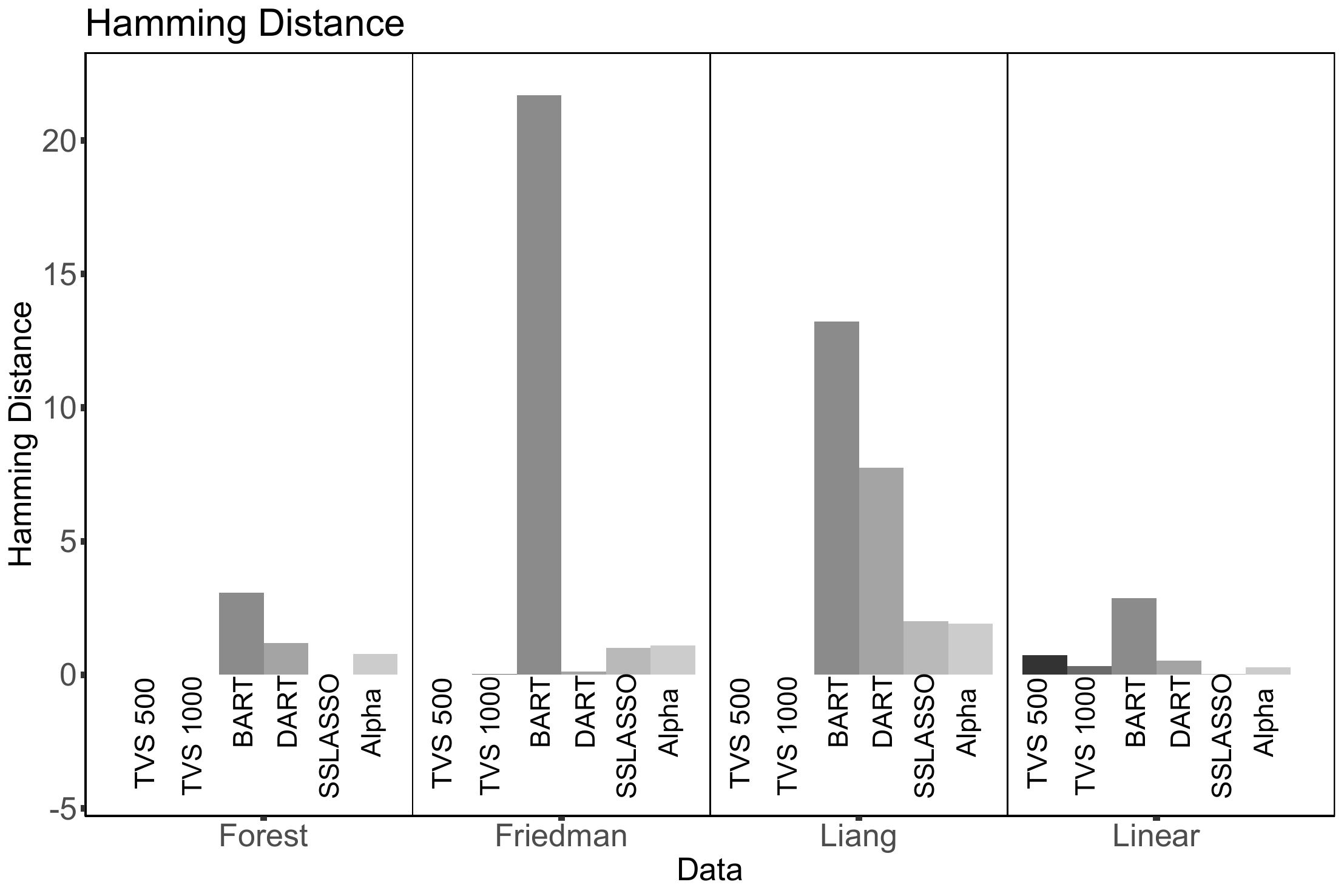}
		\caption{Hamming Distance}\label{fig:ham_online}
	\end{subfigure}
	~
	\begin{subfigure}[b]{0.50\columnwidth}
		\centering
		\includegraphics[width=1.0\columnwidth]{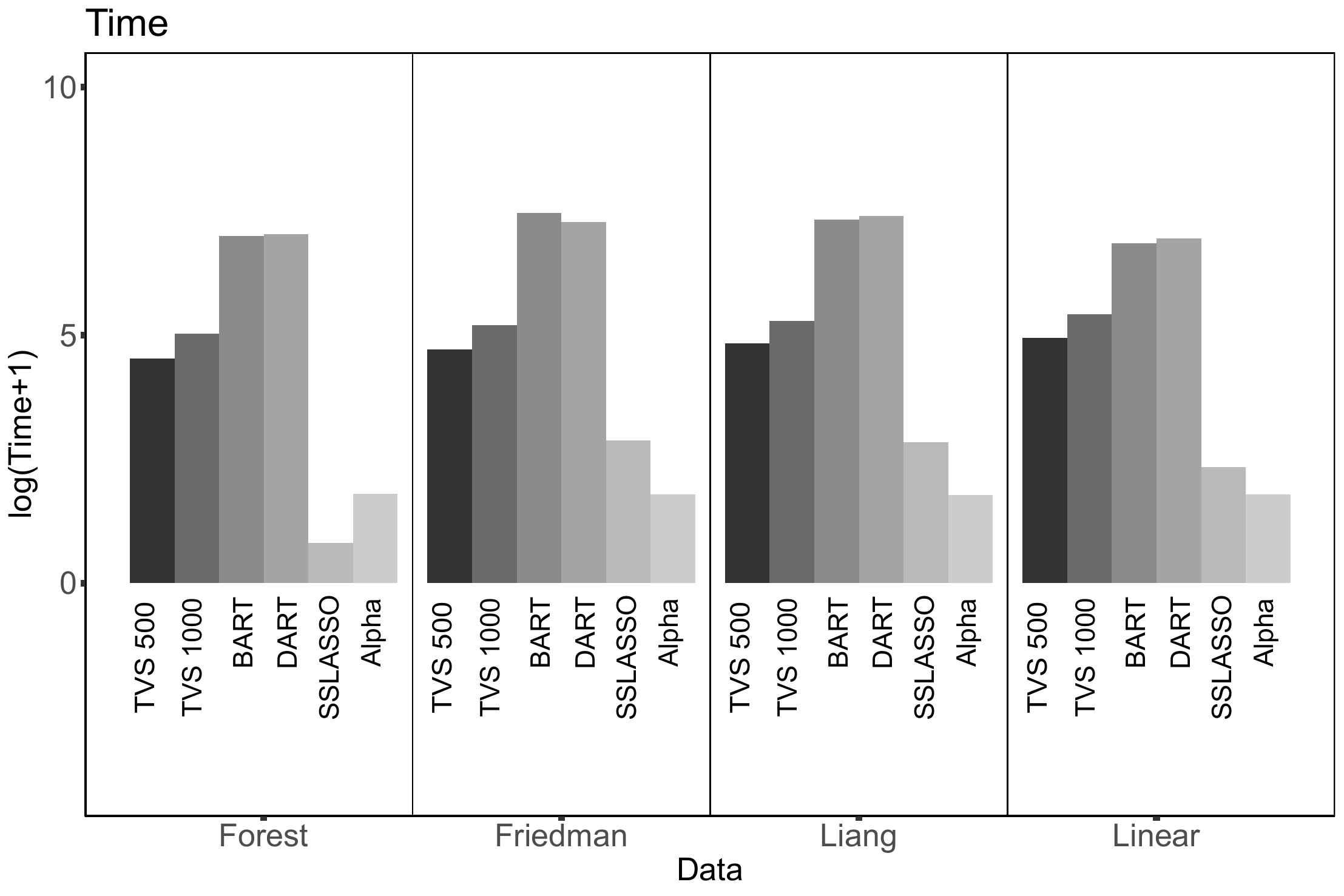}
		\caption{Time}\label{fig:time_online}
	\end{subfigure}
	\caption{{\small Graphs denoting FDP (\ref{fig:fdr_online}), Power (\ref{fig:power_online}), Hamming Distance (\ref{fig:ham_online}), and Time (\ref{fig:time_online}) for the $4$ choices of $f_0$ assuming $p=1\,000$ and $n =10\,000$. The $x$-axis denotes the choice of $f_0$ and the various methods are marked with various shades of gray. For TVS, we have two choices $M=500$ and $M=1\,000$, both with $s=1000$.}}
	\label{tab:image_online}
\end{figure}

\subsection{Online Cases}
\label{sec:simu_online}
	We now consider a simulation scenario where $n>>p$, i.e. $p =1\,000$ and $n=10\,000$. 
As described earlier in Section \ref{sec:online}, we partition the data into minibatches $(\Y^{(b)},\X^{(b)})$ of size $s$, where $\Y^{(b)}=(Y_i: (b-1)s+1\leq i\leq bs)$ and $\X^{(b)}=[\bm x_{i}:(b-1)s+1\leq i\leq bs]'$ for $b=1,\dots, n/s$ with $s\in \{500,1000\}$ and $M \in \{500,1000\} $  using $D=10$ trees. In this study, we consider the same four set-ups as in Section \ref{sec:simu_offline}. We implemented TVS  with a fixed number of rounds $r\in\{1,5,10\}$ and a version with a stopping criterion based on the stabilization of the inclusion probabilities $\pi_i(t)=\frac{a_i(t)}{a_i(t) +b_i(t)}$. This means that TVS will terminate when the estimated model $\widehat{\mS}_t$ obtained by truncating $\pi(t)$'s at $0.5$ has not changed for  $100$ consecutive iterations. The results using the stopping criterion are reported in Figure \ref{tab:image_online} and the rest is in the Appendix (Section \ref{sec:appendix_online_simul} ). As before, we report the best configuration for BART and DART, namely $D =20$ for BART and $D=50$ for DART (both with  $50\,000$ MCMC iterations). For both methods, there are non-negligible false discoveries and the timing comparisons are not encouraging. In addition, we could not apply both BART and DART  with $n\geq 50\,000$ observations due to insufficient memory. For TVS,  we found the batch size $s= 1\,000$ to work better, as well as  running the algorithm for enough rounds until the inclusion probabilities have stabilized (Figure \ref{tab:image_online} reports the results with a stopping criterion). The results are very encouraging. 
	
Streaming feature selection methods  are still being developed  (\cite{wang2018provable,fahy2019dynamic,ramdas2017online,javanmard2018online}).
We have compared our online variant with the $\alpha$-investing method for streaming variable selection of \cite{zhou2006streamwise}. This method dynamically adjusts $p$-value thresholds for adding features to the model. We run this method using $p$-values from applying linear regression on batches of streaming data. The results are shown in Figure \ref{tab:image_online}. This procedure performs well in Forest and Linear set-ups but misses the key non-linear variables in Friedman and Liang setups. While SSLASSO's performance is very strong, we notice that in the non-linear setup of  \cite{liang2018bayesian} there are false non-discoveries.

\section{Application on Real Data}
\label{sec:real_data}
\subsection{HIV Data}
\label{sec:hiv}
We will apply (offline) TVS on  a benchmark Human Immunodeficiency Virus Type I (HIV-I) data described and analyzed in \cite{rhee_genotypic_2006} and \cite{barber_controlling_2015}. 	
This  publicly available\footnote{Stanford HIV Drug Resistance Database \url{https://hivdb.stanford.edu/pages/published_analysis/genophenoPNAS2006/}} dataset consists of genotype 
and resistance measurements (decrease in susceptibility on a log scale)  to three drug classes: (1) protease inhibitors (PIs),  (2) nucleoside reverse transcriptase inhibitors (NRTIs), and (3) non-nucleoside reverse transcriptase inhibitors (NNRTIs). 

 The goal in this analysis is to  find mutations in the HIV-1 protease or reverse transcriptase that are associated with drug resistance.
 Similarly as in \cite{barber_controlling_2015}
 we analyze each drug separately. The response $Y_i$ is given by the log-fold increase of lab-tested drug resistance in the $i^{th}$ sample with the design matrix $X$ consisting of binary indicators $x_{ij}\in \{0,1\}$  for whether or not the $j^{th}$ mutation  has occurred at the $i^{th}$ sample.\footnote{As suggested in the analysis of \cite{barber_controlling_2015}, when analyzing each drug, only mutations that appear $3$ or more times in the samples are taken into consideration.} 

{In an independent experimental study,  \citep{soo-yon_rhee_hiv-1_2005}   identified mutations that are present at a significantly higher frequency in virus samples from patients who have been treated with each class of drugs as compared to patients who never received such treatments.  While, as with any other real data experiment, the ground truth is unknown, we treat this independent study are a good approximation to the ground truth. Similarly as \cite{barber_controlling_2015},  we only compare mutation positions since multiple mutations in the same position are often associated with the same drug resistance outcomes.

{For illustration, we now focus on one particular drug called Lopinavir (LPV). There are $p=206$ mutations and $n=824$ independent samples available for this drug. 
 TVS was applied to this data for $T=500$ iterations with  $M=1\,000$ inner BART iterations. In Figure \ref{tab:outcomes2}, we differentiate those mutations whose position were identified
by  \cite{soo-yon_rhee_hiv-1_2005}  and mutations which were not identified with blue and red  colors, respectively. 
From the plot of inclusion probabilities in Figure \ref{fig:LPV}, it is comforting to see  that only one unidentified mutation has a posterior probability $\pi_j(t)$ stabilized above the $0.5$ threshold.  Generally, we observe the experimentally identified mutations (blue curves) to have higher inclusion probabilities.   }

\begin{figure}[t!]
	\begin{subfigure}[b]{0.5\columnwidth}
		\centering
		\includegraphics[width=1\columnwidth]{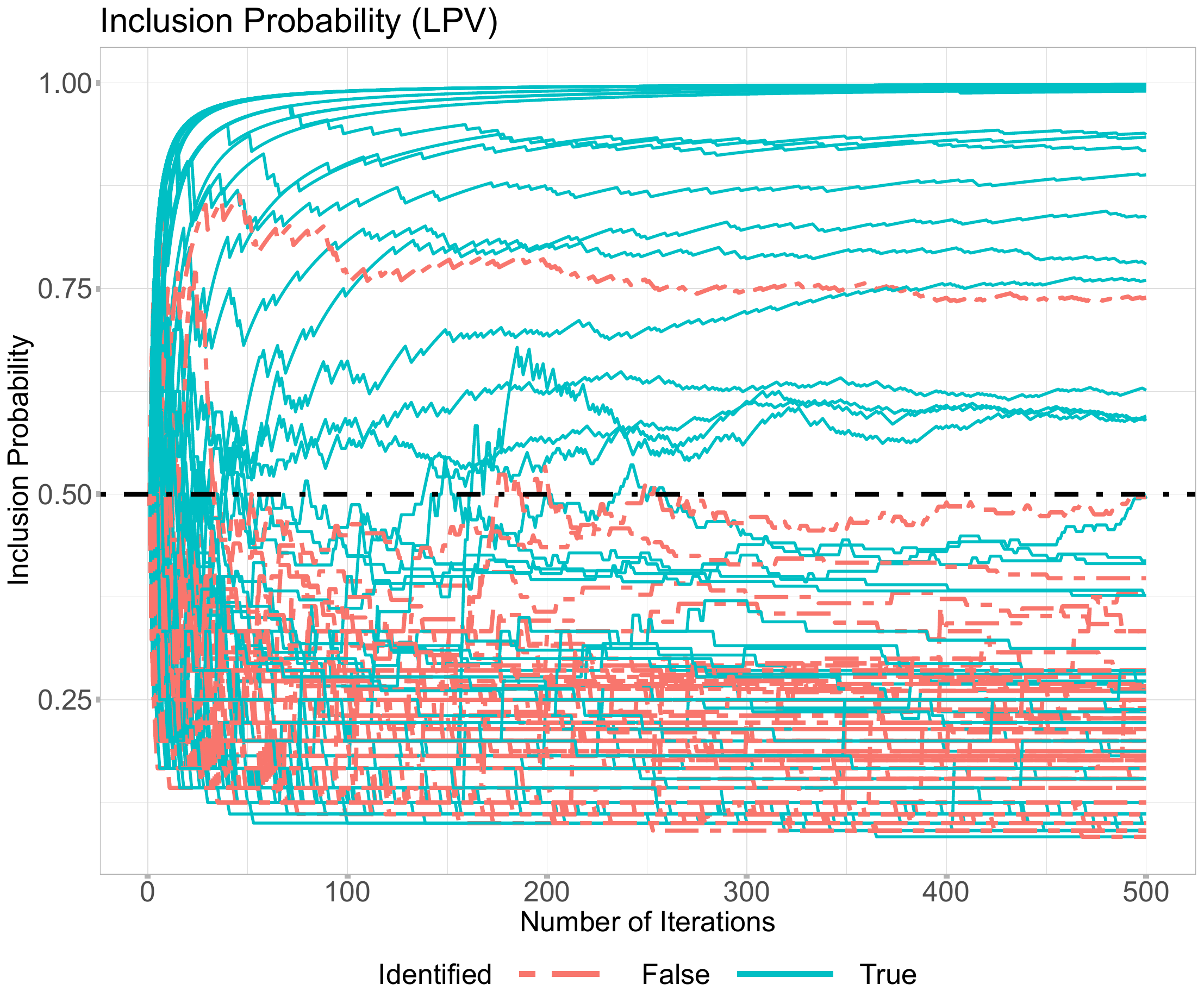}
		\caption{TVS Inclusion Probabilities}\label{fig:LPV}
	\end{subfigure}
	~
	\begin{subfigure}[b]{0.50\columnwidth}
		\centering
		\includegraphics[width=1\columnwidth]{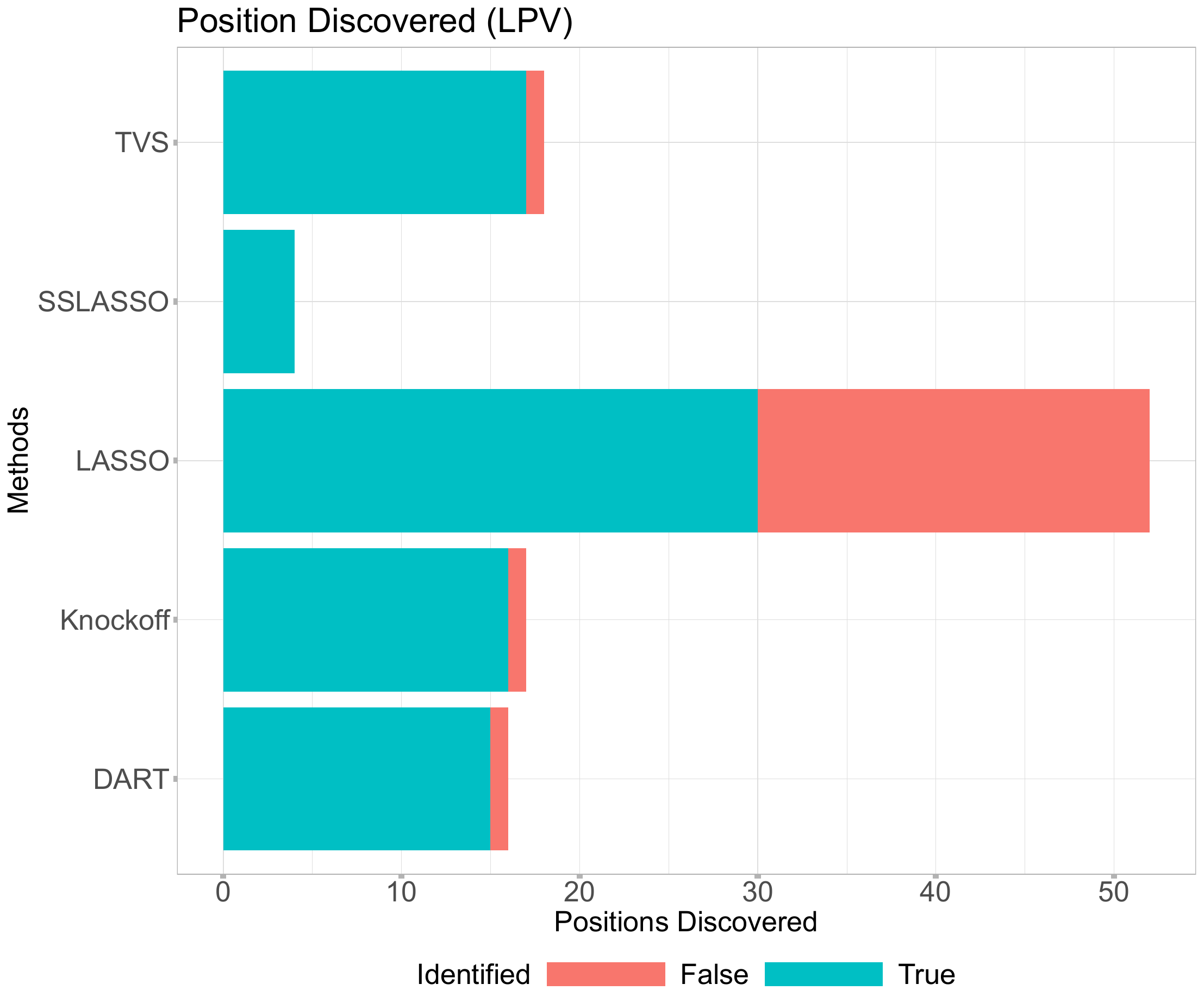}
		\caption{Comparisons with other Methods}\label{fig:LPVbar}
	\end{subfigure}
	
	\caption{{\small Thompson Variable Selection on the LPV dataset. (Left) Trajectories of the inclusion probabilities $\pi_j(t)$. (Right) Number of signals discovered, where blue denotes the experimentally validated signals and red are the unvalidated ones.}}
	\label{tab:outcomes2}
\end{figure}	

Comparisons are made with DART, Knockoffs \citep{barber_controlling_2015}, LASSO (10-fold cross-validation), and Spike-and-Slab LASSO (\cite{rockova_spike-and-slab_2018}), choosing $\lambda_1=0.1$ and $\lambda_0\in\{\lambda_1+10\times k; k=0,1,\dots, p\}$). Knockoffs, LASSO, and  the Spike-and-Slab LASSO assume a linear model with no interactions.  DART was implemented using $T=50$ trees and $50\,000$ MCMC iterations, where we select those variables whose average number splits was at least one.  The numbers of discovered Positions for each method are plotted in Figure \ref{fig:LPVbar}. While LASSO selects many more experimentally validated  mutations, it also includes many unvalidated ones. TVS, on the other hand, has a very small number of ``false discoveries" while maintaining good power. Additional results are included in the Appendix (Section \ref{sec:supplement_hiv}).

}

\subsection{Durable Goods Marketing Data Set}
\label{sec:marketing_intro}
Our second application examines a cross-sectional dataset described in  Ni et al. (2012) consisting of durable goods sales data from a major anonymous  U.S. consumer electronics retailer. The dataset features the results of a direct-mail promotion campaign in November 2003 where
 roughly half of the $n=176\,961$ households received a promotional mailer with $10\$$ off their purchase during the promotion time period (December 4-15). If they did purchase, they would get $10\%$ off on a subsequent purchase through December. The treatment assignment was random. The data contains $p=146$ descriptors of all customers including prior purchase history, purchase of warranties etc. We will investigate the effect of the promotional campaign (as well as other covariates) on December sales. In addition, we will interact the promotion mail indicator with customer characteristics to identify the ``mail-deal-prone" customers.

We dichotomized December purchase (in dollars) to create a binary outcome  $Y_i=\mathbb{I}(\texttt{December-sales}_i>0)$ for whether or not  the $i^{th}$ customer  made any purchase in December.
Regarding predictor variables, we removed any variables with missing values and any binary variables with less than $10$ samples in one group. 
This pre-filtering leaves us with $114$ variables whose names and descriptive statistics are reported in Section \ref{sec:additional_marketing_data} in the Appendix. We interact the promotion mail indicator with these variables to obtain $p=227$ predictors. Due to the large volume of data ($n\approx 180\,000$), we were unable to run  DART and BART (\texttt{BART} package implementation) due to memory problems. This highlights the need for  TVS as a variable selector which can handle such  voluminous data.

\begin{figure}[t!]
	\begin{subfigure}[b]{0.31\columnwidth}
		\centering
		\includegraphics[width=1\columnwidth]{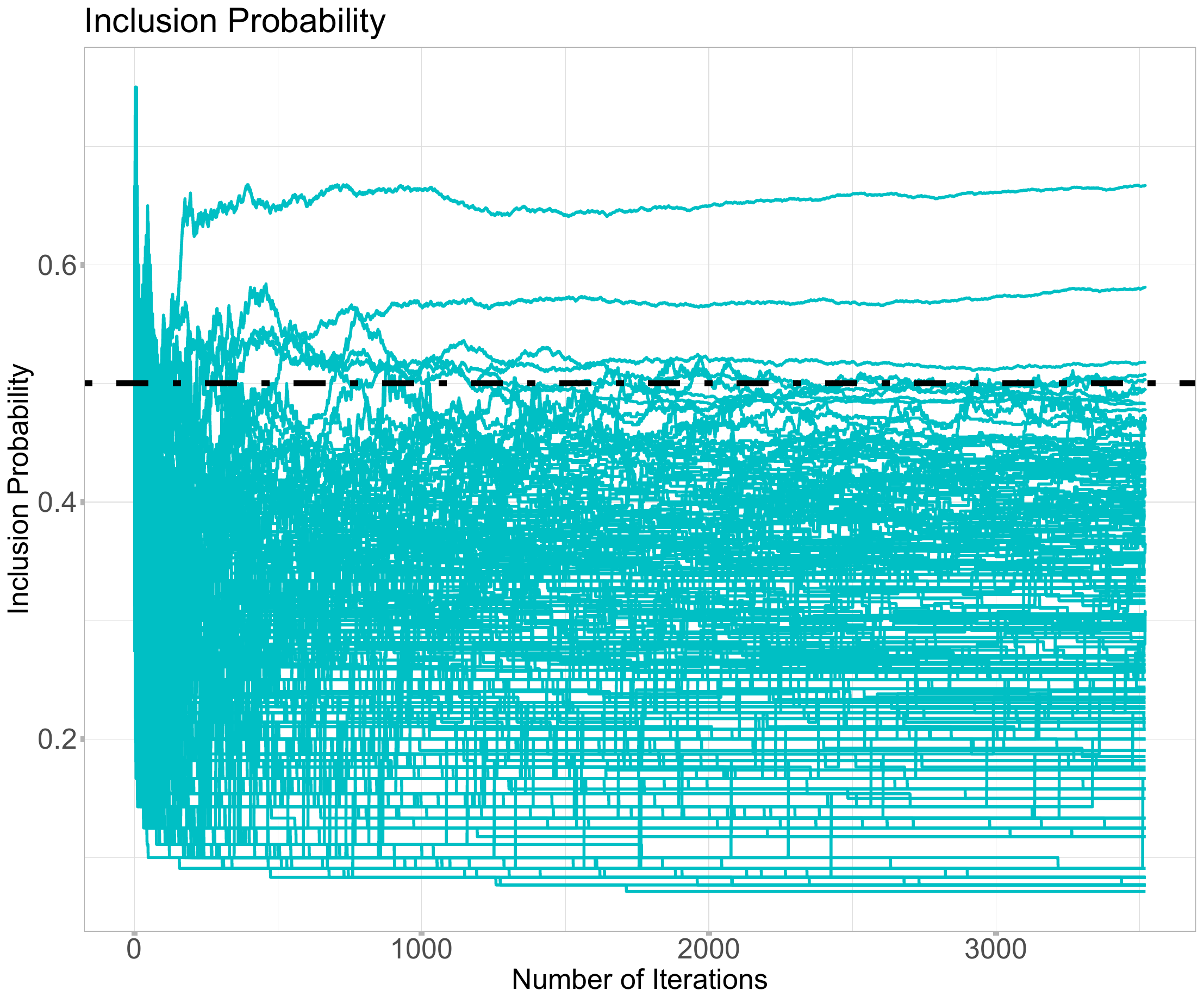}
		\caption{$s=1\,000$}\label{fig:noknockoff}
	\end{subfigure}
	\begin{subfigure}[b]{0.31\columnwidth}
		\centering
		\includegraphics[width=1\columnwidth]{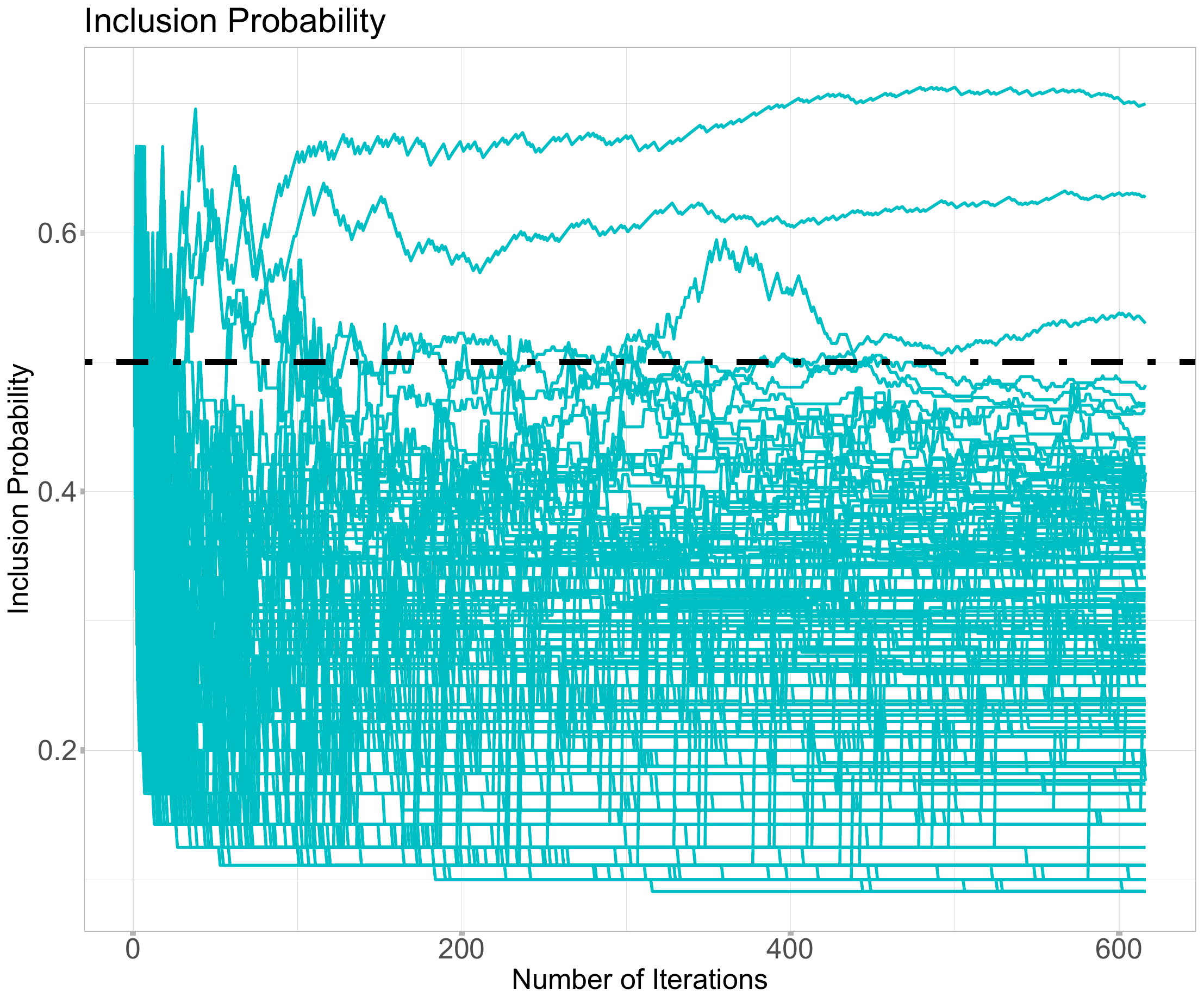}
		\caption{$s=2\,000$}
	\end{subfigure}
	\begin{subfigure}[b]{0.31\columnwidth}
		\centering
		\includegraphics[width=1\columnwidth]{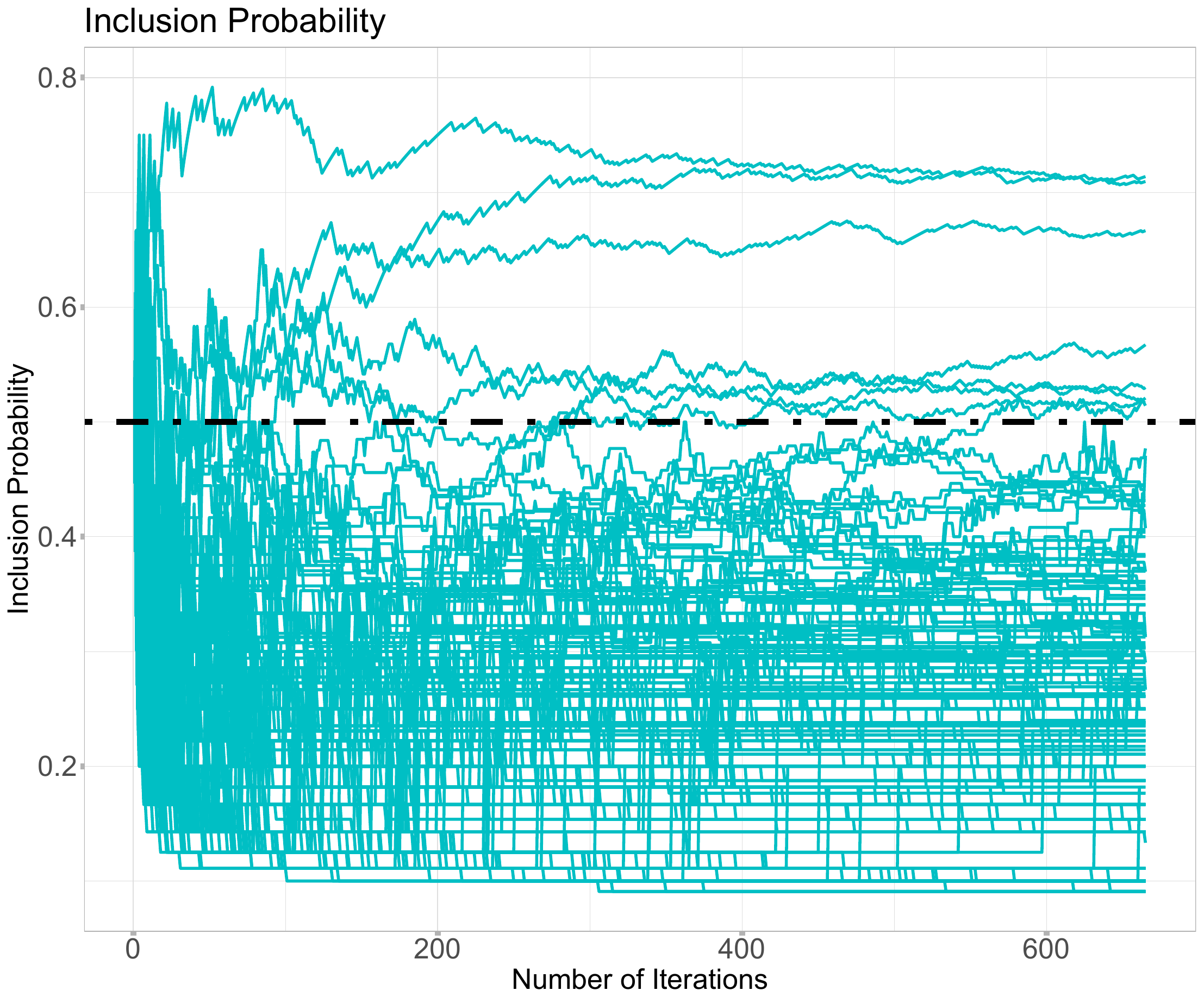}
		\caption{$s=5\,000$}
	\end{subfigure}

	\begin{subfigure}[b]{0.31\columnwidth}
		\centering
		\includegraphics[width=1\columnwidth]{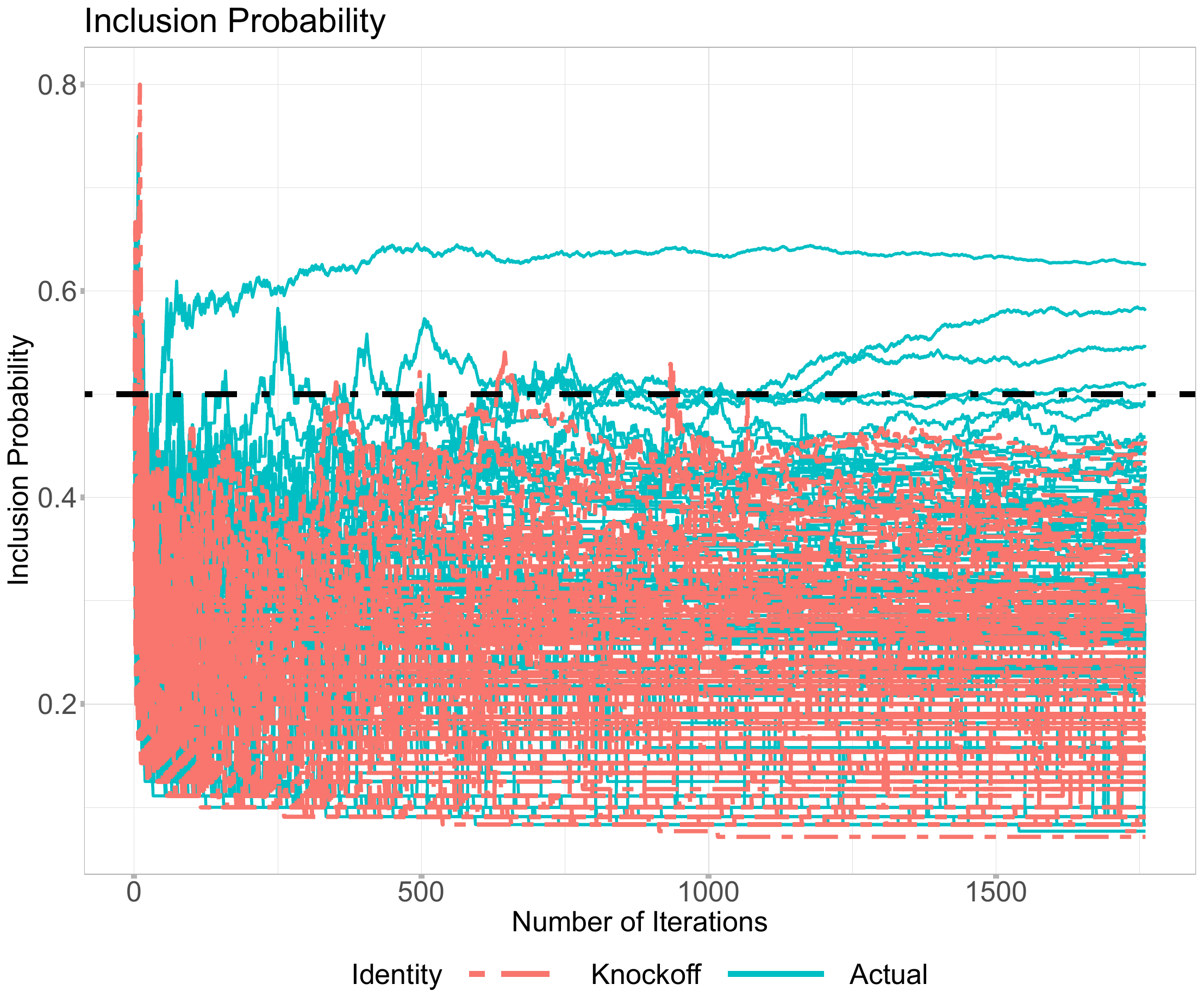}
		\caption{$s=1\,000$}\label{fig:1000knockoff}
	\end{subfigure}	
	\begin{subfigure}[b]{0.31\columnwidth}
		\centering
		\includegraphics[width=1\columnwidth]{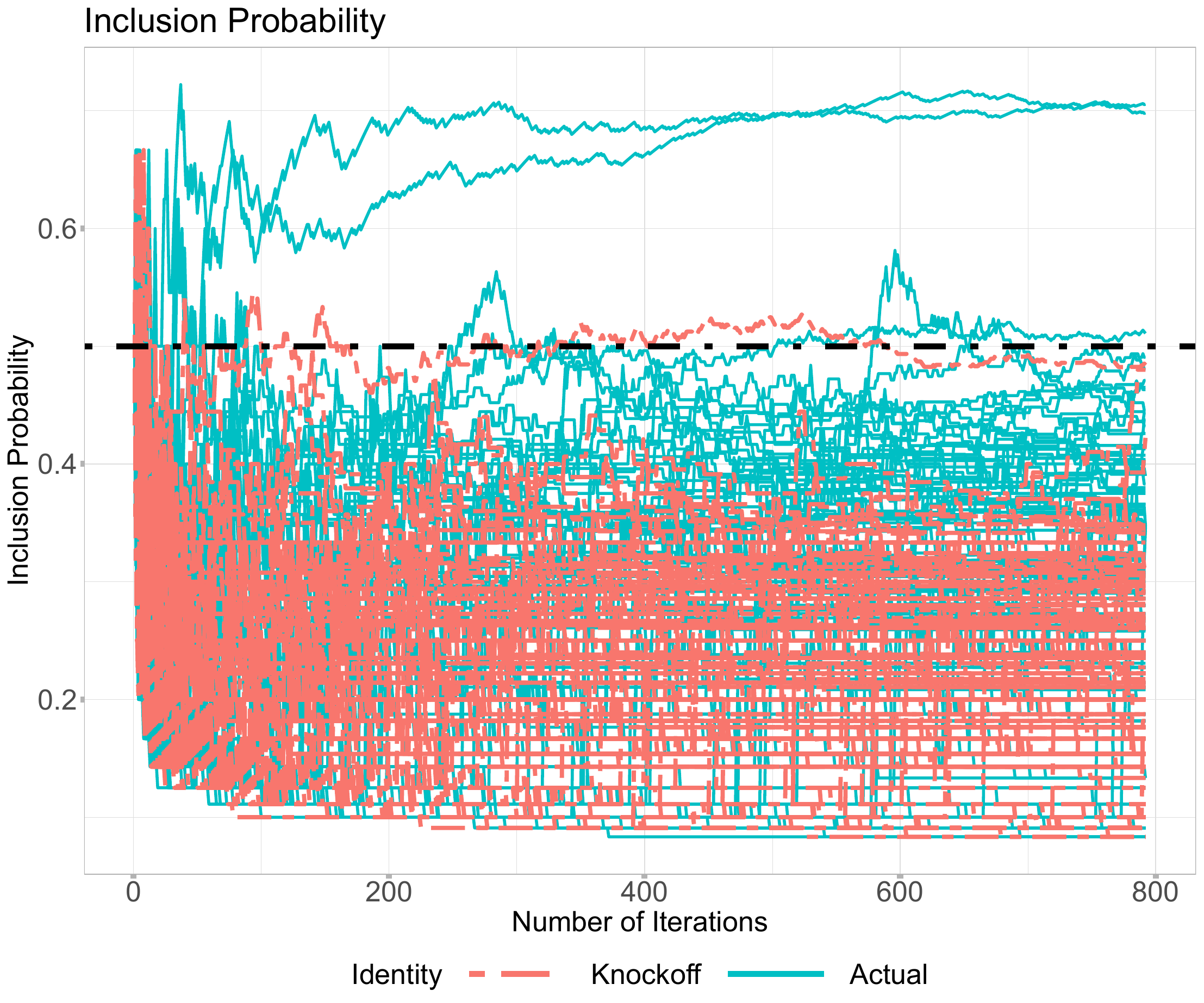}
		\caption{$s=2\,000$}
	\end{subfigure}	
	\begin{subfigure}[b]{0.31\columnwidth}
		\centering
		\includegraphics[width=1\columnwidth]{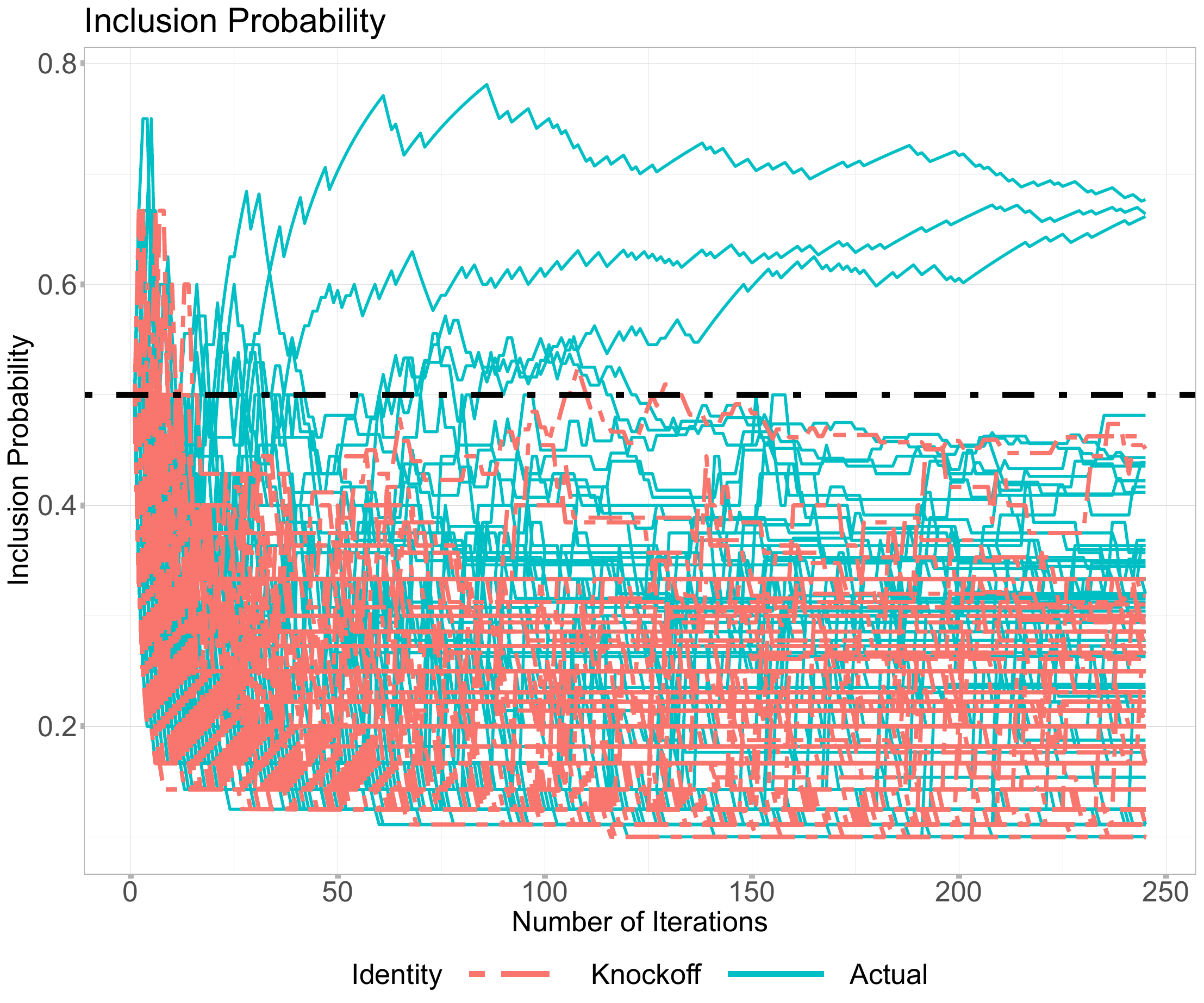}
		\caption{$s=5\,000$}
	\end{subfigure}
	\caption{{\small Thompson Variable Selection on the marketing data. (Left) Trajectories of the inclusion probabilities $\pi_j(t)$ without knockoffs. (Right) Trajectories of the inclusion probabilities $\pi_j(t)$ with knockoffs (in red).}}
	\label{tab:outcomes3}
\end{figure}

Unlike the HIV-I data in Section \ref{sec:hiv}, there is no proxy for the ground truth. 
To understand the performance quality of TVS, we added  $227$ normally distributed knockoffs. The knockoffs are generated using \texttt{create.second\_order} function in the \texttt{knockoff} R package (\cite{knockoff_package}) using a Gaussian Distributions with the same mean and covariance structure (\cite{candes_panning_2018}).  We run TVS with a batch size $s\in\{1\,000,2\,000,5\,000\}$ and  $M = 1\,000$ inner iterations until the posterior probabilities have stabilized. The inclusion probabilities are plotted in Figure \ref{tab:outcomes3} for two cases (a) without knockoffs (the first row) and (b) with knockoffs (the second row).  It is interesting to note that, apart from one setting with $s=1\,000$,  the knockoff trajectories are safely suppressed below $0.5$ (dashed line). Both with and without knockoffs, TVS chooses `the number of months with purchases in past 24 month' and `the November Promotion Sales' as important variables.  The selected variables for each combination of settings are summarized in Table \ref{tab:market_select}.

\begin{table}[!t]
\centering
	\scalebox{0.8}{\begin{tabular}{|l|ll|ll|ll|}
			\hline
			& \multicolumn{2}{l|}{$s=1\,000$} & \multicolumn{2}{l|}{$s=2\,000$} & \multicolumn{2}{l|}{$s=5\,000$} \\ \hline
			Knockoff                                                                    & Yes            & No             & Yes            & No             & Yes            & No             \\ \hline
			total number of medium ticket items in previous 60 months                   & 0.30           & \textbf{0.51}           & \textbf{0.51}           & 0.46           & 0.44           & \textbf{0.51 }          \\
			total number of small ticket items in previous 60 months                    & 0.49           & \textbf{0.52 }          & 0.40           & 0.41           & 0.25           &\textbf{ 0.53 }          \\
			number of months shopped once in previous 12 months                         & 0.34           & 0.41           & 0.44           & 0.42           & 0.41           & \textbf{0.57}           \\
			number of months shopped once in previous 24 months                         & \textbf{0.63}           & \textbf{0.67}           & \textbf{0.70}           &\textbf{ 0.63}           & \textbf{0.66 }          &\textbf{ 0.71}           \\
			count of unique purchase trips in previous 24 months                        & \textbf{0.55 }          & 0.26           & 0.48           & \textbf{0.53  }         & \textbf{0.68}           & \textbf{0.67  }         \\
			total number of items purchased in previous 12 months                       & \textbf{0.51}           & 0.30           & 0.24           & 0.21           & 0.24           & 0.11           \\
			promo\_nov period:  total sales                                             &\textbf{ 0.58}           & \textbf{0.58 }          & \textbf{0.71 }          & \textbf{0.70 }          & 0.33           & 0.45           \\
			mailed in holiday 2001 mailer                                               & 0.25           & 0.43           & 0.08           & 0.41           & 0.42           & \textbf{0.52}           \\
			percent audio category sales of total sales $\times$ mail indicator         & 0.41           & \textbf{0.50}           & 0.15           & 0.28           & 0.13           & 0.10           \\
			promo\_nov period:  total sales $\times$ mail indicator                     & 0.15           & 0.35           & 0.46           & 0.41           & \textbf{0.66}           & \textbf{0.71}           \\
			indicator of holiday mailer 2002 promotion response $\times$ mail indicator & 0.19           & 0.38           & 0.44           & 0.21           & 0.44           & \textbf{0.52}           \\ \hline
	\end{tabular}}
	\caption{\small Variables Selected by TVS with different $s$ and with/without knockoff. The numbers report conditional inclusion probabilities $\pi_i(t)$ after convergence, where values above $0.5$ are in bold.}\label{tab:market_select}
\end{table}

Finally, we used  the same set of variables (including knockoff variables) for different variable selection methods and recorded the number of knockoffs chosen by each one. We used BART ($D=20$, MCMC iteration = $50\,000$), and DART  ($D=50$, MCMC iteration = $50\,000$) with the same selection criteria as before, i.e. a variable is selected if it was split on average at least once. We also consider LASSO where the sparsity penalty $\lambda$ was chosen by cross-validation. BART and DART cannot be run on the entire data set so we only run it on a random subset of $10\,000$ data points. While TVS with large enough $s$ does not include any of the knockoffs, LASSO does include $14$ and  DART includes $4$.

\section{Discussion}\label{sec:discussion}
Our work pursues an intriguing connection  between spike-and-slab variable selection and  binary bandit problems. This pursuit has lead to  a proposal of  Thompson Variable Selection,  a reinforcement learning wrapper algorithm for fast variable selection in high dimensional non-parametric problems.   
In related work, \cite{liu_abc_2018} developed an ABC sampler for variable subsets through a split-sample approach by (a) first proposing a subset $\mS_t$ from a prior, (b) keeping only those subsets that yield pseudo-data that are sufficiently close to the left-out sample. TVS can be broadly regarded as a reinforcement learning elaboration of this strategy where, instead of sampling from a (non-informative) prior $\pi(\mS_t)$,  one  ``updates the prior $\pi(\mS_t)$"  by learning from previous successes. 

TVS can be regarded as a stochastic optimization approach to subset selection which balances exploration and exploitation. TVS is suitable in settings when very many predictors and/or very many observations can be too overwhelming for machine learning. By sequentially parsing subsets of data and reinforcing promising covariates, TVS can effectively separate signal from noise, providing a platform for interpretable machine learning. TVS minimizes regret by sequentially computing a median probability model rule obtained by truncating  sampled mean rewards.
We provide bounds for this regret without necessarily assuming that the mean arm rewards be unrelated. We observe strong   empirical performance of TVS under various scenarios, both on real and simulated data. {\color{blue}
The TVS approach, coupled with BART,  captures non-linearities and may thus be beneficial over linear techniques. In addition, TVS scales to very large datasets.

Thompson sampling is ultimately designed to minimize regret, not to select the best arms.  \cite{bubeck2009pure} point out that algorithms satisfying regret bounds of order $\log(T)$ can be far from optimal for finding the best arm.  \cite{russo2016simple} proposed a `top-two' sampling version of Thompson sampling which is tailored for single best-arm identification.  Alternative algorithms have been proposed including the Successive Elimination algorithm of \cite{even2006action} for single top-arm identification and the SAR (Successive Accepts Rejects) algorithm of \cite{bubeck2013multiple} for top $m$-arm identification. These works suggest possible refinements of our approach for directly targeting multiple best arms in correlated combinatorial bandits.

}

\bibliographystyle{chicago}
\bibliography{Thompson1}

\end{document}


{
		\title{\bf \sf Variable Selection via Thompson Sampling}
		\author{Yi Liu\footnote{Yi Liu is a \yi{$4^{th}$} year PhD student at the Department of Statistics at the University of Chicago}\, and Veronika Ro\v{c}kov\'{a}\footnote{Veronika Rockova is Assistant Professor in Econometrics and Statistics at the Booth School of Business of the University of Chicago. The author gratefully acknowledges the support from the James S. Kemper Foundation Research Fund at the Booth School of Business. } }
		\maketitle
	} 
	\appendix
	
	\begin{center}
		{\Huge \bf Supplementary Material}
	\end{center}
	\section{Proof of Theorem \ref{the:corr}}\label{sec:proofs}
	
	We will denote with $\mS^\star=\arg\max_{\mS}r^C_{\bm\theta^\star}(\mS)$ the optimal model where 
	$$
	r^C_{\bm\theta^\star}(\mS)=\sum_{i\in \mS}\left[\theta^\star_i(\mS)\log\left(\frac{C+1}{C}\right)-\log\left(\frac{1}{C}\right)\right].
	$$
	First, we define the reward gap  of a set of arms $\mS_t$ as
	$$
	\Delta_{\mS_t} = \E[r_{\bm\theta^\star}^C(\mS^\star) - r_{\bm\theta}^C(\mS_t)].
	$$
	and write the expected regret in  \eqref{eq:regret}  as 
	$
	Reg(T)=\E \sum_{t=1}^T \Delta_{\mS_t}.
	$
	Before proceeding, we need to introduce some notation. We denote with $N_i(t)= \sum_{k< t} \mathbb{I}(i\in \mS_k )$  the number of times an arm $i$ has been pulled up to time $t$. 
	Next, 
	\begin{equation}\label{eq:hat}
	\hat{\mu}_i(t) =\frac{a_i(t)-1}{N_i(t)}= \frac{1}{N_i(t)}\sum_{k<t}\mathbb{I}(i\in \mS_k)\gamma_{i}^t
	\end{equation}
	denotes the empirical mean of an arm $i$, i.e. the proportion of times an arm $i$ has yielded a reward when pulled, i.e. $\gamma_{i}^t =1$ when $i\in \mS_t$.
	We will be using the following usual Chernoff-Hoeffding bounds which we state without a proof.
	\begin{lemma}[Chernoff-Hoeffding Bound]
		\label{thm:CHbound}
		Let $X_1,..., X_n$ be independent Bernoulli random variables with $\EE[X_i] = p_i$ and denote with $X =\frac{1}{n}\sum_{i=1}^n = X_i$ and $\mu =\frac{1}{n}\sum_{i=1}^np_i $. Then, for any $0 < \lambda < 1-\mu $, we have
		$$\Prob(X \geq \mu + \lambda ) \leq  \exp\{-nd(\mu + \lambda, \mu)\},$$
		and, for any $0 < \lambda < \mu$,
		$$\Prob(X \leq \mu - \lambda ) \leq \exp\{-nd(\mu - \lambda, \mu)\},$$
		where $d(a, b) = a \ln a/b + (1-a) \ln (1-a)/(1-b)$.
	\end{lemma}
	
	{Similarly as in other regret bound proofs \citep{komiyama_optimal_2015, wang2018thompson}
		we will bound the expected regret separately on the intersection of combinations of the following events: 
		\begin{align*}
		\mathcal{A}(t)&=\{\mS_t \neq \mS^*\},\\
		\mathcal{B}(t) &=\left\{\exists i \in  \mS^*\,\, s.t.\,\, \hat{\mu}_i(t)<0.5+\frac{\alpha }{2} \text{ or } \exists i \in \mS_t \backslash \mS^* \,\, s.t.\,\,\hat{\mu}_i(t)>0.5-\frac{\alpha }{2} \right\},\\
		\mathcal{C}(t)&= \left\{ \exists i \in \mS^*\,\, s.t.\,\, \hat{\mu}_i(t) -\theta_i(t) > \frac{\alpha }{2} \text{ or }   \exists i \in \mS_t\backslash \mS^*\,\, s.t.\,\,  \theta_i(t)-\hat{\mu}_i(t) > \frac{\alpha }{2} \right\},\\
		\mathcal{D}(t) &= 
		\bigcap_{i\in\mS_t}\left\{N_i(t)>  \frac{8\log T}{\alpha ^2}\right\},
		\end{align*} 
		where $\alpha$ occurred in Assumption \ref{assumption_identifiability}. 
	}
	First, we focus on the following term
	\begin{equation}\label{term1}
	Reg_1(T)=\sum_{t=1}^T\E[\Delta_{\mS_t}\times\mathbb{I}(\mathcal{A}(t)\cap \mathcal{B}(t))].
	\end{equation}
	The following Lemma \label{lemma:R1}  finds  an upper bound on $Reg_1(T)$:
	\begin{lemma}\label{lemma:R1}
		Under the Assumption 1, the TVS sampling policy in Table \ref{alg:vanillathompson} with   $C=(\sqrt{5}-1)/2$  yields, for $\Delta_{max}=\max_{\mS}\Delta_{\mS}$
		\begin{align}
		Reg_1(T)
		\leq&\, \Delta_{\max}\,  \sum_{i \in \mS^*} \EE\left(\sum_{t=1}^T\mathbb{I}\left\{ i \in \mS_t, \hat{\mu}_i(t) < 0.5 + \frac{\alpha}{2} \right\}\right) \\
		&+ \Delta_{\max} \sum_{i \notin \mS^*} \EE\left(\sum_{t=1}^T\mathbb{I}\left\{ i \in \mS_t, \hat{\mu}_i(t) > 0.5 - \frac{\alpha}{2}  \right\}\right) \label{eq:second_one} \\
		\leq& \Delta_{\max}\times p\left(1 + \frac{4}{\alpha_{\max}^2} \right).
		\end{align}
	\end{lemma}
	\begin{proof}
		We will first prove  that
		\begin{equation}\label{eq:term1}
		\EE\left(\sum_{t=1}^T\mathbb{I}\left\{ i \in \mS_t, \hat{\mu}_i(t) < 0.5 + \frac{\alpha}{2} \right\}\right) \leq 1 + \frac{4}{\alpha_{\max}^2}.
		\end{equation}
		The second inequality for the term in \eqref{eq:second_one} can be obtained analogously. With $\theta_i(\mS_t)$ as in   \eqref{eq:probs_S}  we denote with 
		$$
		\bar\theta_i(T)=\frac{1}{N_i(T)}\sum_{t=1}^T\mathbb{I}(i\in\mS_t)\theta_i(\mS_t)
		$$
		and with $d(a,b)=a\log[a/b]+(1-a)\log[(1-a)/(1-b)]$. Note that for any $b\in(0,1)$  both  functions $d(x,b)$ and $d(b,x)$ are monotone increasing on $[b,1]$ and, at the same time, monotone decreasing on $[0,b]$.
		Under the Assumption 1 and for $i\in\mS^\star$  we have $\bar\theta_i(T)>0.5+\alpha$ and thereby 
		$$
		d(0.5+\alpha/2,\bar\theta_i(T))>d(0.5+\alpha/2,0.5+\alpha). 
		$$
		Similarly as in the proof of Lemma 3 in Wang and Chen (2018), we denote with $\tau_1, \tau_2,...$  the times such that $i\in \mS_t$,  define $\tau_0=0$
		and write
		\begin{align*}
		\EE\left(\sum_{t=1}^T\mathbb{I}\{ i \in \mS_t, \hat{\mu}_i(t) < 0.5+\alpha/2\}\right) &\leq 1 + \sum_{w = 0}^T \Prob(\hat{\mu}_i(\tau_w) < 0.5+\alpha/2, N_i(t) = w )\\
		&\leq 1 + \sum_{w = 0}^T \exp\left(-w\,d\left(0.5 + \alpha/2, \bar\theta(T)\right)\right) \\
		&\leq 1+ \sum_{w = 0}^T \exp\left(-w\, d(0.5+\alpha/2, 0.5+ \alpha)\right)\\
		&\leq  1+ \sum_{w = 0}^\infty \exp\left(-w\, d(0.5+\alpha/2, 0.5+ \alpha)\right)\\
		&\leq 1 + \frac{\exp\left(- d(0.5+\alpha/2, 0.5+ \alpha)\right)}{1-\exp\left(- d(0.5+\alpha/2, 0.5+\alpha)\right)}\\
		&\leq 1+ \frac{4}{\alpha^2}. 
		\end{align*}	
		This concludes the proof of  \eqref{eq:term1}. The second term can be bounded analogously, which concludes the proof of the Lemma. \hfill $\square$
	\end{proof}
	
	\smallskip
	
	Next, we  focus on the following term
	\begin{equation}\label{term2}
	Reg_2(T)=\sum_{t=1}^T\E[\Delta_{\mS_t}\times\mathbb{I}(\mathcal{A}(t)\cap \mathcal{B}^c(t)\cap\mathcal{C}(t)\cap \mathcal D(t))].
	\end{equation}
	To bound this term, we will be using the following Lemma (Lemma 4 from \cite{wang2018thompson}) which we, again, state without a proof.
	
	\begin{lemma} \label{lemma:theta} 
		Denote with $\theta_i(t)$ the mean reward for an arm $i$ sampled from  $\mathcal B(a_i(t),b_i(t))$ during the step C1 in Table \ref{alg:vanillathompson}.
		Using the TVS algorithm from Table \ref{alg:vanillathompson},  we have the following two inequalities for any base arm $i$:
		$$\Prob\left[\theta_i(t) - \hat{\mu}_i(t) > \sqrt{\frac{2\log T}{N_i(t)}}\right]\leq \frac{1}{T}, $$
		$$\Prob\left[ \hat{\mu}_i(t)- \theta_i(t) > \sqrt{\frac{2\log T}{N_i(t)}}\right]\leq \frac{1}{T}. $$
	\end{lemma}
	\proof The proof relies on the observation that  $\theta_i(t)$'s  only depend on values $a_i(t)$ and $b_i(t)$. The proof is then the same as in Lemma 4 in \cite{wang2018thompson}.
	\medskip 
	
	The following lemma bounds the regret term  \eqref{term2}.

	\begin{lemma} \label{lemma:R2} Using the TVS algorithm from Table \ref{alg:vanillathompson},  we have the following bound:
		$$
		Reg_2(T)=\sum_{t=1}^T\E[\Delta_{\mS_t}\times\mathbb{I}(\mathcal{A}(t)\cap \mathcal{B}^c(t)\cap\mathcal{C}(t)\cap \mathcal D(t))]\leq \Delta_{\max}\times p.
		$$
	\end{lemma}
	\begin{proof}
		On the event $\mathcal D(t)$, we have $N_i(t)> \frac{8\log T}{\alpha ^2}$ and thereby
		$\frac{\alpha}{2}>\sqrt{2\frac{\log T}{N_i(t)}}$. The set $\mathcal{B}^c(t)\cap \mathcal{C}(t)\cap \mathcal D(t)$ is then subsumed within 
		$$
		\left\{ \exists i \in \mS_t\backslash \mS^* :\theta_i(t) - \hat{\mu}_i(t) > \sqrt{\frac{2\log T}{N_i(t)}} \text{ or }\exists i \in \mS_t\cap \mS^*: \hat{\mu}_i(t)- \theta_i(t) > \sqrt{\frac{2\log T}{N_i(t)}} \right\}. 
		$$ 
		We can then directly apply Lemma \ref{lemma:theta} to write
		\begin{align*}
		\sum_{t=1}^T \Delta_{\mS_t}\P\left(\mathcal{A}(t) \cap \mathcal{B}^c(t)\cap \mathcal{C}(t)\cap \mathcal{D}(t)\right)&\leq \Delta_{\max} \sum_{t=1}^T \P\left( \mathcal{B}^c(t)\cap \mathcal{C}(t)\cap\mathcal{D}(t)\right)\\
		&\leq \Delta_{\max} \sum_{t=1}^T p/T.\quad\quad\quad\quad\quad\quad \square
		\end{align*}
	\end{proof}
	
	Finally, we focus on the following term

	\begin{equation}\label{eq:reg3}
	Reg_3 = \sum_{t=1}^T\EE \left(\Delta_{\mS_t} \times \mathbb{I}\left(\mathcal{A}(t) \cap\mathcal{B}^c(t)\cap \mathcal{C}(t) \cap \mathcal{D}^c(t)\right)\right). 
	\end{equation}

	\begin{lemma}
		\label{lemma:full_bernoulli_inequality}
		Let $i \in \mS^*$ and let $f^*_i(j,s)$ be the probability that after $j$ pulls of an arm $i$, $s$ of those  pulls result in a reward. If $s \leq \lfloor 0.5 j\rfloor$ , then
		\begin{equation}
		f^*_i(j,s) \leq \binom{j}{s}(0.5+ \alpha)^s (0.5-\alpha)^{j-s}
		\end{equation}
	\end{lemma}
	
	\begin{proof}
		We denote with $\tau_j^i$  the $j^{th}$ time such that  the arm $i$ has been pulled (i.e. $\theta_i(t)>0.5$).
		We denote the probability of yielding a reward at time $\tau_j$ as $p_j = \Prob\left(\gamma^{\tau_j}_i = 1\mid \mS_{\tau_j} \right)$
		and, for a given $j$ and $s$  write $ f^*_{j,s}(p_1,\cdots,p_j):=f^*_{i}(j,s)$.
		Since we are studying one particular arm, we have dropped the subscript $i$ without any loss of generality.
		Consider now a vector of binary indicators $\bm{B} = (B_1, B_2, B_3, \cdots, B_j)'\in\{0,1\}^j$ where $B_k=\gamma^{\tau_k}_i\in \{0,1\}$ for whether or not the $k^{th}$ pull of an arm $i$ has yielded a reward. Denoting $\mid\bm B\mid=\sum_{j=1}^pB_k$, we can write 
		$$
		f^*_{j,s}(p_1 ,\cdots, p_j) = \sum_{\bm{B}: \mid \bm B\mid = s} \prod_{l=1}^j p_l^{B_l}(1-p_l)^{1-B_l}.
		$$
		We want to show that $\bm p^\star=(p_1^\star,\dots,p_j^\star)'=\arg\max f^*_{j,s}(p_1 ,\cdots, p_j)$  when $p^\star_i=0.5+\alpha$ for all $1\leq i\leq j$.
		First, we notice is that this is a multi-linear polynomial in the sense that $\frac{\partial f^*_{j,s}(p_1 ,\cdots, p_j)}{\partial p_k} $ is independent of $p_k$. Keeping every other coordinate constant,  the value $p_k^\star$ that maximizes $f^*_{j,s}(\cdot)$ in the $k^{th}$ direction has to be either $0.5+\alpha$ or $1$. The vector $(p^\star_1, \cdots, p^\star_j)'$ maximizing  $f^*_{j,s}(p_1 ,\cdots, p_j)$ will thus have each coordinate $p^*_k  $ either equal to $1$ or $ 0.5+\alpha$.  Let $r\in \mathbb{N} \cup \{0\}$ be the number of coordinates $k$ for which  $p^\star_k = 1$ and $j-r$ be the number of coordinates $k$ for which $p^\star_k = 0.5+\alpha$ (notice that $r\leq s$). Since  $f^*_{j,s}(p_1 ,\cdots, p_j)$ is  a symmetric polynomial (i.e. the value of the function is not affected by a permutation of its argument) we assume, without loss of generality,   that $p_1^\star= p_2^\star= \cdots =  p_r^\star =1$ and $p_{r+1}^\star = p_{r+2}^\star= \cdots =p_{j}^\star = 0.5+\alpha$. In this case, we have the constraint on the binary indicators $\bm B$ where the first $r$ indices have to be $1$ and the remaining  $s-r$ $1'$s can  be anywhere between the  index $r+1$ and $j$ ($j-r$ indices). Therefore, we have 
		$$ f^*_{j,s,r}(\bm p^\star) = \binom{j-r}{s-r}(0.5 +\alpha)^{s-r}(0.5-\alpha)^{j-s}.$$
		It is sufficient to prove that this function is maximized at $r=0$. We have
		\begin{align*}
		\frac{f^*_{j,s,r+1}(\bm p^\star)}{f^*_{j,s,r}(\bm p^\star)} &= \frac{ \binom{j-r-1}{s-r-1}(0.5 +\alpha)^{s-r-1}(0.5-\alpha)^{j-s}}{\binom{j-r}{s-r}(0.5 +\alpha)^{s-r}(0.5-\alpha)^{j-s}} =\frac{s-r}{(j-r)(1/2 +\alpha)}\\
		&\leq \frac{j/2-r}{(j-r)(1/2 +\alpha)} \quad\text{(using the assumption $s \leq \lfloor  j/2\rfloor $)}\\
		&=1 - \frac{(1/2-\alpha)r + \alpha j}{(j-r)(1/2+\alpha)}< 1
		\end{align*}
		since $1/2-\alpha \geq 0$ and $\alpha >0$. Since this is true for all $r$, the function $\frac{f^*_{j,s,r+1}(p_1 ,\cdots, p_j)}{f^*_{j,s,r}(p_1 ,\cdots, p_j)}$ is maximized at $r=0$. This concludes the proof.  \hfill $\square$

	\end{proof}
	
	\begin{lemma} 
		\label{the:ineqn}
		Let $i \in \mS^*$ and let $\tau_j^i$ be the $j^{th}$ time such that $\theta_i(t)>0.5$.  Suppose that Assumption 1 is true, then   the TVS Algorithm 1 with $C=(\sqrt{5}-1)/2$ satisfies
		\begin{equation}
		\EE\left[\tau_{j+1}- \tau_j \right] \leq \begin{cases}
		4 + \frac{1}{\alpha} \quad \text{when}\quad j \leq \frac{8}{\alpha}\\
		1 + \frac{1}{\e^{\alpha^2j/4}-1}+ \e^{-\alpha^2j/2}\left(C_1+C_2\frac{1-2\alpha}{4\alpha^2(j+1)}\right) \quad \text{when}\quad j >\frac{8}{\alpha},
		\end{cases}
		\end{equation}
		where constants $C_1,C_2>0$ are not related to Algorithm 1.
	\end{lemma}
	\begin{proof}
		We denote with $\tau_j^i$  the $j^{th}$ time such that  the arm $i$ has been pulled (i.e. $\theta_i(t)>0.5$).
		First, we consider the time interval $[\tau_j^i,\tau_{j+1}^i)$. For any $t\in[\tau_j^i,\tau_{j+1}^i)$ we know 
		that the arm $i$ has been played $j$ times and, thereby,  $\theta_i(t)$ comes from a beta distribution
		$$
		\theta_i(t)\sim  \mathcal{B}\left[a_i(t), b_i(t) \right],
		$$
		where $j = a_i(t) + b_i(t) -2$. The parameters of the beta distribution are only updated if the arm $i$ is pulled and the distribution thus does not change until we reach the iteration $\tau_{j+1}$. Therefore, given $\hat{\mu}_i(\tau_{j}^i)$ the expected difference between $\tau_{j+1}^i - \tau_{j}^i$ has a geometric distribution with an expectation  
		$$ 
		\EE\left[\tau_{j+1} - \tau_{j} \,\mid \, \hat{\mu}_i({\tau_{j}^i})\right]=\frac{1}{\Prob(B_{ij}> 0.5)}=\frac{1}{p_{i,j}(0.5)} 
		$$
		where $B_{ij} \sim \mathcal{B}\left[a_i(\tau_{j+1}), b_i(\tau_{j+1}) \right]$. We let $F_{n,p}(\cdot)$ and  $f_{n,p}(\cdot)$ denote the cumulative distribution function (CDF) and the probability density function  of a Binomial distribution with parameters $(n,p)$.
		We now recall the following  fact  (see e.g. Fact 3 in \cite{agrawal_analysis_2012}) about the CDF  $F_{\alpha, \beta}^{\text {beta}}(x)$
		of a beta distribution with parameters $(\alpha,\beta)$. We have the following identity which links the CDF of a beta distribution and  a CDF of a binomial distribution:
		\begin{equation}\label{eq:identity}
		F_{\alpha, \beta}^{\text {beta}}(y)=1-F_{\alpha+\beta-1, y}(\alpha-1)
		\end{equation}
		for all positive integers $\alpha, \beta$. 
		Let $f^*_i(j,s)$ be the probability that after $j$ pulls of an arm $i$, $s$ out of those $j$ pulls result in a reward. Here, we have the following relationship
		$a_i(\tau_{j}) = s + 1  $ and
		$b_i(\tau_{j}) = j + 1 - s $. Using the identity \eqref{eq:identity} and given $s$ successes among $j$ pulls, we can write 
		$p_{i,j}(0.5)=F_{s+1,j-s+1}^{\text {beta}}(0.5)=1-F_{j+1, 0.5}(s)$ and thereby
		\begin{equation}\label{eq:conundrum}
		\EE\left[\frac{1}{p_{i,j}(0.5)}\right] = \sum_{s=0}^j \frac{f^*_i(j,s)}{F_{j+1,0.5}(s)}.
		\end{equation}
		
		First, we consider the case when $j \leq \frac{8}{\alpha}$.
		In the following calculations, we will use the result from Lemma \ref{lemma:full_bernoulli_inequality}. Let $ p_{\max} = \alpha+0.5$ and $R = \frac{p_{\max}}{1-p_{max}}$. Using the fact  $F_{j+1, 0.5}(s)\geq 0.5 F_{j, 0.5}(s)$ and $F_{j, 0.5}(s)\geq1/2$ when $s\geq\lceil j/2\rceil$ (since  the median of Binomial distribution with parameters $(j,1/2)$ is either $\lfloor j/2 \rfloor$ or $ \lceil j/2 \rceil$) we have 
		\begin{align} 
		\EE\left[\frac{1}{p_{i,j}(0.5)}\right]
		\leq & 2 \sum_{s=0}^{j} \frac{f^*_i(j,s)}{F_{j, 0.5}(s)} \\
		\leq & 2 \sum_{s=0}^{\lfloor j/2\rfloor} \frac{\binom{j}{s}p_{\max}^{s}(1-p_{\max})^{j-s}}{f_{j, 0.5}(s)}+4 \sum_{s=\lceil j/2\rceil}^{j} f^*_i(j,s)\\
		\leq & 2 \sum_{s=0}^{\lfloor j/2\rfloor} \frac{p_{\max}^{s}(1-p_{\max})^{j-s} }{1/2^{j}}+4\\
		\leq &  2 \frac{\left(1-p_{\max}\right)^{j}}{1/2^{j}} \sum_{s=0}^{\lfloor j/2\rfloor} R^s+4 \\
		\leq & 2\left(\frac{R^{\lfloor  j/2 \rfloor+ 1}-1}{R-1}\right) \frac{\left(1-p_{\max}\right)^{j}}{1/2^{j}}+4\\
		\leq &2\left(\frac{R}{R-1}\right) R^{  j/2} \frac{\left(1-p_{\max}\right)^{j}}{1/2^{j}}+4 \label{eqn:quickbound1}\\
		\leq & \frac{1}{\alpha}\e^{-j\,d(1/2,\, p_{\max})} +4 \label{eqn:quickbound2} \\
		\leq & \frac{1}{\alpha} + 4,
		\end{align}
		where  from \eqref{eqn:quickbound1} to \eqref{eqn:quickbound2} we have used the following two facts. First, using  the definition of $d(\cdot,\cdot)$ in Lemma 4 and the fact that $d(p_1, p_2) > (p_1-p_2)^2$ we obtain
		\begin{align*}
		\frac{\left(1-p_{\max}\right)^{j}}{1/2^{j}} R^{\lfloor j/2\rfloor} &\leq \frac{\left(1-p_{\max}\right)^{j}}{1/2^{j}} R^{ j/2}= \e^{j \log(1-p_{\max}) - j \log(1/2) + j/2 \log(p_{\max}) -j/2 \log(1-p_{\max})}\\
		&= \e^{-j\left\{\frac{1}{2}\log\left(\frac{1}{2}/p_{\max}\right)+ \frac{1}{2}\log\left[\frac{1}{2}/(1-p_{\max})\right]\right\}}= \e^{-j\,d(1/2, \,p_{\max})}\leq \e^{-\alpha^2 j}.
		\end{align*}
		Second, since $p_{\max}= 0.5 + \alpha$, we have $\frac{R}{R-1} = \frac{p_{\max}}{2\alpha}\leq \frac{1}{2\alpha}.$


		\noindent When $j > \frac{8}{\alpha}$, we will divide the sum $\Sigma(0, j) \equiv \sum_{s=0}^j \frac{f^*_i(j,s)}{F_{j+1,0.5}(s)}$ into 
		$4$ pieces and bound each one of them
		\begin{align}
		\Sigma\left(0,\lfloor j/2\rfloor- 1\right)&\leq c_2\left[\e^{-\alpha^2 j}  \frac{1-2\alpha}{4\alpha^2(j+1)}\right] + c_3\,\e^{-2\alpha^{2} j}\label{eq:first_one}\\
		\Sigma\left(\lfloor j/2\rfloor,\lfloor j/2\rfloor\right) &\leq 3\, \e^{-\alpha^2 j}\label{eq:second_one}\\
		\Sigma\left(\lceil j/2 \rceil, \left\lfloor (1/2+\alpha/2) j\right\rfloor\right) &\leq c_3\left(\e^{-\alpha^2 j / 2}\right)\label{eq:third_one}\\
		\Sigma\left(\left\lceil (1/2+\alpha/2)j\right\rceil, j \right)&\leq 1+\frac{1}{\e^{\alpha^2j/4}-1}\label{eq:last_one},
		\end{align}
		where $c_2>0$ and $c_3>0$ are constants unrelated to Algorithm \ref{alg:vanillathompson}.
		This will complete the proof. We now prove the bounds in the last display. 
		We start with the first inequality in \eqref{eq:first_one}. 
		When $s \leq (j+1)/2-\sqrt{(j+1)/4}$, we use  the following bound for the Binomial CDF (\cite{jevrabek2004dual}) 
		$$
		F_{j+1, 0.5}(s)\geq \frac{1}{c_2}\left[\frac{j+1-s}{j+1-2s} \left( \begin{array}{c}{j+1} \\ {s}\end{array}\right) \frac{1}{2^{j+1}}\right],
		$$
		for some $c_2 >0$. When $s \geq (j+1)/2-\sqrt{(j+1)/4}$ we use that fact that, for some $c_3>0$,
		$$
		F_{j+1, 0.5}(s)\geq \frac{1}{c_3} >0.
		$$
		Altogether, we arrive at the following bound (using again Lemma \ref{lemma:full_bernoulli_inequality} and denoting  $p_{\max}=  1/2+ \alpha $ and $R =\frac{p_{\max}}{1-p_{max}}$)
		\begin{align}
		\Sigma\left(0,\lfloor j/2\rfloor- 1\right) &\leq c_2\sum_{s=0}^{\lceil (j+1)/2-\sqrt{(j+1)/4}\rceil}\frac{f^*_i(j,s)}{\frac{j+1-s}{j+1-2s} \left( \begin{array}{c}{j+1} \\ {s}\end{array}\right) \frac{1}{2^{j+1}}} +  c_3\sum_{s=\lceil (j+1)/2-\sqrt{(j+1)/4}\rceil+1}^{\lfloor j/2\rfloor- 1}  f^*_i(j,s)\nonumber\\
		&\leq c_2\sum_{s=0}^{\lfloor j/2\rfloor- 1}\frac{f^*_i(j,s)}{\frac{j+1-s}{j+1-2s} \left( \begin{array}{c}{j+1} \\ {s}\end{array}\right) \frac{1}{2^{j+1}}} +  c_3\sum_{s=0}^{\lfloor j/2\rfloor- 1}  f^*_i(j,s)\nonumber\\
		&\leq c_2 \left(\frac{\left(1-p_{\max}\right)^{j}}{1/2^{j+1}} \sum_{s=0}^{\lfloor j/2\rfloor- 1}\left(1-\frac{2s}{j+1}\right) R^{s}\right) +  c_3 \sum_{s=0}^{\lfloor  j/2\rfloor- 1}  f^*_i(j,s).\label{eq_two_terms}
		\end{align}
		Now we bound the first term in  \eqref{eq_two_terms} to obtain
		\begin{align*}
		&\frac{\left(1-p_{\max}\right)^{j}}{1/2^{j+1}} \sum_{s=0}^{\lfloor  j/2\rfloor- 1}\left(1-\frac{2s}{j+1}\right) R^{s} \\
		&=\frac{\left(1-p_{\max}\right)^{j}}{1/2^{j+1}} \left\{\frac{R^{\lfloor j/2\rfloor}-1}{R-1} -\frac{2}{j+1}\left[\frac{\left(\lfloor  j/2 \rfloor-1) R^{\lfloor  j/2\rfloor}\right.}{R-1}-\frac{R^{\lfloor j/2\rfloor}-R}{(R-1)^{2}}\right]\right\}\\
		&\leq\frac{\left(1-p_{\max}\right)^{j}}{1/2^{j+1}} \left\{\frac{R^{\lfloor j/2\rfloor}}{R-1} -\frac{2}{j+1}\left[\frac{\left(\lfloor  j/2 \rfloor-1) R^{\lfloor  j/2\rfloor}\right.}{R-1}-\frac{R^{\lfloor j/2\rfloor}}{(R-1)^{2}}\right]\right\}\\
		&\leq \frac{\left(1-p_{\max}\right)^{j}}{1/2^{j+1}} \left[\frac{2}{j+1} \frac{R^{\lfloor j/2\rfloor}}{(R-1)^{2}} +\frac{2[(j+1)/2-\lfloor  j/2\rfloor+ 1]}{j+1} \frac{R^{\lfloor j/2\rfloor}}{R-1}\right]\\
		&\leq \frac{\left(1-p_{\max}\right)^{j}}{1/2^{j+1}}\frac{6}{j+1} \frac{R^{\lfloor  j/2\rfloor+1}}{(R-1)^{2}}\\
		&\leq \e^{-\alpha^2 j} \frac{12}{j+1} \frac{R}{(R-1)^{2}},
		\end{align*}
		where we have used the following facts. First,
		for any $x >1$	we have
		$$
		\sum_{s =0}^n s\, x^s = \frac{n x^{n+2}-(n+1) x^{n+1}+x}{(1-x)^{2}}= \frac{nx^{n+1}}{x-1} -\frac{x^{n+1}-x}{(x-1)^2}.
		$$
		Second,
		$j/2+1/2-\lfloor  j/2\rfloor+ 1<3$ and (similarly as before)
		$$
		\frac{\left(1-p_{\max}\right)^{j}}{1/2^{j}} R^{\lfloor j/2\rfloor} \leq \e^{-j\,d(1/2,\, p_{\max})}\leq \e^{-\alpha^2 j}.
		$$
		Finally, since $R/(R-1) \leq \frac{1}{2\alpha} $ and $1/(R-1) = \frac{1-2\alpha}{4\alpha}$, we have
		$$
		\frac{1}{(j+1)/4} \frac{R}{(R-1)^{2}} \leq   
		\frac{1-2\alpha}{4\alpha^2(j+1)}.
		$$
		For the second term in \eqref{eq_two_terms}, we notice that $\sum_{s=0}^{\lfloor  j/2\rfloor- 1}  f^*_i(j,s)$ is equal to the probability that the total number of successes is less than $\lfloor  j/2\rfloor- 1$. Here, we invoke Lemma \ref{thm:CHbound} and note that the success probability of each pull is always greater than $1/2 + \alpha$ and the difference between the average success probability  over the $j$ pulls and $1/2$ is thereby greater than $\alpha$. Hence, $\sum_{s=0}^{\lfloor  j/2\rfloor- 1}  f^*_i(j,s) \leq \e^{-2\alpha^{2} j}.$ We put the two terms together to finally obtain
		\begin{align}
		\Sigma(0,\lfloor  j/2\rfloor- 1) &\leq  c_2\left[\e^{-\alpha^2 j}  \frac{1-2\alpha}{4\alpha^2(j+1)}\right]+ c_3\sum_{s=0}^{\lfloor  j/2\rfloor- 1}  f^*_i(j,s)\\
		&\leq  c_2\left[\e^{-\alpha^2 j}  \frac{1-2\alpha}{4\alpha^2(j+1)}\right] + c_3\,\e^{-2\alpha^{2} j}.
		\end{align}

		Next,  to bound the term $\Sigma(\lfloor j/2\rfloor,\lfloor j/2\rfloor)$ in \eqref{eq:second_one}, we use Lemma \ref{lemma:full_bernoulli_inequality} and the fact that 
		$p_{max}>1/2$ to find that for $s=\lfloor j/2\rfloor$
		\begin{align*}
		\Sigma(\lfloor j/2\rfloor,\lfloor j/2\rfloor)=   \frac{  f^*_i(j,s)}{F_{j+1, 0.5}(s)} \leq \frac{ f^*_i(j,s)}{f_{j+1, 0.5}(s)} &\leq 2\,\left(1-\frac{s}{j+1}\right) R^{s}\left( \frac{1- p_{\max}}{1/2}\right)^{j} \\
		&\leq\frac{2}{j+1}\left(\frac{j}{2}+2\right) R^{ j/2} \left(\frac{1- p_{\max}}{1/2}\right)^{j}\\
		&  \leq  \left(1+\frac{3}{j+1}\right) \e^{-\alpha^2 j}\leq 2\,  \e^{-\alpha^2 j},
		\end{align*}
		where we used the assumption $j\geq 1/\alpha>2$.
		
		In order to bound the third term $\Sigma(\lceil  j/2 \rceil, \left\lfloor (\alpha+1)j/2\right\rfloor)$ in \eqref{eq:third_one}, we 
		first  note that if $j \geq \frac{8}{\alpha}> \frac{1}{\alpha} \geq 2$ (our assumption above), we have  $\sqrt{(j+1) /4}>\sqrt{3/4} \geq\sqrt{1/2}>1/2$ and thereby  $(j+1)/2-\sqrt{(j+1)/4}< j/2 \leq\lceil j/2\rceil \leq s$. This implies that  the condition in \cite{jevrabek2004dual} is satisfied and we can apply the bound $F_{j+1, 0.5}(s)\geq \frac{1}{c_3}$.   Then  we have
		\begin{align*}
		\Sigma(\lceil  j/2 \rceil, \left\lfloor (\alpha+1)j/2\right\rfloor)&\leq c_3\left(\sum_{s=\lceil j/2\rceil}^{\left\lfloor (\alpha+1)j/2\right\rfloor} f^*_i(j,s)\right)
		\leq c_3\left(\sum_{s=0}^{\left\lfloor (\alpha+1)j/2\right\rfloor} f^*_i(j,s)\right)\\
		&\leq c_3\left(\e^{-\alpha^2 j / 2}\right),
		\end{align*}
		where the last inequality stems from the Chernoff-Hoeffding inequality in Lemma \ref{thm:CHbound} and Assumption 1 which guarantees that the success probability of each pull is greater than $1/2 + \alpha$. This implies that the difference between the average probability of success over all the $j$ pulls and $1/2+\alpha/2$ is greater than $\alpha/2$.

		Finally, to bound the term $\Sigma(\left\lceil  (\alpha+1)j/2\right\rceil,j)$  in \eqref{eq:last_one}  we use the Hoeffding inequality in Lemma \ref{thm:CHbound}  with $\lambda=  (\alpha+1)j/[2(j+1)] -1/2\leq s/(j+1)-1/2$ to  find  (for a r.v. $X\sim Bin(j+1,1/2)$) that
		\begin{align*}
		F_{j+1, 0.5}(s)&\geq 1-\P\left(\frac{X}{j+1}-\frac{1}{2}>\lambda\right)\geq 1-\e^{-2(j+1)\lambda^2}
		= 1- \e^{(-\frac{j\alpha^2}{2} + \frac{j\alpha^2}{2(j+1)}  + \alpha \frac{j}{j+1} - \frac{1}{2(j+1)})}\\
		&\geq 1-\e^{-\alpha^2j/2},
		\end{align*}
		where we used the fact that  $1/j\leq \alpha/8$ and thereby  $2\alpha \geq   \frac{j\alpha^2}{2(j+1)}+\alpha \frac{j}{j+1} - \frac{1}{2(j+1)}$.
		Finally, we write
		\begin{align*}
		\Sigma(\left\lceil (\alpha+1)j/2\right\rceil, j )\leq \sum_{s=\left\lceil (\alpha+1)j/2\right\rceil}^{j} \frac{f^*_i(j,s)}{F_{j+1, 0.5}(s)}
		\leq 	\frac{1}{1-\e^{-\alpha^2j/4}}=1+\frac{1}{\e^{\alpha^2j/4}-1}.
		\end{align*}
		Now, denoting $C_1 =2+2\,c_3$ and $C_2=c_2$ we get the statement in the Lemma. This concludes our proof. \hfill $\square$
	\end{proof}
	
	\medskip
	
	Using Lemma \ref{the:ineqn}, we can achieve a similar bound in Lemma 6 of \cite{wang2018thompson},

	\begin{lemma} \label{lemma:quick_inequ} 
		Under Assumption 1, the TVS  Algorithm \ref{alg:vanillathompson} with $C=(\sqrt{5}-1)/2$  satisfies the following property. For any signal arm $i \in \mS^*$, the expected number of total pulls before the given arm $i$ is pulled $\frac{8\log(T)}{\alpha^2}$  times is bounded by 
		$$
		\left\lceil\frac{8\log(T)}{\alpha^2}\right\rceil+ \left\lceil\frac{8}{\alpha}\right\rceil\left(3+\frac{1}{\alpha}\right) +\wt C\,\frac{\e^{-4\alpha}}{1 - \e^{-\alpha^2 /2}} +\frac{8}{\alpha^2} \frac{1}{\e^{2\alpha}-1} + \frac{\e^{-1}}{1 - \e^{-\alpha/8}}, 
		$$
		where $\wt C= C_1+C_2\frac{1-2\alpha}{32\,\alpha}$ for some $C_1>0$ and $C_2>0$ not related to the Algorithm \ref{alg:vanillathompson}. 
	\end{lemma}
	\begin{proof}
		We use the notation from Lemma  \ref{the:ineqn}, where  $\tau_j^i$  is the  time when  the arm $i$ has been pulled for the $j^{th}$ time. 
		Denoting with $\wt T=\lfloor\frac{8\log(T)}{\alpha^2}\rfloor$, we want to find an upper bound for $\EE [\tau_{\wt T}^i]$. 
		We can write 
		$$
		\EE\left[\tau_{\wt T}^i \right] = \sum_{j=0}^{\wt T} \EE\left[\tau_{j+1}^i - \tau_{j}^i \right]   = \sum_{j=0}^{\wt T} \EE\left(\frac{1}{p_{i,j}(0.5)}\right) 
		$$
		and using Lemma \ref{the:ineqn} we obtain
		\begin{equation}\label{eq:EEE1}
		\EE\left[\tau_{\wt T}^i \right]  \leq \sum_{j=0}^{\lfloor\frac{8}{\alpha}\rfloor} \left(4+\frac{1}{\alpha}\right)+ 
		\sum_{j = \lfloor\frac{8}{\alpha}\rfloor+1} ^{\wt T} \left[1 + \frac{1}{\e^{\alpha^2j/4}-1}+ \e^{-\alpha^2j/2}\left(C_1+C_2\frac{1-2\alpha}{4\alpha^2(j+1)}\right)\right]. 
		\end{equation}
		First, we note that	
		$\sum_{j=0}^{\lfloor\frac{8}{\alpha}\rfloor}\left(4+\frac{1}{\alpha}\right) + \sum_{j = \lfloor\frac{8}{\alpha}\rfloor+1}^{\wt T} 1  \leq  \lceil\frac{8\log(T)}{\alpha^2}\rceil+ \lceil\frac{8}{\alpha}\rceil(3+\frac{1}{\alpha})$.
		Next, we write
		$$ 
		\sum_{j = \lceil \frac{8}{\alpha}\rceil+1}^{\wt T} \e^{ - \frac{j\alpha^2}{2}} \leq  
		\frac{\e^{-4\alpha}}{1 - \e^{-\alpha^2 /2}} 
		$$
		and	
		$$ \sum_{j =\lceil \frac{8}{\alpha}\rceil +1}^{\wt T}
		{\e^{- \frac{j\alpha^2}{2}}}\left(C_1+C_2\frac{1-2\alpha}{4\alpha^2(j+1)}\right) <
		\sum_{j =\lceil \frac{8}{\alpha}\rceil +1 }^{\wt T} {\e^{- \frac{j\alpha^2}{2}} }\left(C_1+C_2\frac{1-2\alpha}{32\,\alpha}\right)\leq 
		\wt C\frac{\e^{-4\alpha}}{1 - \e^{-\alpha^2 /2}}.
		$$
		Finally, we use the fact that $\frac{1}{\e^{x} -1 } \leq \e^{-x/2}$ for $x \geq 1$ to obtain
		\begin{align*}
		\sum_{j =\lceil \frac{8}{\alpha}\rceil +1} ^{\wt T} \frac{1}{\e^{\frac{j\alpha^2}{4}}-1} &\leq  \sum_{j = \lceil \frac{8}{\alpha}\rceil +1}^{\lfloor\frac{8}{\alpha^2}\rfloor} \frac{1}{\e^{\frac{j\alpha^2}{4}}-1} + \sum_{j = \lceil\frac{8}{\alpha^2}\rceil}^{\lfloor\frac{8\log(T)}{\alpha^2}\rfloor} \frac{1}{\e^{\frac{j\alpha^2}{4}}-1}  \\
		&\leq \frac{8}{\alpha^2} \frac{1}{\e^{2\alpha}-1} + \sum_{j = \lceil\frac{8}{\alpha^2}\rceil}^{\lfloor\frac{8\log(T)}{\alpha^2}\rfloor}\e^{-\alpha^2 j/8}\leq \frac{8}{\alpha^2} \frac{1}{\e^{2\alpha}-1} + \frac{\e^{-1}}{1 - \e^{-\alpha^2/8}}\hspace{2cm}\square
		\end{align*}
		
	\end{proof}
	
	\medskip
	
	Using these lemmas we can prove the following lemma about $Reg_3(T)$.

	\begin{lemma} \label{lemma:R3}
		We denote with $Reg_3(T)$ the regret term in \eqref{eq:reg3}. Under Assumption 1, the TVS  Algorithm \ref{alg:vanillathompson} with $C=(\sqrt{5}-1)/2$  yields
		\begin{align*}
		Reg_3(T) &\leq  \Delta_{\max} \left\{\frac{8 p\log(T)}{\alpha^2} + \wt C\, q^*\frac{\e^{-4\alpha}}{1 - \e^{-\alpha^2 /2}} + q^*\left[\frac{8}{\alpha^2} \frac{1}{\e^{2\alpha}-1} + \frac{\e^{-1}}{1 - \e^{-\alpha/8}}+\left\lceil\frac{8}{\alpha}\right\rceil\left(3+\frac{1}{\alpha}\right)\right] \right\},
		\end{align*}
		where $\wt C =  C_1+C_2\frac{1-2\alpha}{32\,\alpha}$ for some $C_1>0$ and $C_2>0$ not related to the Algorithm \ref{alg:vanillathompson}. 
	\end{lemma}
	\begin{proof}
		We start with the following facts  $Reg_3(T)\leq \sum_{t=1}^T\Delta_{\mS_t}\EE \left[ \mathbb{I}\left(\mathcal{A}(t) \cap \mathcal{D}^c(t)\right)\right]$ and 
		$$
		Reg_3(T)\leq \sum_{t=1}^T \sum_{i\in\mS_t}\Delta_{\mS_t}\EE \left[ \mathbb{I}\left(\mathcal{A}(t) \cap\left\{N_i(t)\leq\frac{8\log(T)}{\alpha^2} \right\}\right)\right], 
		$$
		where we used the fact that on the event $\mathcal{D}^c(t)$,  there exists at least one arm $i\in \mS_t$ such that $N_i(t)\leq  8\frac{\log(T)}{\alpha^2}$.  We now decompose the sum above into signal arms and noise arms
		\begin{align}
		Reg_3(T)\leq&  \sum_{t=1}^T \left[\sum_{i\in\mS_t\cap\mS^\star}\Delta_{\mS_t}\EE \left[ \mathbb{I}\left(\mathcal{A}(t) \cap\left\{N_i(t)\leq\frac{8\log(T)}{\alpha^2} \right\}\right)\right]\right]\label{eq:term1}\\
		&+ \sum_{t=1}^T \left[\sum_{i\in\mS_t\backslash\mS^\star}\Delta_{\mS_t}\EE \left[ \mathbb{I}\left(\mathcal{A}(t) \cap\left\{N_i(t)\leq\frac{8\log(T)}{\alpha^2} \right\}\right)\right]\right].\label{eq:term2}
		\end{align}
		If $i \in\mS_t\backslash \mS^\star$, then $\mS_t$ contributes to the regret  but this can {\em only} happen $\frac{8\log(T)}{\alpha^2}$ times so the total regret contribution  of pulling a subset $\mS_t$ including an arm $i$ before $N_i(t) > \frac{8\log(T)}{\alpha^2}$ is bounded by 
		$
		\max_{\mS: i\in\mS}\frac{8\Delta_{\mS}\log(T)}{\alpha^2}.
		$
		There are $p-q^*$ noise arms $i\notin \mS^*$ and the term \eqref{eq:term2} can be thus bounded by $(p-q^*)\Delta_{max}\frac{8\log(T)}{\alpha^2}$.  

		If $i \in \mS^\star\cap\mS_t$, then the arm $i$ contributes to the regret when $\mS_t\neq\mS^\star$. However, by Lemma \ref{the:ineqn} and Lemma \ref{lemma:quick_inequ} we can bound the expected number of pulls of an arm $i$  before $N_i(t)$ reaches $\frac{8\log(T)}{\alpha^2}$. This means that the contribution to the regret when $i\in\mS^\star\cap\mS_t$ is bounded by 
		$\Delta_{\max} \left(	\left\lceil\frac{8\log(T)}{\alpha^2}\right\rceil+ \left\lceil\frac{8}{\alpha}\right\rceil\left(3+\frac{1}{\alpha}\right) +\wt C\,\frac{\e^{-4\alpha}}{1 - \e^{-\alpha^2 /2}} +\frac{8}{\alpha^2} \frac{1}{\e^{2\alpha}-1} + \frac{\e^{-1}}{1 - \e^{-\alpha/8}}\right)$. 
		Because there are $q^\star$ signal arms, we can combine \eqref{eq:term1} and \eqref{eq:term2} to arrive at the bound in the statement of this lemma.\hfill $\square$	
	\end{proof}

	\medskip
	
	We now put the various pieces together to finally  prove Theorem \ref{the:corr}. 
	
	\begin{proof}
		We start by noticing that
		\begin{align*}
		\mathbb{I}\left[\mathcal{A}(t)\right] &= \mathbb{I}[\mathcal{A}(t) \cap\mathcal{B}(t) ] +  \mathbb{I}\left[\mathcal{A}(t) \cap\mathcal{B}^c(t) \right]\\
		& =   \mathbb{I}[\mathcal{A}(t) \cap\mathcal{B}(t)] +  \mathbb{I}\left[\mathcal{A}(t) \cap\mathcal{B}^c(t)\cap \mathcal{C}(t) \right] + \mathbb{I}[\mathcal{A}(t) \cap\mathcal{B}^c(t)\cap \mathcal{C}^c(t)].
		\end{align*}
		Now we note that $\mathbb{I}[\mathcal{A}(t) \cap\mathcal{B}^c(t)\cap \mathcal{C}^c(t)]=0$ because
		$$\mathcal{B}^c(t)\cap \mathcal{C}^c(t)=\left\{ \forall i \in \mS^*\,\,\text{we have}\,\, \theta_i(t) >0.5 \quad \text{and}\quad \forall i \in \mS_t\backslash \mS^*\,\,\text{we have}\,\, \theta_i(t) < 0.5    \right\} = \mathcal{A}^c(t). $$
		Thereby we can write 
		$$
		\mathbb{I}\left[\mathcal{A}(t)\right] {=} \mathbb{I}[\mathcal{A}(t) \cap\mathcal{B}(t)] + 
		\mathbb{I}\left[\mathcal{A}(t) \cap\mathcal{B}^c(t)\cap \mathcal{C}(t)\cap  \mathcal{D}^c(t)  \right]\\
		+  \mathbb{I}\left[\mathcal{A}(t) \cap\mathcal{B}^c(t)\cap \mathcal{C}(t) \cap  \mathcal{D}(t)  \right]
		$$	
		which leads to the following decomposition	
		\begin{align*}Reg(T) &= \sum_{t=1}^T \Delta_{\mS_t}\EE\left[\mathbb{I}\left(\mathcal{A}(t) \right)\right]\\
		&\leq  \sum_{t=1}^T \EE\Bigg[\Delta_{\mS_t}\times \mathbb{I}[\mathcal{A}(t) \cap \mathcal{B}(t)] +
		\Delta_{\mS_t}\times \mathbb{I}\left[\mathcal{A}(t) \cap \mathcal{B}^c(t)\cap \mathcal{C}(t)\cap \mathcal{D}(t)\right]\\
		& \quad +\Delta_{\mS_t}\times\mathbb{I}\left[\mathcal{A}(t) \cap\mathcal{B}^c(t)\cap \mathcal{C}(t) \cap \mathcal{D}^c(t)\right]\Bigg]\\
		&= Reg_1(T) + Reg_2(T) + Reg_3(T).
		\end{align*}
		Now, we bound $Reg_1(T)$, $Reg_2(T)$ and $Reg_3(T)$ with Lemma \ref{lemma:R1},Lemma \ref{lemma:R2}, and Lemma \ref{lemma:R3}. This gives us Theorem \ref{the:corr}. 
	\end{proof}

	\subsection{Proof of Theorem \ref{thm:almostsureconvergence}}\label{sec:proof_thm:almostsureconvergence}
		
		\proof 
		
			Let's define the event $\mathcal{A}(t)=\{\mS_t \neq \mS^*\}$. It is sufficient to show that the probability of infinitely many events $\mA(t)$ occurring is zero, i.e. $\P\left(\lim\sup\limits_{t\rightarrow\infty}\mathcal{A}(t)\right)=0$. This implies that  $\P\left(\lim\inf\limits_{t\rightarrow\infty}\{\mS_t=\mS^*\}\right)=1$.
By the Borel-Cantelli Lemma, it suffices to show that
			\begin{equation}
			\sum_{t=1}^\infty\mathbb{P}(\mathcal{A}(t)) < \infty. 
			\end{equation}
		
For each $i\in\{1,\dots, p\}$ we  define an event 
			$$
			\mathcal{A}_i(t) = \begin{cases}
				\{\theta_i(t) > 0.5\} \quad \text{when} \quad i\in \mS\backslash\mS^*\\
				\{\theta_i(t) < 0.5\} \quad \text{when} \quad i\in \mS^*\\
			\end{cases}.
			$$ 			
			
			Using the fact that \begin{equation}
			\mathcal{A}(t) \subseteq \bigcup_{i=1}^p \mathcal{A}_i(t)
			\end{equation}
			one has 
			$$
			\left(\sum_{t=1}^\infty\mathbb{P}(\mathcal{A}(t))\right) \leq\left(\sum_{t=1}^\infty\sum_{i=1}^p\mathbb{P}(\mathcal{A}_i(t))\right) = 
			\sum_{i=1}^p\sum_{t=1}^\infty\mathbb{P}\left(\mathcal{A}_i(t)\right). 
			$$

We now show that the sum on the right hand side is finite, splitting the arms into the true arms $i\in\mS^*$ and the fake arms $i\in\mS\backslash\mS^*$.

		\begin{lemma}
			\label{lem:almostsureconvergenceinequalityOne}
		 Under assumptions of Theorem \ref{thm:almostsureconvergence} we have, for each $i\in \mS\backslash\mS^*$,
		 $$
		 \sum_{t=1}^\infty\P(\mathcal{A}_i(t)) < 1 + \frac{4}{\alpha^2} + \frac{1}{1-\exp\left(-\alpha^2/8\right)}.
		 $$ 
		\end{lemma}
		\begin{proof}
		
			We can write
			\begin{align*}
			\mathcal{A}_i(t)&=\{\theta_i(t) > 0.5 \}\\
						     &=\{\hat{\mu}_i(t) \geq 0.5-\alpha/2, \theta_i(t)>0.5\}\cup\{\hat{\mu}_i(t) \leq 0.5-\alpha/2, \theta_i(t)>0.5\} \\
						     &=\{\hat{\mu}_i(t) \geq 0.5-\alpha/2, \theta_i(t)>0.5\}\cup\{\hat{\mu}_i(t) \leq 0.5-\alpha/2, \theta_i(t) -\hat{\mu}_i(t) >\alpha/2, \theta_i(t) >0.5\}\\
						     &\subseteq \{\hat{\mu}_i(t) \geq 0.5-\alpha/2, \theta_i(t)>0.5\}\cup\{ \theta_i(t) -\hat{\mu}_i(t) >\alpha/2, \theta_i(t) >0.5\}.
			\end{align*}
		  Since $\{\theta_i(t)>0.5\}=\{i \in \mS_t\}$, we first show that 
		 \begin{equation}
		 \sum_{t = 1}^{\infty} \P\left(i\in\mS_t,\hat{\mu}_i(t) \geq 0.5-\alpha/2\right) \leq 1+ \frac{4}{\alpha^2}
		 \end{equation}
		 
		 We can write, using our Lemma 2 in the Supplement,
		 $$
		  \sum_{t = 1}^{\infty} \P\left(i\in\mS_t,\hat{\mu}_i(t) \geq 0.5-\alpha/2\right)=\lim_{T\rightarrow\infty} \E \sum_{t=1}^{T} \mathbb I\left(i\in\mS_t,\hat{\mu}_i(t) \geq 0.5-\alpha/2\right)\leq 1+\frac{4}{\alpha^2}.
		 $$
		 Recall the definition of $N_i(t)= \sum_{k< t} \mathbb{I}(i\in \mS_k ) =a_i(t) + b_i(t)-2$ and  $\hat\mu_i(t)=\frac{a_i(t)-1}{N_i(t)}$.

		 Now, we want to show that for any fixed $a_i(t)\geq 1$ and $b_i(t)\geq 1$, we have 
		 \begin{align*}
		 \mathbb{P}\left(\theta_i(t) >\hat{\mu}_i(t)+ \alpha/2 \right)&= 1 - F^{\texttt{beta}}_{(a_i(t), b_i(t))}(\hat{\mu}_i(t)+ \alpha/2)\\
		 &= 1 - \left(1 -F_{a_i(t) + b_i(t)-1,\hat{\mu}_i(t)+ \alpha/2 } (a_i(t)-1)\right)\\
		 &\text{(Using equation (11) in the Appendix)}\\
		 &= F_{a_i(t) + b_i(t)-1,\hat{\mu}_i(t)+ \alpha/2 }(\hat{\mu}_i(t)(a_i(t)+b_i(t)-2))\\
		 &\text{(Using the Chernoff-Hoeffding Bound)}\\
		 &\leq \exp(-(a_i(t)+b_i(t)-2)d(\hat{\mu}_i(t)+ \alpha/2,\hat{\mu}_i(t)))\\
		 &\leq \exp(-N_i(t)\alpha^2/8)
		 \end{align*}
		 Then 
		 \begin{align*}
		  \sum_{t = 1}^{\infty} \mathbb{E}\left(\mathbb{I}\left(\{\theta_i(t) -\hat{\mu}_i(t) >\alpha/2,i\in\mS_t\}\right)\right) &\leq \sum_{N_i=0}^\infty \exp(-N_i\alpha^2/8)\\
		  & \leq \frac{1}{1- \exp\left(-\alpha^2/8\right)}.
		 \end{align*}
		 Combining the two cases, we finally obtain
		 $\sum_{t=1}^\infty\mathbb{E}\left(\mathbb{I}(\mathcal{A}_i(t))\right) < 1 + \frac{4}{\alpha^2} + \frac{1}{1-\exp\left(-\alpha^2/8\right)}.$\qed 
		\end{proof}
		\begin{lemma}
			\label{lem:almostsureconvergenceinequalityTwo}
			 Under the Assumption 1, the TVS sampling policy in Table \ref{alg:vanillathompson} with   $C=(\sqrt{5}-1)/2$ satisfies, for each $i\in \mS^*$ and for some suitable $C(\alpha)>0$,
			 \begin{equation}\label{eq:calpha}
			 \sum_{t=1}^\infty\mathbb{P}\left(\mathcal{A}_i(t)\right) <  C(\alpha).
			 \end{equation}
		\end{lemma}
		
		\begin{proof}
			For a signal arm $i\in \mS^*$ we have $\mathcal{A}_i(t) = \{\theta_i(t)< 0.5\}$. We recall the arguments from Lemma \ref{the:ineqn} in the Supplement. There we denoted with $\tau_j^i$  the $j^{th}$ time  the arm $i$ has been pulled (i.e. $\theta_i(t)>0.5$). Now, we note that in between $[\tau^i_j+1,\tau^i_{j+1}-1]$ we have $ \{\theta_i(t)< 0.5\}$ since the arm $i$ is not being pulled. 
			One can deduce that 
			 
			\begin{equation}\label{eq:summable}
			 \sum_{t=0}^\infty\mathbb{E}\left(  \mathbb{I}(\mathcal{A}_i(t))\right) =\sum_{j=0}^\infty \mathbb{E}\left(\tau^i_{j+1}-\tau^i_{j}-1\right). 
			 \end{equation}
		
		Then note that using the result from Lemma \ref{the:ineqn}. 
		\begin{equation}
		\EE\left[\tau_{j+1}- \tau_j -1\right] \leq \begin{cases}
		3 + \frac{1}{\alpha} \quad \text{when}\quad j \leq \frac{8}{\alpha}\\
		 \frac{1}{\e^{\alpha^2j/4}-1}+ \e^{-\alpha^2j/2}\left(C_1+C_2\frac{1-2\alpha}{4\alpha^2(j+1)}\right) \quad \text{when}\quad j >\frac{8}{\alpha}.
		\end{cases}
		\end{equation}
		From this expression we can immediately see that \eqref{eq:summable} is summable. Below, we provide more details on the upper bound  $C(\alpha)$  in \eqref{eq:calpha}.
		\begin{equation}
		\sum_{j=0}^\infty \mathbb{E}\left(\tau^i_{j+1}-\tau^i_{j}-1\right)  \leq\sum_{j=0}^{\lceil\frac{8}{\alpha}\rceil} \mathbb{E}\left(\tau^i_{j+1}-\tau^i_{j}-1\right) + \sum_{j=\lceil\frac{8}{\alpha}\rceil+1}^{\lceil\frac{4}{\alpha^2}\rceil} \mathbb{E}\left(\tau^i_{j+1}-\tau^i_{j}-1\right) +  \sum_{j=\lceil\frac{4}{\alpha^2}\rceil+1}^{\infty} \mathbb{E}\left(\tau^i_{j+1}-\tau^i_{j}-1\right).
		\end{equation}
		Then we can bound each term separately as:
		$ \sum_{j=0}^{\lceil\frac{8}{\alpha}\rceil} \mathbb{E}\left(\tau^i_{j+1}-\tau^i_{j}-1\right)\leq \lceil\frac{8}{\alpha}\rceil (3 + \frac{1}{\alpha}),$ 
		\begin{align*}
		\sum_{j=\lceil\frac{8}{\alpha}\rceil+1}^{\lceil\frac{4}{\alpha^2}\rceil} \mathbb{E}\left(\tau^i_{j+1}-\tau^i_{j}-1\right) &
		\leq \lceil\frac{4}{\alpha^2}\rceil \left(  \frac{1}{\e^{\alpha^2(\lceil\frac{8}{\alpha}\rceil)/4}-1}+ \e^{-\alpha^2(\lceil\frac{8}{\alpha}\rceil)/2}\left(C_1+C_2\frac{1-2\alpha}{4\alpha^2(\lceil\frac{8}{\alpha}\rceil+1)}  \right)\right),
		\end{align*} 
		\begin{align*}
		\sum_{j=\lceil\frac{4}{\alpha^2}\rceil+1}^{\infty} \mathbb{E}\left(\tau^i_{j+1}-\tau^i_{j}-1\right) &\leq \sum_{j=\lceil\frac{4}{\alpha^2}\rceil+1}^{\infty} \left[\frac{1}{\e^{\alpha^2j/4}-1}+ \e^{-\alpha^2j/2}\left(C_1+C_2\frac{1-2\alpha}{4\alpha^2(j+1)}\right)\right]\\
		&\text{Using the fact that }\exp(-x/2) \geq 1/(\exp(x)-1) \text{ for }x\geq 1\\
		& \leq \sum_{j=\lceil\frac{4}{\alpha^2}\rceil+1}^{\infty}\left[ \e^{-\alpha^2j/8}+ \e^{-\alpha^2j/2}\left(C_1+C_2\frac{1-2\alpha}{4\alpha^2(\lceil\frac{4}{\alpha^2}\rceil+1)}\right)\right]\\
		&\leq \frac{1}{1 -\e^{-\alpha^2/8} } + \frac{1}{1- \e^{-\alpha^2/2}}\left(C_1+C_2\frac{1-2\alpha}{4\alpha^2(\lceil\frac{4}{\alpha^2}\rceil+1)}\right).
		\end{align*}

		\end{proof}

	\newpage
	\section{Additional  Simulation Results}
	\subsection{Offline Cases}
	
	The Figure \ref{fig:sup1} below shows simulation results for $p=1\,000$ under the same settings as Figure 6. 
	In addition, we report  convergence diagnostics (number of iterations until convergence) for the simulation study from Section 6.1, using the convergence criterion ``$\wh S_t$ stays the same for $100$ consecutive TVS iterations", are included in Table  \ref{tab:offline_convergence}.
	
	\begin{figure}[H]
		\includegraphics[width=0.95\columnwidth]{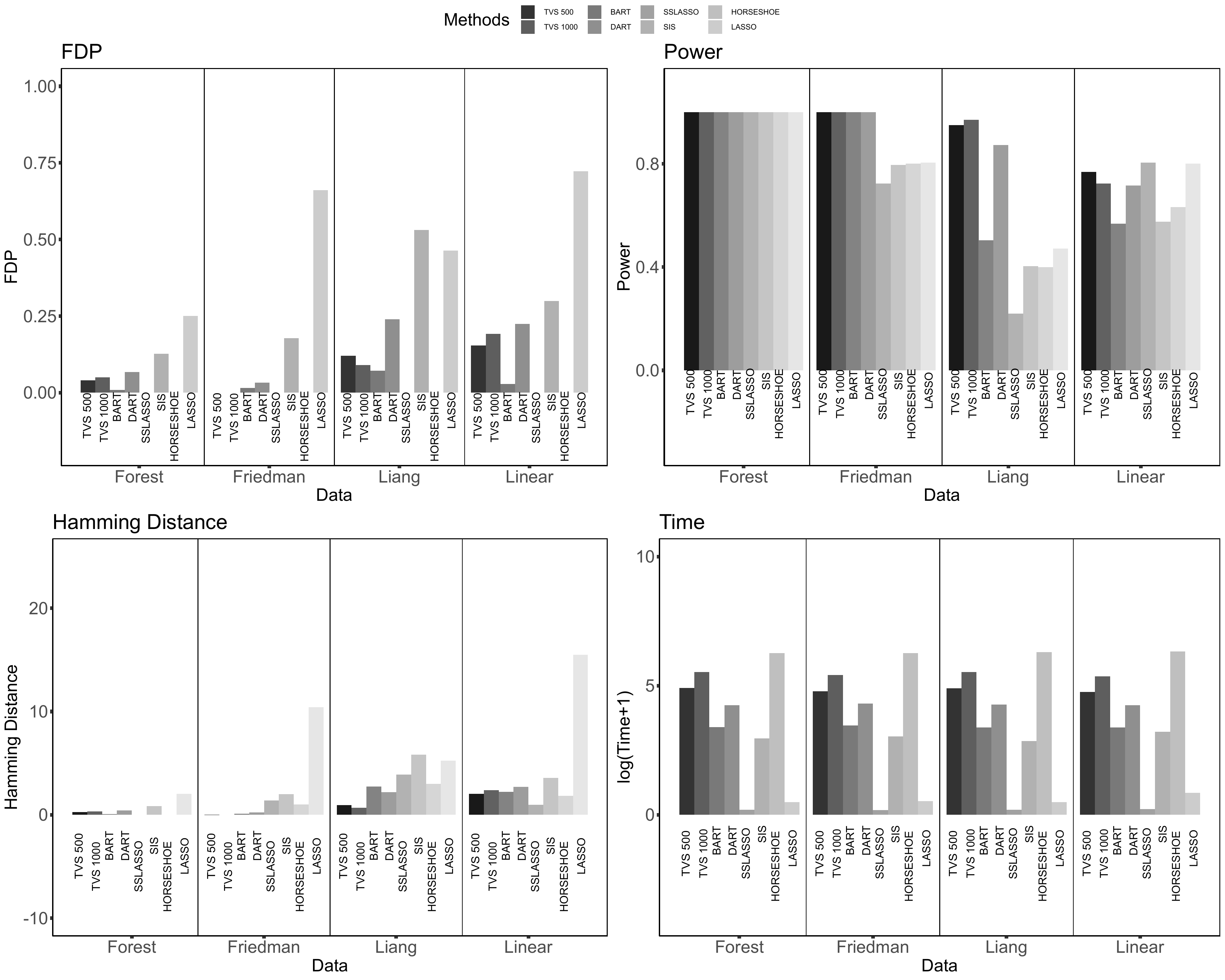}
		\caption{{\small Graphs denoting FDP (\ref{fig:fdr}), Power (\ref{fig:power}), Hamming Distance (\ref{fig:ham}), and Time (\ref{fig:time}) for the $4$ choices of $f_0$ assuming $p=1\,000$ and $n =300$. The $x$-axis denotes the choice of $f_0$ and the various methods are marked with various shades of gray. For TVS, we have two choices $M=500$ and $M=1\,000$.}}\label{fig:sup1}
	\end{figure}

	\begin{table}[ht!]
		\centering
		\small
		\begin{tabular}{llllllllll}
			\hline 
			&                     & \multicolumn{2}{l}{Friedman}                                  & \multicolumn{2}{l}{Linear}                                    & \multicolumn{2}{l}{Forest}                                   & \multicolumn{2}{l}{Liang}                                     \\
			\multirow{-2}{*}{$p$}     & \multirow{-2}{*}{$M$} & Mean                          & SD                            & Mean                          & SD                            & Mean                          & SD                           & Mean                          & SD                            \\ \hline
			& 500                 & {\color[HTML]{000000} 216.40} & {\color[HTML]{000000} 112.69} & {\color[HTML]{000000} 216.50} & {\color[HTML]{000000} 112.68} & {\color[HTML]{000000} 126.16} & {\color[HTML]{000000} 31.11} & {\color[HTML]{000000} 283.58} & {\color[HTML]{000000} 162.44} \\
			\multirow{-2}{*}{1000}  & 1000                & {\color[HTML]{000000} 196.34} & {\color[HTML]{000000} 94.79}  & {\color[HTML]{000000} 197.48} & {\color[HTML]{000000} 94.20}  & 115.30                        & 11.02                        & {\color[HTML]{000000} 232.06} & {\color[HTML]{000000} 130.52} \\ \hline
			& 500                 & {\color[HTML]{000000} 298.04} & {\color[HTML]{000000} 100.51} & {\color[HTML]{000000} 299.58} & {\color[HTML]{000000} 101.01} & 230.14                        & {\color[HTML]{000000} 39.74} & {\color[HTML]{000000} 247.20} & {\color[HTML]{000000} 156.05} \\
			\multirow{-2}{*}{10000} & 1000                & {\color[HTML]{000000} 280.46} & {\color[HTML]{000000} 82.96}  & {\color[HTML]{000000} 280.28} & {\color[HTML]{000000} 82.98}  & {\color[HTML]{000000} 205.10} & {\color[HTML]{000000} 31.86} & {\color[HTML]{000000} 235.74} & {\color[HTML]{000000} 109.29} \\ \hline
		\end{tabular}
		\caption{The table records the number of iterations needed for TVS to converge in the simulation study in Section \ref{sec:simu_offline}. }
		\label{tab:offline_convergence}
	\end{table}
	
	In addition to the convergence criterion, we tried different number of trees $(D)$ for DART and BART. We also considered a different variable selection rule, i.e. the Median Probability Model using the inclusion probability of BART and DART (as mentioned in Linero (2018)). 
	Due to space constraints, we showed only the best settings for BART and  DART in Section \ref{sec:simu_offline}.
	The following tables present the entire simulation study and show that TVS yields better Hamming distance and computational speed gains.
	Tables 5-8 present the offline simulation study, one table for each data setup. 
	\newpage
	\begin{table}[ht!]
		\caption{Linear Setup: BART and DART are implemented using \texttt{Prob} (median probability model) rule or \texttt{Avg Split} (truncating the importance measure at $1$)}
		\centering
		\scalebox{0.67}{
			\begin{tabular}{lllllllllll}
				\hline
				\multicolumn{11}{c}{$p$ = 1000}                                                                                                                                                       \\ \hline
				Method                & \begin{tabular}[c]{@{}l@{}}Convergence\\ Criteria\end{tabular} & M    & Mean Time & SD Time & Mean FDP & SD FDP & Mean Power & SD Power & Mean Ham & SD Ham \\ \hline
				\multirow{6}{*}{TVS}  & YES                                                            & 100  & 30.27     & 11.14   & 0.16     & 0.15   & 0.76       & 0.10     & 2.10     & 1.04   \\
				& NO                                                             & 100  & 37.67     & 5.35    & 0.22     & 0.14   & 0.82       & 0.09     & 2.20     & 1.11   \\
				& YES                                                            & 500  & 84.02     & 39.67   & 0.03     & 0.10   & 0.71       & 0.14     & 1.64     & 1.01   \\
				& NO                                                             & 500  & 115.41    & 8.44    & 0.15     & 0.16   & 0.77       & 0.12     & 2.04     & 1.18   \\
				& YES                                                            & 1000 & 147.72    & 71.70   & 0.07     & 0.12   & 0.66       & 0.16     & 2.04     & 0.90   \\
				& NO                                                             & 1000 & 213.62    & 14.82   & 0.19     & 0.16   & 0.72       & 0.15     & 2.38     & 1.23   \\ \hline
				& \multicolumn{2}{l}{Avg Split/ Prob}                                   & \multicolumn{8}{l}{}                                                                \\ \hline
				\multirow{6}{*}{DART} & \multicolumn{2}{l}{Avg Split}                                         & 38.25     & 2.17    & 0.07     & 0.15   & 0.52       & 0.15     & 2.62     & 0.97   \\
				& \multicolumn{2}{l}{Prob}                                              & 38.25     & 2.17    & 0.16     & 0.17   & 0.57       & 0.16     & 2.80     & 1.09   \\
				& \multicolumn{2}{l}{Avg Split}                                         & 83.47     & 4.68    & 0.23     & 0.19   & 0.72       & 0.16     & 2.78     & 1.56   \\
				& \multicolumn{2}{l}{Prob}                                              & 83.47     & 4.68    & 0.32     & 0.20   & 0.76       & 0.16     & 3.38     & 2.01   \\
				& \multicolumn{2}{l}{Avg Split}                                         & 281.94    & 34.48   & 0.42     & 0.19   & 0.66       & 0.17     & 4.46     & 1.96   \\
				& \multicolumn{2}{l}{Prob}                                              & 281.94    & 34.48   & 0.48     & 0.20   & 0.69       & 0.19     & 5.44     & 2.59   \\ \hline
				\multirow{6}{*}{BART} & \multicolumn{2}{l}{Avg Split}                                         & 34.93     & 3.13    & 0.03     & 0.10   & 0.57       & 0.15     & 2.24     & 0.87   \\
				& \multicolumn{2}{l}{Prob}                                              & 34.93     & 3.13    & 0.17     & 0.17   & 0.70       & 0.12     & 2.34     & 1.22   \\
				& \multicolumn{2}{l}{Avg Split}                                         & 83.39     & 6.19    & 0.03     & 0.08   & 0.66       & 0.15     & 1.82     & 0.83   \\
				& \multicolumn{2}{l}{Prob}                                              & 83.39     & 6.19    & 0.24     & 0.15   & 0.76       & 0.11     & 2.60     & 1.14   \\
				& \multicolumn{2}{l}{Avg Split}                                         & 321.77    & 21.53   & 0.06     & 0.11   & 0.72       & 0.11     & 1.68     & 0.77   \\
				& \multicolumn{2}{l}{Prob}                                              & 321.77    & 21.53   & 0.39     & 0.17   & 0.81       & 0.07     & 3.98     & 1.83   \\ \hline
				\multicolumn{3}{l}{SSLASSO}                                                                   & 0.40      & 0.06    & 0.00     & 0.00   & 0.80       & 0.10     & 0.98     & 0.51   \\ \hline
				\multicolumn{11}{c}{$p$=10000}                                                                                                                                                        \\ \hline
				\multirow{6}{*}{TVS}  & YES                                                            & 100  & 43.18     & 13.83   & 0.08     & 0.16   & 0.21       & 0.20     & 4.22     & 0.84   \\
				& NO                                                             & 100  & 54.28     & 7.37    & 0.27     & 0.20   & 0.49       & 0.13     & 3.56     & 1.15   \\
				& YES                                                            & 500  & 113.64    & 33.34   & 0.15     & 0.22   & 0.48       & 0.17     & 3.14     & 1.31   \\
				& NO                                                             & 500  & 146.72    & 18.23   & 0.13     & 0.20   & 0.57       & 0.14     & 2.70     & 1.25   \\
				& YES                                                            & 1000 & 184.65    & 52.60   & 0.07     & 0.15   & 0.50       & 0.12     & 2.74     & 0.90   \\
				& NO                                                             & 1000 & 262.24    & 30.22   & 0.07     & 0.15   & 0.59       & 0.15     & 2.38     & 1.09   \\ \hline
				& \multicolumn{2}{l}{Avg Split/ Prob}                                   & \multicolumn{8}{l}{}                                                                \\ \hline
				\multirow{6}{*}{DART} & \multicolumn{2}{l}{Avg Split}                                         & 251.30    & 30.96   & 0.17     & 0.23   & 0.44       & 0.14     & 3.30     & 1.23   \\
				& \multicolumn{2}{l}{Prob}                                              & 251.30    & 30.96   & 0.24     & 0.23   & 0.48       & 0.16     & 3.48     & 1.42   \\
				& \multicolumn{2}{l}{Avg Split}                                         & 544.92    & 79.06   & 0.23     & 0.20   & 0.66       & 0.15     & 2.84     & 1.60   \\
				& \multicolumn{2}{l}{Prob}                                              & 544.92    & 79.06   & 0.44     & 0.19   & 0.70       & 0.17     & 4.84     & 2.48   \\
				& \multicolumn{2}{l}{Avg Split}                                         & 1690.91   & 303.85  & 0.74     & 0.09   & 0.68       & 0.19     & 11.74    & 3.46   \\
				& \multicolumn{2}{l}{Prob}                                              & 1690.91   & 303.85  & 0.84     & 0.05   & 0.70       & 0.19     & 21.52    & 5.68   \\ \hline
				\multirow{6}{*}{BART} & \multicolumn{2}{l}{Avg Split}                                         & 238.08    & 54.78   & 0.00     & 0.00   & 0.25       & 0.15     & 3.76     & 0.74   \\
				& \multicolumn{2}{l}{Prob}                                              & 238.08    & 54.78   & 0.29     & 0.24   & 0.48       & 0.18     & 3.92     & 1.41   \\
				& \multicolumn{2}{l}{Avg Split}                                         & 542.44    & 104.17  & 0.06     & 0.16   & 0.37       & 0.15     & 3.26     & 0.90   \\
				& \multicolumn{2}{l}{Prob}                                              & 542.44    & 104.17  & 0.25     & 0.21   & 0.58       & 0.15     & 3.32     & 1.42   \\
				& \multicolumn{2}{l}{Avg Split}                                         & 2057.58   & 368.99  & 0.01     & 0.05   & 0.42       & 0.15     & 2.92     & 0.78   \\
				& \multicolumn{2}{l}{Prob}                                              & 2057.58   & 368.99  & 0.15     & 0.18   & 0.57       & 0.12     & 2.82     & 1.06   \\ \hline
				\multicolumn{3}{l}{SSLASSO}                                                                   & 9.59      & 0.61    & 0.00     & 0.00   & 0.45       & 0.09     & 2.74     & 0.44   \\ \hline
			\end{tabular}
		}
	\end{table}
	
	\begin{table}[ht!]
		\centering
		\caption{Liang Setup: BART and DART are implemented using \texttt{Prob} (median probability model) rule or \texttt{Avg Split} (truncating the importance measure at $1$)}
		\scalebox{0.67}{
			\begin{tabular}{lllllllllll}
				\hline
				\multicolumn{11}{c}{$p$ = 1000}                                                                                                                                                       \\ \hline
				Method                & \begin{tabular}[c]{@{}l@{}}Convergence\\ Criteria\end{tabular} & M    & Mean Time & SD Time & Mean FDP & SD FDP & Mean Power & SD Power & Mean Ham & SD Ham \\ \hline
				\multirow{6}{*}{TVS}  & YES                                                            & 100  & 36.24     & 14.22   & 0.15     & 0.15   & 0.82       & 0.18     & 1.80     & 1.23   \\
				& NO                                                             & 100  & 41.61     & 9.80    & 0.20     & 0.13   & 0.87       & 0.15     & 1.78     & 1.23   \\
				& YES                                                            & 500  & 97.16     & 44.16   & 0.08     & 0.13   & 0.86       & 0.17     & 1.20     & 1.14   \\
				& NO                                                             & 500  & 129.34    & 18.88   & 0.14     & 0.12   & 0.93       & 0.13     & 1.18     & 1.16   \\
				& YES                                                            & 1000 & 143.95    & 81.27   & 0.06     & 0.14   & 0.87       & 0.16     & 1.04     & 1.35   \\
				& NO                                                             & 1000 & 239.32    & 32.67   & 0.11     & 0.15   & 0.94       & 0.13     & 0.96     & 1.47   \\ \hline
				& \multicolumn{2}{l}{Avg Split/ Prob}                                   & \multicolumn{8}{l}{}                                                                \\ \hline
				\multirow{6}{*}{DART} & \multicolumn{2}{l}{Avg Split}                                         & 42.08     & 6.82    & 0.14     & 0.18   & 0.56       & 0.16     & 2.74     & 1.23   \\
				& \multicolumn{2}{l}{Prob}                                              & 42.08     & 6.82    & 0.26     & 0.21   & 0.59       & 0.16     & 3.24     & 1.61   \\
				& \multicolumn{2}{l}{Avg Split}                                         & 91.97     & 14.56   & 0.24     & 0.17   & 0.87       & 0.15     & 2.22     & 1.72   \\
				& \multicolumn{2}{l}{Prob}                                              & 91.97     & 14.56   & 0.41     & 0.20   & 0.90       & 0.14     & 4.36     & 2.83   \\
				& \multicolumn{2}{l}{Avg Split}                                         & 285.21    & 46.55   & 0.33     & 0.21   & 0.91       & 0.14     & 3.18     & 2.69   \\
				& \multicolumn{2}{l}{Prob}                                              & 285.21    & 46.55   & 0.45     & 0.21   & 0.92       & 0.13     & 5.22     & 3.74   \\ \hline
				\multirow{6}{*}{BART} & \multicolumn{2}{l}{Avg Split}                                         & 37.46     & 6.25    & 0.06     & 0.13   & 0.50       & 0.16     & 2.70     & 0.99   \\
				& \multicolumn{2}{l}{Prob}                                              & 37.46     & 6.25    & 0.43     & 0.18   & 0.69       & 0.17     & 4.68     & 1.96   \\
				& \multicolumn{2}{l}{Avg Split}                                         & 87.44     & 13.23   & 0.09     & 0.15   & 0.62       & 0.17     & 2.28     & 1.13   \\
				& \multicolumn{2}{l}{Prob}                                              & 87.44     & 13.23   & 0.44     & 0.22   & 0.76       & 0.15     & 5.10     & 2.94   \\
				& \multicolumn{2}{l}{Avg Split}                                         & 334.66    & 50.66   & 0.01     & 0.06   & 0.58       & 0.16     & 2.16     & 0.87   \\
				& \multicolumn{2}{l}{Prob}                                              & 334.66    & 50.66   & 0.39     & 0.16   & 0.79       & 0.17     & 3.98     & 2.06   \\ \hline
				\multicolumn{3}{l}{SSLASSO}                                                                   & 0.35      & 0.05    & 0.00     & 0.00   & 0.22       & 0.08     & 3.90     & 0.42   \\ \hline
				\multicolumn{11}{c}{$p$=10000}                                                                                                                                                        \\ \hline
				\multirow{6}{*}{TVS}  & YES                                                            & 100  & 36.28     & 18.21   & 0.01     & 0.07   & 0.12       & 0.13     & 4.44     & 0.67   \\
				& NO                                                             & 100  & 58.38     & 7.85    & 0.02     & 0.12   & 0.37       & 0.09     & 3.22     & 0.65   \\
				& YES                                                            & 500  & 129.30    & 49.27   & 0.02     & 0.07   & 0.33       & 0.17     & 3.42     & 0.73   \\
				& NO                                                             & 500  & 160.50    & 28.76   & 0.13     & 0.18   & 0.52       & 0.13     & 2.88     & 1.08   \\
				& YES                                                            & 1000 & 215.42    & 77.68   & 0.02     & 0.08   & 0.41       & 0.16     & 3.02     & 0.80   \\
				& NO                                                             & 1000 & 281.03    & 33.14   & 0.17     & 0.20   & 0.62       & 0.17     & 2.64     & 1.55   \\ \hline
				& \multicolumn{2}{l}{Avg Split/ Prob}                                   & \multicolumn{8}{l}{}                                                                \\ \hline
				\multirow{6}{*}{DART} & \multicolumn{2}{l}{Avg Split}                                         & 264.51    & 45.73   & 0.12     & 0.21   & 0.33       & 0.14     & 3.64     & 1.01   \\
				& \multicolumn{2}{l}{Prob}                                              & 264.51    & 45.73   & 0.50     & 0.24   & 0.37       & 0.15     & 5.34     & 1.78   \\
				& \multicolumn{2}{l}{Avg Split}                                         & 568.41    & 105.58  & 0.41     & 0.24   & 0.58       & 0.18     & 4.44     & 2.36   \\
				& \multicolumn{2}{l}{Prob}                                              & 568.41    & 105.58  & 0.65     & 0.17   & 0.60       & 0.19     & 8.62     & 3.72   \\
				& \multicolumn{2}{l}{Avg Split}                                         & 1713.13   & 388.03  & 0.67     & 0.16   & 0.68       & 0.19     & 9.94     & 4.49   \\
				& \multicolumn{2}{l}{Prob}                                              & 1713.13   & 388.03  & 0.81     & 0.11   & 0.70       & 0.19     & 19.02    & 7.74   \\ \hline
				\multirow{6}{*}{BART} & \multicolumn{2}{l}{Avg Split}                                         & 235.79    & 48.24   & 0.02     & 0.10   & 0.21       & 0.05     & 3.98     & 0.32   \\
				& \multicolumn{2}{l}{Prob}                                              & 235.79    & 48.24   & 0.59     & 0.19   & 0.32       & 0.12     & 6.06     & 1.65   \\
				& \multicolumn{2}{l}{Avg Split}                                         & 538.84    & 103.64  & 0.05     & 0.17   & 0.25       & 0.09     & 3.90     & 0.65   \\
				& \multicolumn{2}{l}{Prob}                                              & 538.84    & 103.64  & 0.59     & 0.21   & 0.38       & 0.09     & 6.82     & 2.67   \\
				& \multicolumn{2}{l}{Avg Split}                                         & 2053.60   & 373.80  & 0.05     & 0.13   & 0.29       & 0.10     & 3.66     & 0.56   \\
				& \multicolumn{2}{l}{Prob}                                              & 2053.60   & 373.80  & 0.16     & 0.21   & 0.40       & 0.09     & 3.56     & 0.91   \\ \hline
				\multicolumn{3}{l}{SSLASSO}                                                                   & 2.45      & 0.22    & 0.00     & 0.00   & 0.20       & 0.00     & 4.00     & 0.00   \\ \hline
		\end{tabular}}
	\end{table}
	
	\begin{table}[ht!]
		\caption{Friedman  Setup: BART and DART are implemented using \texttt{Prob} (median probability model) rule or \texttt{Avg Split} (truncating the importance measure at $1$)}
		\scalebox{0.67}{
			\begin{tabular}{lllllllllll}
				\hline
				\multicolumn{11}{c}{$p$ = 1000}                                                                                                                                                       \\ \hline
				Method                & \begin{tabular}[c]{@{}l@{}}Convergence\\ Criteria\end{tabular} & M    & Mean Time & SD Time & Mean FDP & SD FDP & Mean Power & SD Power & Mean Ham & SD Ham \\ \hline
				\multirow{6}{*}{TVS}  & YES                                                            & 100  & 31.06     & 12.07   & 0.17     & 0.14   & 0.77       & 0.10     & 2.08     & 1.03   \\
				& NO                                                             & 100  & 38.42     & 6.13    & 0.23     & 0.14   & 0.82       & 0.10     & 2.26     & 1.14   \\
				& YES                                                            & 500  & 86.28     & 41.23   & 0.04     & 0.11   & 0.70       & 0.15     & 1.72     & 1.05   \\
				& NO                                                             & 500  & 119.10    & 12.98   & 0.16     & 0.16   & 0.77       & 0.13     & 2.08     & 1.21   \\
				& YES                                                            & 1000 & 150.88    & 73.61   & 0.08     & 0.13   & 0.65       & 0.16     & 2.12     & 0.92   \\
				& NO                                                             & 1000 & 221.15    & 24.51   & 0.20     & 0.16   & 0.71       & 0.16     & 2.48     & 1.22   \\ \hline
				& \multicolumn{2}{l}{Avg Split/ Prob}                                   & \multicolumn{8}{l}{}                                                                \\ \hline
				\multirow{6}{*}{DART} & \multicolumn{2}{l}{Avg Split}                                         & 41.52     & 5.41    & 0.07     & 0.15   & 0.52       & 0.15     & 2.62     & 0.97   \\
				& \multicolumn{2}{l}{Prob}                                              & 41.52     & 5.41    & 0.17     & 0.17   & 0.57       & 0.16     & 2.80     & 1.09   \\
				& \multicolumn{2}{l}{Avg Split}                                         & 92.19     & 12.73   & 0.24     & 0.18   & 0.72       & 0.17     & 2.82     & 1.56   \\
				& \multicolumn{2}{l}{Prob}                                              & 92.19     & 12.73   & 0.33     & 0.19   & 0.76       & 0.17     & 3.46     & 1.99   \\
				& \multicolumn{2}{l}{Avg Split}                                         & 309.60    & 48.65   & 0.43     & 0.18   & 0.67       & 0.17     & 4.52     & 1.96   \\
				& \multicolumn{2}{l}{Prob}                                              & 309.60    & 48.65   & 0.49     & 0.19   & 0.69       & 0.19     & 5.52     & 2.56   \\ \hline
				\multirow{6}{*}{BART} & \multicolumn{2}{l}{Avg Split}                                         & 38.17     & 5.52    & 0.03     & 0.10   & 0.56       & 0.15     & 2.26     & 0.90   \\
				& \multicolumn{2}{l}{Prob}                                              & 38.17     & 5.52    & 0.17     & 0.17   & 0.70       & 0.12     & 2.38     & 1.28   \\
				& \multicolumn{2}{l}{Avg Split}                                         & 91.71     & 13.07   & 0.03     & 0.09   & 0.66       & 0.15     & 1.84     & 0.84   \\
				& \multicolumn{2}{l}{Prob}                                              & 91.71     & 13.07   & 0.25     & 0.15   & 0.75       & 0.11     & 2.68     & 1.17   \\
				& \multicolumn{2}{l}{Avg Split}                                         & 356.32    & 52.27   & 0.07     & 0.11   & 0.72       & 0.11     & 1.72     & 0.78   \\
				& \multicolumn{2}{l}{Prob}                                              & 356.32    & 52.27   & 0.39     & 0.18   & 0.82       & 0.08     & 4.02     & 1.86   \\ \hline
				\multicolumn{3}{l}{SSLASSO}                                                                   & 0.39      & 0.08    & 0.00     & 0.00   & 0.80       & 0.10     & 1.00     & 0.49   \\ \hline
				\multicolumn{11}{c}{$p$=10000}                                                                                                                                                        \\ \hline
				\multirow{6}{*}{TVS}  & YES                                                            & 100  & 42.50     & 13.82   & 0.08     & 0.16   & 0.20       & 0.20     & 4.24     & 0.85   \\
				& NO                                                             & 100  & 53.41     & 6.40    & 0.26     & 0.20   & 0.50       & 0.12     & 3.50     & 1.15   \\
				& YES                                                            & 500  & 113.50    & 35.77   & 0.15     & 0.22   & 0.48       & 0.17     & 3.14     & 1.31   \\
				& NO                                                             & 500  & 145.87    & 15.29   & 0.13     & 0.20   & 0.57       & 0.14     & 2.70     & 1.25   \\
				& YES                                                            & 1000 & 183.96    & 56.91   & 0.08     & 0.16   & 0.50       & 0.12     & 2.76     & 0.92   \\
				& NO                                                             & 1000 & 259.48    & 26.15   & 0.08     & 0.16   & 0.58       & 0.16     & 2.42     & 1.11   \\ \hline
				& \multicolumn{2}{l}{Avg Split/ Prob}                                   & \multicolumn{8}{l}{}                                                                \\ \hline
				\multirow{6}{*}{DART} & \multicolumn{2}{l}{Avg Split}                                         & 269.16    & 39.30   & 0.17     & 0.23   & 0.44       & 0.14     & 3.30     & 1.23   \\
				& \multicolumn{2}{l}{Prob}                                              & 269.16    & 39.30   & 0.24     & 0.23   & 0.48       & 0.16     & 3.48     & 1.42   \\
				& \multicolumn{2}{l}{Avg Split}                                         & 590.78    & 96.73   & 0.24     & 0.19   & 0.66       & 0.15     & 2.90     & 1.58   \\
				& \multicolumn{2}{l}{Prob}                                              & 590.78    & 96.73   & 0.45     & 0.19   & 0.70       & 0.17     & 4.92     & 2.45   \\
				& \multicolumn{2}{l}{Avg Split}                                         & 1839.84   & 382.64  & 0.74     & 0.09   & 0.69       & 0.19     & 11.78    & 3.45   \\
				& \multicolumn{2}{l}{Prob}                                              & 1839.84   & 382.64  & 0.84     & 0.05   & 0.71       & 0.19     & 21.58    & 5.70   \\ \hline
				\multirow{6}{*}{BART} & \multicolumn{2}{l}{Avg Split}                                         & 250.06    & 46.46   & 0.00     & 0.00   & 0.24       & 0.15     & 3.78     & 0.76   \\
				& \multicolumn{2}{l}{Prob}                                              & 250.06    & 46.46   & 0.29     & 0.24   & 0.47       & 0.18     & 3.94     & 1.42   \\
				& \multicolumn{2}{l}{Avg Split}                                         & 573.10    & 104.74  & 0.06     & 0.16   & 0.36       & 0.14     & 3.30     & 0.89   \\
				& \multicolumn{2}{l}{Prob}                                              & 573.10    & 104.74  & 0.26     & 0.21   & 0.58       & 0.15     & 3.34     & 1.41   \\
				& \multicolumn{2}{l}{Avg Split}                                         & 2211.11   & 394.91  & 0.01     & 0.05   & 0.42       & 0.15     & 2.92     & 0.78   \\
				& \multicolumn{2}{l}{Prob}                                              & 2211.11   & 394.91  & 0.14     & 0.18   & 0.56       & 0.12     & 2.82     & 1.06   \\ \hline
				\multicolumn{3}{l}{SSLASSO}                                                                   & 9.38      & 0.64    & 0.00     & 0.00   & 0.45       & 0.09     & 2.76     & 0.43   \\ \hline
		\end{tabular} }
	\end{table}
	
	\begin{table}[ht!]
		\caption{Forest  Setup: BART and DART are implemented using \texttt{Prob} (median probability model) rule or \texttt{Avg Split} (truncating the importance measure at $1$)}
		\scalebox{0.67}{
			\begin{tabular}{lllllllllll}
				\hline
				\multicolumn{11}{c}{$p$ = 1000}                                                                                                                                                       \\ \hline
				Method                & \begin{tabular}[c]{@{}l@{}}Convergence\\ Criteria\end{tabular} & M    & Mean Time & SD Time & Mean FDP & SD FDP & Mean Power & SD Power & Mean Ham & SD Ham \\ \hline
				\multirow{6}{*}{TVS}  & YES                                                            & 100  & 18.34     & 12.51   & 0.00     & 0.02   & 1.00       & 0.00     & 0.02     & 0.14   \\
				& NO                                                             & 100  & 39.30     & 3.53    & 0.05     & 0.08   & 1.00       & 0.00     & 0.32     & 0.51   \\
				& YES                                                            & 500  & 45.56     & 34.00   & 0.01     & 0.04   & 1.00       & 0.00     & 0.06     & 0.24   \\
				& NO                                                             & 500  & 126.55    & 10.66   & 0.06     & 0.09   & 1.00       & 0.00     & 0.36     & 0.56   \\
				& YES                                                            & 1000 & 84.82     & 67.87   & 0.00     & 0.02   & 1.00       & 0.00     & 0.02     & 0.14   \\
				& NO                                                             & 1000 & 235.24    & 19.79   & 0.08     & 0.10   & 1.00       & 0.00     & 0.50     & 0.61   \\ \hline
				& \multicolumn{2}{l}{Avg Split/ Prob}                                   & \multicolumn{8}{l}{}                                                                \\ \hline
				\multirow{6}{*}{DART} & \multicolumn{2}{l}{Avg Split}                                         & 42.52     & 5.71    & 0.00     & 0.00   & 0.99       & 0.04     & 0.04     & 0.20   \\
				& \multicolumn{2}{l}{Prob}                                              & 42.52     & 5.71    & 0.00     & 0.02   & 1.00       & 0.00     & 0.02     & 0.14   \\
				& \multicolumn{2}{l}{Avg Split}                                         & 91.69     & 12.63   & 0.07     & 0.09   & 1.00       & 0.00     & 0.44     & 0.61   \\
				& \multicolumn{2}{l}{Prob}                                              & 91.69     & 12.63   & 0.17     & 0.15   & 1.00       & 0.00     & 1.24     & 1.20   \\
				& \multicolumn{2}{l}{Avg Split}                                         & 302.68    & 51.01   & 0.33     & 0.14   & 1.00       & 0.00     & 2.78     & 1.59   \\
				& \multicolumn{2}{l}{Prob}                                              & 302.68    & 51.01   & 0.42     & 0.14   & 1.00       & 0.00     & 4.14     & 2.00   \\ \hline
				\multirow{6}{*}{BART} & \multicolumn{2}{l}{Avg Split}                                         & 38.90     & 6.03    & 0.01     & 0.05   & 1.00       & 0.00     & 0.06     & 0.31   \\
				& \multicolumn{2}{l}{Prob}                                              & 38.90     & 6.03    & 0.24     & 0.15   & 1.00       & 0.00     & 1.88     & 1.52   \\
				& \multicolumn{2}{l}{Avg Split}                                         & 92.01     & 13.74   & 0.02     & 0.06   & 1.00       & 0.00     & 0.14     & 0.35   \\
				& \multicolumn{2}{l}{Prob}                                              & 92.01     & 13.74   & 0.39     & 0.17   & 1.00       & 0.00     & 3.94     & 2.82   \\
				& \multicolumn{2}{l}{Avg Split}                                         & 355.26    & 55.65   & 0.03     & 0.06   & 1.00       & 0.00     & 0.18     & 0.39   \\
				& \multicolumn{2}{l}{Prob}                                              & 355.26    & 55.65   & 0.52     & 0.11   & 1.00       & 0.00     & 5.86     & 2.33   \\ \hline
				\multicolumn{3}{l}{SSLASSO}                                                                   & 0.36      & 0.08    & 0.00     & 0.00   & 1.00       & 0.00     & 0.00     & 0.00   \\ \hline
				\multicolumn{11}{c}{$p=10\,000$}                                                                                                                                                        \\ \hline
				\multirow{6}{*}{TVS}  & YES                                                            & 100  & 38.14     & 10.73   & 0.00     & 0.00   & 0.78       & 0.34     & 1.12     & 1.72   \\
				& NO                                                             & 100  & 57.82     & 7.10    & 0.00     & 0.00   & 0.98       & 0.07     & 0.12     & 0.33   \\
				& YES                                                            & 500  & 85.29     & 16.72   & 0.00     & 0.00   & 0.96       & 0.09     & 0.18     & 0.44   \\
				& NO                                                             & 500  & 158.70    & 17.03   & 0.00     & 0.02   & 0.99       & 0.04     & 0.06     & 0.24   \\
				& YES                                                            & 1000 & 143.99    & 41.27   & 0.00     & 0.00   & 0.97       & 0.08     & 0.14     & 0.40   \\
				& NO                                                             & 1000 & 282.44    & 30.03   & 0.01     & 0.05   & 1.00       & 0.03     & 0.10     & 0.36   \\ \hline
				& \multicolumn{2}{l}{Avg Split/ Prob}                                   & \multicolumn{8}{l}{}                                                                \\ \hline
				\multirow{6}{*}{DART} & \multicolumn{2}{l}{Avg Split}                                         & 262.89    & 38.21   & 0.00     & 0.00   & 0.94       & 0.10     & 0.28     & 0.50   \\
				& \multicolumn{2}{l}{Prob}                                              & 262.89    & 38.21   & 0.02     & 0.06   & 0.98       & 0.06     & 0.22     & 0.46   \\
				& \multicolumn{2}{l}{Avg Split}                                         & 585.06    & 100.00  & 0.15     & 0.14   & 1.00       & 0.00     & 1.10     & 1.15   \\
				& \multicolumn{2}{l}{Prob}                                              & 585.06    & 100.00  & 0.41     & 0.13   & 1.00       & 0.00     & 3.82     & 1.79   \\
				& \multicolumn{2}{l}{Avg Split}                                         & 1868.72   & 308.73  & 0.69     & 0.09   & 1.00       & 0.00     & 12.00    & 3.81   \\
				& \multicolumn{2}{l}{Prob}                                              & 1868.72   & 308.73  & 0.81     & 0.05   & 1.00       & 0.00     & 22.84    & 6.44   \\ \hline
				\multirow{6}{*}{BART} & \multicolumn{2}{l}{Avg Split}                                         & 251.37    & 53.89   & 0.00     & 0.00   & 0.76       & 0.17     & 1.20     & 0.86   \\
				& \multicolumn{2}{l}{Prob}                                              & 251.37    & 53.89   & 0.28     & 0.17   & 0.99       & 0.05     & 2.34     & 1.85   \\
				& \multicolumn{2}{l}{Avg Split}                                         & 571.27    & 115.64  & 0.01     & 0.05   & 0.95       & 0.12     & 0.32     & 0.62   \\
				& \multicolumn{2}{l}{Prob}                                              & 571.27    & 115.64  & 0.38     & 0.15   & 1.00       & 0.00     & 3.54     & 2.06   \\
				& \multicolumn{2}{l}{Avg Split}                                         & 2206.52   & 426.63  & 0.00     & 0.00   & 1.00       & 0.03     & 0.02     & 0.14   \\
				& \multicolumn{2}{l}{Prob}                                              & 2206.52   & 426.63  & 0.09     & 0.12   & 1.00       & 0.00     & 0.62     & 0.85   \\ \hline
				\multicolumn{3}{l}{SSLASSO}                                                                   & 2.99      & 0.72    & 0.00     & 0.00   & 0.90       & 0.15     & 0.52     & 0.74   \\ \hline
		\end{tabular}}
	\end{table}
	
	\clearpage
	\subsection{Online Cases}\label{sec:appendix_online_simul}
	In Table \ref{tab:online_convergence}, we report  convergence diagnostics of the simulation from Section 6.2, where the convergence criterion is chosen as ``$\wh S_t$ stays the same for $100$ consecutive TVS iterations". Table 10,12,14 and 16 report the results with $10\,000$, comparing TVS with BART. These results show that TVS at the very least highly competitive with DART in terms of Hamming distance but offers vast computational benefits compared to BART and DART.  Tables 11, 13,15 and 17 report TVS results with $n=50\,000$ and $n=100\,000$. We could not run BART on these large datasets and thereby we report only on TVS.
	
	\begin{table}[H]
		\centering
		
		\scalebox{1.0}{	\begin{tabular}{lllllllllll}
				\hline
				\multirow{2}{*}{$n$}      & \multirow{2}{*}{$M$} & \multirow{2}{*}{$s$} & \multicolumn{2}{l}{Friedman} & \multicolumn{2}{l}{Forest} & \multicolumn{2}{l}{Linear} & \multicolumn{2}{l}{Liang} \\
				&                    &                    & Mean          & SD           & Mean         & SD          & Mean         & SD          & Mean         & SD         \\ \hline
				\multirow{4}{*}{$10\,000$}  & 500                & 500                & 9.40          & 2.74         & 6.56         & 1.16        & 9.40         & 2.74        & 10.00        & 1.60       \\
				& 1000               & 500                & 8.32          & 1.32         & 6.06         & 0.24        & 8.32         & 1.32        & 8.54         & 1.31       \\
				& 500                & 1000               & 23.46         & 7.99         & 13.64        & 2.94        & 23.46        & 7.99        & 18.14        & 3.83       \\
				& 1000               & 100                & 21.14         & 7.15         & 12.44        & 2.15        & 21.14        & 7.15        & 15.58        & 2.92       \\ \hline
				\multirow{4}{*}{$50\,000$}  & 500                & 500                & 2.42          & 0.76         & 2.00         & 0.00        & 2.42         & 0.76        & 2.28         & 0.45       \\
				& 1000               & 500                & 2.14          & 0.35         & 2.00         & 0.00        & 2.14         & 0.35        & 2.12         & 0.33       \\
				& 500                & 1000               & 4.94          & 1.75         & 3.10         & 0.36        & 4.94         & 1.75        & 4.04         & 0.83       \\
				& 1000               & 100                & 4.10          & 1.25         & 3.08         & 0.27        & 4.10         & 1.25        & 3.48         & 0.71       \\ \hline
				\multirow{4}{*}{$100\,000$} & 500                & 500                & 1.20          & 0.45         & 1.00         & 0.00        & 1.22         & 0.46        & 1.38         & 0.49       \\
				& 1000               & 500                & 1.18          & 0.39         & 1.00         & 0.00        & 1.18         & 0.39        & 1.10         & 0.30       \\
				& 500                & 1000               & 2.72          & 0.73         & 2.04         & 0.20        & 2.74         & 0.72        & 2.24         & 0.48       \\
				& 1000               & 100                & 2.66          & 0.87         & 2.04         & 0.20        & 2.64         & 0.88        & 2.14         & 0.35       \\ \hline
		\end{tabular}}
		
		\caption{\small The table records the number of rounds needed for TVS to converge in the simulation study in Section \ref{sec:simu_online}. }
		\label{tab:online_convergence}
	\end{table}
	
	In addition to the results shown in Section 6.2, we tried a different number of trees $D$ for DART and BART. We also considered a different variable selection rule, i.e. the Median Probability Model using the inclusion probability of BART and DART (as mentioned in \cite{linero_bayesian_2018}). 
	Due to space constraints, we showed only the best settings for BART and  DART in Section \ref{sec:simu_online}.
	Now we present additional  simulation results for $n=50\,000$ and $n=100\,000$. Since BART and DART cannot be run with such large $n$, we only show the results for TVS. 
	

\begin{table}[ht!]
	\caption{Linear Setup with $n=10\,000$, BART and DART are implemented using \texttt{Prob} (median probability model) rule or \texttt{Avg Split} (truncating the importance measure at $1$); $s$ is the batch size, $r$ is the number of rounds and $M$ is the number of internal MCMC iterations in TVS.}
	\scalebox{0.7}{
\begin{tabular}{llllllllllll}
	\hline
	Method                & M    & s                                 & r                                         & Mean Time & SD Time & Mean FDR & SD FDR & Mean Power & SD Power & Mean Ham & SD Ham \\ \hline
	\multirow{16}{*}{TVS} & 500  & 500                               & Till Converge                             & 52.80     & 12.24   & 0.00     & 0.02   & 1.00       & 0.00     & 0.02     & 0.14   \\
	& 500  & 500                               & 1                                         & 8.33      & 1.21    & 0.00     & 0.00   & 0.96       & 0.09     & 0.18     & 0.44   \\
	& 500  & 500                               & 5                                         & 40.29     & 5.31    & 0.00     & 0.00   & 1.00       & 0.00     & 0.00     & 0.00   \\
	& 500  & 500                               & 10                                        & 79.77     & 10.26   & 0.00     & 0.00   & 1.00       & 0.00     & 0.00     & 0.00   \\
	& 1000 & 500                               & Till Converge                             & 88.68     & 11.37   & 0.00     & 0.00   & 1.00       & 0.00     & 0.00     & 0.00   \\
	& 1000 & 500                               & 1                                         & 14.79     & 1.95    & 0.00     & 0.00   & 0.98       & 0.05     & 0.08     & 0.27   \\
	& 1000 & 500                               & 5                                         & 73.30     & 9.15    & 0.00     & 0.00   & 1.00       & 0.00     & 0.00     & 0.00   \\
	& 1000 & 500                               & 10                                        & 145.98    & 18.12   & 0.00     & 0.00   & 1.00       & 0.00     & 0.00     & 0.00   \\
	& 500  & 1000                              & Till Converge                             & 91.11     & 23.10   & 0.00     & 0.00   & 1.00       & 0.00     & 0.00     & 0.00   \\
	& 500  & 1000                              & 1                                         & 6.42      & 1.01    & 0.03     & 0.09   & 0.69       & 0.19     & 1.68     & 1.00   \\
	& 500  & 1000                              & 5                                         & 33.02     & 5.05    & 0.00     & 0.00   & 1.00       & 0.00     & 0.00     & 0.00   \\
	& 500  & 1000                              & 10                                        & 66.82     & 9.69    & 0.00     & 0.02   & 1.00       & 0.00     & 0.02     & 0.14   \\
	& 1000 & 1000                              & Till Converge                             & 152.94    & 33.50   & 0.00     & 0.00   & 1.00       & 0.00     & 0.00     & 0.00   \\
	& 1000 & 1000                              & 1                                         & 11.15     & 1.61    & 0.02     & 0.05   & 0.83       & 0.15     & 0.94     & 0.77   \\
	& 1000 & 1000                              & 5                                         & 60.73     & 8.66    & 0.00     & 0.00   & 1.00       & 0.00     & 0.00     & 0.00   \\
	& 1000 & 1000                              & 10                                        & 122.70    & 16.98   & 0.00     & 0.02   & 1.00       & 0.00     & 0.02     & 0.14   \\ \hline
	& D    & \multicolumn{2}{l}{\begin{tabular}[c]{@{}l@{}}Avg Split\\ /Prob\end{tabular}} & \multicolumn{8}{l}{}                                                                \\ \hline
	\multirow{6}{*}{DART} & 10   & \multicolumn{2}{l}{Avg Split}                                                 & 67.86     & 11.51   & 0.00     & 0.02   & 0.98       & 0.06     & 0.12     & 0.33   \\
	& 10   & \multicolumn{2}{l}{Prob}                                                      & 196.43    & 43.41   & 0.00     & 0.00   & 1.00       & 0.00     & 0.00     & 0.00   \\
	& 50   & \multicolumn{2}{l}{Avg Split}                                                 & 12.77     & 2.04    & 0.15     & 0.25   & 0.45       & 0.18     & 3.30     & 1.49   \\
	& 50   & \multicolumn{2}{l}{Prob}                                                      & 196.80    & 43.34   & 0.00     & 0.00   & 1.00       & 0.00     & 0.00     & 0.00   \\
	& 200  & \multicolumn{2}{l}{Avg Split}                                                 & 61.14     & 9.85    & 0.00     & 0.00   & 0.94       & 0.10     & 0.30     & 0.51   \\
	& 200  & \multicolumn{2}{l}{Prob}                                                      & 196.89    & 43.85   & 0.00     & 0.00   & 1.00       & 0.00     & 0.00     & 0.00   \\ \hline
	\multirow{6}{*}{BART} & 10   & \multicolumn{2}{l}{Avg Split}                                                 & 125.31    & 19.75   & 0.00     & 0.02   & 0.98       & 0.05     & 0.10     & 0.30   \\
	& 10   & \multicolumn{2}{l}{Prob}                                                      & 949.38    & 128.93  & 0.63     & 0.06   & 1.00       & 0.00     & 9.00     & 2.18   \\
	& 50   & \multicolumn{2}{l}{Avg Split}                                                 & 1715.54   & 218.52  & 0.37     & 0.13   & 1.00       & 0.00     & 3.32     & 1.66   \\
	& 50   & \multicolumn{2}{l}{Prob}                                                      & 1715.54   & 218.52  & 0.66     & 0.06   & 1.00       & 0.00     & 10.26    & 2.95   \\
	& 200  & \multicolumn{2}{l}{Avg Split}                                                 & 5717.13   & 529.43  & 0.29     & 0.14   & 1.00       & 0.00     & 2.36     & 1.43   \\
	& 200  & \multicolumn{2}{l}{Prob}                                                      & 5717.13   & 529.43  & 0.73     & 0.06   & 1.00       & 0.00     & 14.38    & 3.76   \\ \hline
	SSLASSO               & \multicolumn{3}{l}{}                                                                 & 9.40      & 0.93    & 0.00     & 0.02   & 1.00       & 0.00     & 0.02     & 0.14   \\ \hline
	Alpha                 & \multicolumn{3}{l}{}                                                                 & 4.98      & 0.17    & 0.04     & 0.09   & 1.00       & 0.00     & 0.28     & 0.61   \\ \hline
\end{tabular}
	}
\end{table}
\begin{table}[ht!]
	\caption{Linear Setup with $n=50\,000$ and $n=100\,000$; $s$ is the batch size, $r$ is the number of rounds and $M$ is the number of internal MCMC iterations in TVS.}
	\scalebox{0.7}{
		\begin{tabular}{llllllllllll}
			\hline
			\multicolumn{12}{c}{$n = 50\,000$}                                                                                                            \\ \hline
			Method                & $M$    & $s$    & $r$             & Mean Time & SD Time & Mean FDR & SD FDR & Mean Power & SD Power & Mean Ham & SD Ham \\ \hline
			\multirow{16}{*}{TVS} & 500  & 500  & Till Converge & 89.20     & 31.75   & 0.00     & 0.00   & 0.82       & 0.07     & 0.92     & 0.34   \\
			& 500  & 500  & 1             & 37.34     & 5.51    & 0.04     & 0.08   & 0.76       & 0.11     & 1.38     & 0.64   \\
			& 500  & 500  & 5             & 181.80    & 26.01   & 0.00     & 0.00   & 0.88       & 0.10     & 0.60     & 0.49   \\
			& 500  & 500  & 10            & 364.25    & 51.93   & 0.00     & 0.00   & 0.96       & 0.08     & 0.18     & 0.39   \\
			& 1000 & 500  & Till Converge & 140.49    & 31.04   & 0.00     & 0.00   & 0.80       & 0.03     & 0.98     & 0.14   \\
			& 1000 & 500  & 1             & 66.15     & 9.06    & 0.00     & 0.00   & 0.76       & 0.08     & 1.22     & 0.42   \\
			& 1000 & 500  & 5             & 326.66    & 43.50   & 0.00     & 0.00   & 0.88       & 0.10     & 0.60     & 0.49   \\
			& 1000 & 500  & 10            & 656.69    & 85.25   & 0.00     & 0.00   & 0.97       & 0.07     & 0.16     & 0.37   \\
			& 500  & 1000 & Till Converge & 150.23    & 63.39   & 0.09     & 0.08   & 0.96       & 0.08     & 0.70     & 0.46   \\
			& 500  & 1000 & 1             & 29.93     & 4.74    & 0.08     & 0.11   & 0.76       & 0.13     & 1.56     & 0.76   \\
			& 500  & 1000 & 5             & 151.28    & 21.95   & 0.09     & 0.08   & 0.98       & 0.05     & 0.60     & 0.49   \\
			& 500  & 1000 & 10            & 304.49    & 43.17   & 0.14     & 0.07   & 1.00       & 0.03     & 0.84     & 0.42   \\
			& 1000 & 1000 & Till Converge & 220.03    & 65.18   & 0.02     & 0.05   & 0.96       & 0.08     & 0.30     & 0.46   \\
			& 1000 & 1000 & 1             & 53.29     & 7.90    & 0.04     & 0.08   & 0.81       & 0.14     & 1.14     & 0.67   \\
			& 1000 & 1000 & 5             & 270.67    & 37.17   & 0.02     & 0.05   & 0.99       & 0.04     & 0.14     & 0.35   \\
			& 1000 & 1000 & 10            & 545.72    & 72.54   & 0.02     & 0.06   & 1.00       & 0.00     & 0.14     & 0.35   \\ \hline
			\multicolumn{12}{c}{$n = 100\,000$}                                                                                                           \\ \hline
			\multirow{16}{*}{TVS} & 500  & 500  & Till Converge & 93.14     & 39.67   & 0.00     & 0.02   & 0.82       & 0.07     & 0.92     & 0.34   \\
			& 500  & 500  & 1             & 76.60     & 12.54   & 0.00     & 0.02   & 0.81       & 0.06     & 0.96     & 0.28   \\
			& 500  & 500  & 5             & 379.48    & 58.92   & 0.00     & 0.00   & 0.99       & 0.05     & 0.06     & 0.24   \\
			& 500  & 500  & 10            & 764.01    & 116.22  & 0.00     & 0.00   & 1.00       & 0.03     & 0.02     & 0.14   \\
			& 1000 & 500  & Till Converge & 159.10    & 55.08   & 0.00     & 0.00   & 0.80       & 0.06     & 1.00     & 0.29   \\
			& 1000 & 500  & 1             & 135.46    & 19.98   & 0.00     & 0.00   & 0.79       & 0.04     & 1.04     & 0.20   \\
			& 1000 & 500  & 5             & 677.18    & 98.13   & 0.00     & 0.00   & 0.95       & 0.09     & 0.26     & 0.44   \\
			& 1000 & 500  & 10            & 1363.99   & 200.85  & 0.00     & 0.00   & 0.99       & 0.05     & 0.06     & 0.24   \\
			& 500  & 1000 & Till Converge & 173.36    & 60.13   & 0.07     & 0.08   & 0.97       & 0.07     & 0.58     & 0.57   \\
			& 500  & 1000 & 1             & 62.23     & 10.77   & 0.06     & 0.09   & 0.84       & 0.13     & 1.12     & 0.77   \\
			& 500  & 1000 & 5             & 315.53    & 51.68   & 0.13     & 0.07   & 1.00       & 0.00     & 0.78     & 0.42   \\
			& 500  & 1000 & 10            & 635.09    & 102.95  & 0.15     & 0.05   & 1.00       & 0.00     & 0.92     & 0.27   \\
			& 1000 & 1000 & Till Converge & 293.65    & 96.59   & 0.01     & 0.04   & 0.98       & 0.05     & 0.14     & 0.35   \\
			& 1000 & 1000 & 1             & 111.14    & 18.43   & 0.01     & 0.05   & 0.90       & 0.11     & 0.60     & 0.61   \\
			& 1000 & 1000 & 5             & 566.08    & 85.57   & 0.03     & 0.06   & 1.00       & 0.00     & 0.18     & 0.39   \\
			& 1000 & 1000 & 10            & 1140.65   & 166.47  & 0.09     & 0.08   & 1.00       & 0.00     & 0.52     & 0.50   \\ \cline{1-12} 
		\end{tabular}
	}
\end{table}
\begin{table}[ht!]
	\caption{Liang Setup with $n=10\,000$, BART and DART are implemented using \texttt{Prob} (median probability model) rule or \texttt{Avg Split} (truncating the importance measure at $1$); $s$ is the batch size, $r$ is the number of rounds and $M$ is the number of internal MCMC iterations in TVS.}
	\scalebox{0.7}{
\begin{tabular}{llllllllllll}
	\hline
	Method                & M    & s                                 & r                                         & Mean Time & SD Time & Mean FDR & SD FDR & Mean Power & SD Power & Mean Ham & SD Ham \\ \hline
	\multirow{16}{*}{TVS} & 500  & 500                               & Till Converge                             & 80.68     & 18.66   & 0.00     & 0.02   & 0.99       & 0.04     & 0.06     & 0.24   \\
	& 500  & 500                               & 1                                         & 9.16      & 1.55    & 0.02     & 0.08   & 0.38       & 0.14     & 3.12     & 0.75   \\
	& 500  & 500                               & 5                                         & 39.81     & 6.38    & 0.00     & 0.00   & 0.92       & 0.11     & 0.40     & 0.57   \\
	& 500  & 500                               & 10                                        & 79.97     & 12.62   & 0.00     & 0.00   & 0.99       & 0.05     & 0.06     & 0.24   \\
	& 1000 & 500                               & Till Converge                             & 125.43    & 30.63   & 0.00     & 0.00   & 0.99       & 0.05     & 0.06     & 0.24   \\
	& 1000 & 500                               & 1                                         & 15.31     & 2.49    & 0.01     & 0.07   & 0.53       & 0.18     & 2.40     & 0.93   \\
	& 1000 & 500                               & 5                                         & 72.52     & 10.90   & 0.00     & 0.00   & 0.97       & 0.07     & 0.14     & 0.35   \\
	& 1000 & 500                               & 10                                        & 146.75    & 21.39   & 0.00     & 0.00   & 1.00       & 0.00     & 0.00     & 0.00   \\
	& 500  & 1000                              & Till Converge                             & 124.65    & 37.15   & 0.00     & 0.00   & 1.00       & 0.00     & 0.00     & 0.00   \\
	& 500  & 1000                              & 1                                         & 7.53      & 1.33    & 0.23     & 0.25   & 0.34       & 0.15     & 4.04     & 1.14   \\
	& 500  & 1000                              & 5                                         & 33.59     & 5.93    & 0.00     & 0.02   & 0.84       & 0.14     & 0.80     & 0.70   \\
	& 500  & 1000                              & 10                                        & 67.86     & 11.51   & 0.00     & 0.02   & 0.98       & 0.06     & 0.12     & 0.33   \\
	& 1000 & 1000                              & Till Converge                             & 196.43    & 43.41   & 0.00     & 0.00   & 1.00       & 0.00     & 0.00     & 0.00   \\
	& 1000 & 1000                              & 1                                         & 12.77     & 2.04    & 0.15     & 0.25   & 0.45       & 0.18     & 3.30     & 1.49   \\
	& 1000 & 1000                              & 5                                         & 61.14     & 9.85    & 0.00     & 0.00   & 0.94       & 0.10     & 0.30     & 0.51   \\
	& 1000 & 1000                              & 10                                        & 125.31    & 19.75   & 0.00     & 0.02   & 0.98       & 0.05     & 0.10     & 0.30   \\ \hline
	& D    & \multicolumn{2}{l}{\begin{tabular}[c]{@{}l@{}}Avg Split\\ /Prob\end{tabular}} & \multicolumn{8}{l}{}                                                                \\ \hline
	\multirow{6}{*}{DART} & 10   & \multicolumn{2}{l}{Avg Split}                                                 & 965.45    & 151.88  & 0.85     & 0.05   & 0.58       & 0.18     & 19.10    & 3.31   \\
	& 10   & \multicolumn{2}{l}{Prob}                                                      & 965.45    & 151.88  & 0.87     & 0.04   & 0.58       & 0.18     & 22.90    & 4.07   \\
	& 50   & \multicolumn{2}{l}{Avg Split}                                                 & 1647.39   & 110.13  & 0.55     & 0.18   & 0.98       & 0.05     & 7.76     & 4.63   \\
	& 50   & \multicolumn{2}{l}{Prob}                                                      & 1647.39   & 110.13  & 0.67     & 0.14   & 0.98       & 0.05     & 11.86    & 5.27   \\
	& 200  & \multicolumn{2}{l}{Avg Split}                                                 & 4695.72   & 626.39  & 0.27     & 0.19   & 1.00       & 0.00     & 2.48     & 2.60   \\
	& 200  & \multicolumn{2}{l}{Prob}                                                      & 4695.72   & 626.39  & 0.42     & 0.20   & 1.00       & 0.00     & 5.18     & 5.29   \\ \hline
	\multirow{6}{*}{BART} & 10   & \multicolumn{2}{l}{Avg Split}                                                 & 1525.58   & 221.31  & 0.71     & 0.07   & 1.00       & 0.00     & 13.22    & 3.77   \\
	& 10   & \multicolumn{2}{l}{Prob}                                                      & 1525.58   & 221.31  & 0.91     & 0.01   & 1.00       & 0.00     & 48.66    & 6.26   \\
	& 50   & \multicolumn{2}{l}{Avg Split}                                                 & 2315.53   & 294.10  & 0.71     & 0.07   & 1.00       & 0.00     & 13.34    & 4.35   \\
	& 50   & \multicolumn{2}{l}{Prob}                                                      & 2315.53   & 294.10  & 0.91     & 0.01   & 1.00       & 0.00     & 52.74    & 9.04   \\
	& 200  & \multicolumn{2}{l}{Avg Split}                                                 & 5801.02   & 605.46  & 0.74     & 0.05   & 1.00       & 0.00     & 15.34    & 3.92   \\
	& 200  & \multicolumn{2}{l}{Prob}                                                      & 5801.02   & 605.46  & 0.90     & 0.01   & 1.00       & 0.00     & 46.92    & 6.89   \\ \hline
	SSLASSO               & \multicolumn{3}{l}{}                                                                 & 16.14     & 1.64    & 0.00     & 0.00   & 0.60       & 0.00     & 2.00     & 0.00   \\ \hline
	Alpha                 & \multicolumn{3}{l}{}                                                                 & 4.91      & 0.66    & 0.00     & 0.00   & 0.62       & 0.05     & 1.92     & 0.27   \\ \hline
\end{tabular}
	}
\end{table}
\begin{table}[ht!]
	\caption{Liang Setup with $n=50\,000$ and $n=100\,000$; $s$ is the batch size, $r$ is the number of rounds and $M$ is the number of internal MCMC iterations in TVS.}
	\scalebox{0.7}{
		\begin{tabular}{llllllllllll}
			\hline
			\multicolumn{12}{c}{n = 50000}                                                                                                            \\ \hline
			Method                & M    & s    & r             & Mean Time & SD Time & Mean FDR & SD FDR & Mean Power & SD Power & Mean Ham & SD Ham \\ \hline
			\multirow{16}{*}{TVS} & 500  & 500  & Till Converge & 92.32     & 20.87   & 0.00     & 0.00   & 0.99       & 0.05     & 0.06     & 0.24   \\
			& 500  & 500  & 1             & 40.68     & 6.15    & 0.00     & 0.00   & 0.90       & 0.13     & 0.52     & 0.65   \\
			& 500  & 500  & 5             & 203.82    & 29.86   & 0.00     & 0.00   & 1.00       & 0.00     & 0.00     & 0.00   \\
			& 500  & 500  & 10            & 410.74    & 59.11   & 0.00     & 0.00   & 1.00       & 0.00     & 0.00     & 0.00   \\
			& 1000 & 500  & Till Converge & 157.52    & 33.75   & 0.00     & 0.00   & 0.98       & 0.07     & 0.08     & 0.34   \\
			& 1000 & 500  & 1             & 73.48     & 10.50   & 0.00     & 0.00   & 0.96       & 0.10     & 0.20     & 0.49   \\
			& 1000 & 500  & 5             & 376.54    & 51.66   & 0.00     & 0.00   & 1.00       & 0.00     & 0.00     & 0.00   \\
			& 1000 & 500  & 10            & 759.76    & 103.03  & 0.00     & 0.00   & 1.00       & 0.00     & 0.00     & 0.00   \\
			& 500  & 1000 & Till Converge & 139.35    & 33.68   & 0.00     & 0.00   & 0.99       & 0.05     & 0.06     & 0.24   \\
			& 500  & 1000 & 1             & 33.99     & 5.70    & 0.00     & 0.00   & 0.82       & 0.16     & 0.88     & 0.82   \\
			& 500  & 1000 & 5             & 172.82    & 26.65   & 0.00     & 0.02   & 1.00       & 0.03     & 0.04     & 0.20   \\
			& 500  & 1000 & 10            & 348.24    & 53.31   & 0.00     & 0.00   & 1.00       & 0.00     & 0.00     & 0.00   \\
			& 1000 & 1000 & Till Converge & 221.64    & 57.18   & 0.00     & 0.02   & 1.00       & 0.00     & 0.02     & 0.14   \\
			& 1000 & 1000 & 1             & 61.49     & 10.03   & 0.00     & 0.00   & 0.96       & 0.08     & 0.20     & 0.40   \\
			& 1000 & 1000 & 5             & 320.88    & 47.84   & 0.01     & 0.03   & 1.00       & 0.00     & 0.04     & 0.20   \\
			& 1000 & 1000 & 10            & 649.88    & 92.55   & 0.00     & 0.02   & 1.00       & 0.00     & 0.02     & 0.14   \\ \hline
			\multicolumn{12}{c}{n = 100000}                                                                                                           \\ \hline
			\multirow{16}{*}{TVS} & 500  & 500  & Till Converge & 120.82    & 49.24   & 0.00     & 0.00   & 0.99       & 0.04     & 0.04     & 0.20   \\
			& 500  & 500  & 1             & 86.77     & 13.57   & 0.00     & 0.00   & 0.99       & 0.05     & 0.06     & 0.24   \\
			& 500  & 500  & 5             & 438.57    & 66.16   & 0.00     & 0.00   & 1.00       & 0.00     & 0.00     & 0.00   \\
			& 500  & 500  & 10            & 883.47    & 129.33  & 0.00     & 0.00   & 1.00       & 0.00     & 0.00     & 0.00   \\
			& 1000 & 500  & Till Converge & 175.25    & 59.12   & 0.00     & 0.00   & 0.99       & 0.04     & 0.04     & 0.20   \\
			& 1000 & 500  & 1             & 158.25    & 22.10   & 0.00     & 0.00   & 0.99       & 0.04     & 0.04     & 0.20   \\
			& 1000 & 500  & 5             & 807.36    & 109.70  & 0.00     & 0.00   & 1.00       & 0.00     & 0.00     & 0.00   \\
			& 1000 & 500  & 10            & 1625.51   & 219.19  & 0.00     & 0.00   & 1.00       & 0.00     & 0.00     & 0.00   \\
			& 500  & 1000 & Till Converge & 167.40    & 46.18   & 0.00     & 0.00   & 1.00       & 0.00     & 0.00     & 0.00   \\
			& 500  & 1000 & 1             & 74.04     & 11.86   & 0.01     & 0.03   & 0.98       & 0.06     & 0.14     & 0.35   \\
			& 500  & 1000 & 5             & 374.54    & 56.43   & 0.00     & 0.00   & 1.00       & 0.00     & 0.00     & 0.00   \\
			& 500  & 1000 & 10            & 752.86    & 112.17  & 0.00     & 0.00   & 1.00       & 0.00     & 0.00     & 0.00   \\
			& 1000 & 1000 & Till Converge & 293.39    & 68.52   & 0.00     & 0.00   & 1.00       & 0.00     & 0.00     & 0.00   \\
			& 1000 & 1000 & 1             & 134.11    & 20.44   & 0.00     & 0.00   & 0.98       & 0.05     & 0.08     & 0.27   \\
			& 1000 & 1000 & 5             & 690.95    & 96.24   & 0.00     & 0.00   & 1.00       & 0.00     & 0.00     & 0.00   \\
			& 1000 & 1000 & 10            & 1380.91   & 192.17  & 0.00     & 0.00   & 1.00       & 0.00     & 0.00     & 0.00   \\ \cline{1-12} 
		\end{tabular}
	}
\end{table}
\begin{table}[ht!]
	\caption{Friedman Setup with $n=10\,000$, BART and DART are implemented using \texttt{Prob} (median probability model) rule or \texttt{Avg Split} (truncating the importance measure at $1$); $s$ is the batch size, $r$ is the number of rounds and $M$ is the number of internal MCMC iterations in TVS.}
	\scalebox{0.7}{
		\begin{tabular}{llllllllllll}
			\hline
			Method                & M    & s                                 & r                                         & Mean Time & SD Time & Mean FDR & SD FDR & Mean Power & SD Power & Mean Ham & SD Ham \\ \hline
			\multirow{16}{*}{TVS} & 500  & 500                               & Till Converge                             & 46.26     & 6.71    & 0.00     & 0.00   & 1.00       & 0.00     & 0.00     & 0.00   \\
			& 500  & 500                               & 1                                         & 7.42      & 0.50    & 0.00     & 0.00   & 0.94       & 0.10     & 0.32     & 0.51   \\
			& 500  & 500                               & 5                                         & 35.04     & 2.46    & 0.00     & 0.00   & 1.00       & 0.00     & 0.00     & 0.00   \\
			& 500  & 500                               & 10                                        & 69.44     & 5.05    & 0.00     & 0.00   & 1.00       & 0.00     & 0.00     & 0.00   \\
			& 1000 & 500                               & Till Converge                             & 80.81     & 6.85    & 0.00     & 0.00   & 1.00       & 0.00     & 0.00     & 0.00   \\
			& 1000 & 500                               & 1                                         & 13.08     & 1.01    & 0.00     & 0.00   & 0.98       & 0.06     & 0.10     & 0.30   \\
			& 1000 & 500                               & 5                                         & 65.29     & 4.65    & 0.00     & 0.00   & 1.00       & 0.00     & 0.00     & 0.00   \\
			& 1000 & 500                               & 10                                        & 129.90    & 9.27    & 0.00     & 0.00   & 1.00       & 0.00     & 0.00     & 0.00   \\
			& 500  & 1000                              & Till Converge                             & 110.23    & 41.32   & 0.00     & 0.00   & 1.00       & 0.00     & 0.00     & 0.00   \\
			& 500  & 1000                              & 1                                         & 6.46      & 0.50    & 0.07     & 0.12   & 0.86       & 0.15     & 1.08     & 1.16   \\
			& 500  & 1000                              & 5                                         & 32.55     & 2.33    & 0.01     & 0.04   & 1.00       & 0.00     & 0.06     & 0.24   \\
			& 500  & 1000                              & 10                                        & 65.30     & 4.27    & 0.00     & 0.00   & 1.00       & 0.00     & 0.00     & 0.00   \\
			& 1000 & 1000                              & Till Converge                             & 181.50    & 50.55   & 0.00     & 0.02   & 1.00       & 0.00     & 0.02     & 0.14   \\
			& 1000 & 1000                              & 1                                         & 11.51     & 0.87    & 0.04     & 0.08   & 0.94       & 0.11     & 0.56     & 0.70   \\
			& 1000 & 1000                              & 5                                         & 60.74     & 4.40    & 0.00     & 0.02   & 1.00       & 0.03     & 0.04     & 0.20   \\
			& 1000 & 1000                              & 10                                        & 122.46    & 8.51    & 0.00     & 0.00   & 1.00       & 0.00     & 0.00     & 0.00   \\ \hline
			& D    & \multicolumn{2}{l}{\begin{tabular}[c]{@{}l@{}}Avg Split\\ /Prob\end{tabular}} & \multicolumn{8}{l}{}                                                                \\ \hline
			\multirow{6}{*}{DART} & 10   & \multicolumn{2}{l}{Avg Split}                                                 & 1022.94   & 194.68  & 0.72     & 0.16   & 0.89       & 0.12     & 15.62    & 7.70   \\
			& 10   & \multicolumn{2}{l}{Prob}                                                      & 1022.94   & 194.68  & 0.80     & 0.08   & 0.90       & 0.12     & 21.00    & 7.48   \\
			& 50   & \multicolumn{2}{l}{Avg Split}                                                 & 1442.99   & 136.58  & 0.02     & 0.05   & 1.00       & 0.00     & 0.12     & 0.33   \\
			& 50   & \multicolumn{2}{l}{Prob}                                                      & 1442.99   & 136.58  & 0.08     & 0.14   & 1.00       & 0.00     & 0.66     & 1.61   \\
			& 200  & \multicolumn{2}{l}{Avg Split}                                                 & 4939.07   & 788.75  & 0.00     & 0.00   & 1.00       & 0.00     & 0.00     & 0.00   \\
			& 200  & \multicolumn{2}{l}{Prob}                                                      & 4939.07   & 788.75  & 0.01     & 0.05   & 1.00       & 0.00     & 0.06     & 0.31   \\ \hline
			\multirow{6}{*}{BART} & 10   & \multicolumn{2}{l}{Avg Split}                                                 & 1751.38   & 274.31  & 0.80     & 0.06   & 1.00       & 0.00     & 21.70    & 6.53   \\
			& 10   & \multicolumn{2}{l}{Prob}                                                      & 1751.38   & 274.31  & 0.89     & 0.02   & 1.00       & 0.00     & 43.30    & 9.50   \\
			& 50   & \multicolumn{2}{l}{Avg Split}                                                 & 1861.83   & 229.61  & 0.24     & 0.13   & 1.00       & 0.00     & 1.80     & 1.20   \\
			& 50   & \multicolumn{2}{l}{Prob}                                                      & 1861.83   & 229.61  & 0.57     & 0.11   & 1.00       & 0.00     & 7.44     & 3.47   \\
			& 200  & \multicolumn{2}{l}{Avg Split}                                                 & 6009.07   & 892.87  & 0.21     & 0.14   & 1.00       & 0.00     & 1.56     & 1.26   \\
			& 200  & \multicolumn{2}{l}{Prob}                                                      & 6009.07   & 892.87  & 0.70     & 0.08   & 1.00       & 0.00     & 12.54    & 4.08   \\ \hline
			SSLASSO               & \multicolumn{3}{l}{}                                                                 & 16.75     & 2.62    & 0.00     & 0.00   & 0.80       & 0.00     & 1.00     & 0.00   \\ \hline
			Alpha                 & \multicolumn{3}{l}{}                                                                 & 4.96      & 0.91    & 0.03     & 0.08   & 0.81       & 0.05     & 1.10     & 0.51   \\ \hline
		\end{tabular}
	}
\end{table}
\begin{table}[ht!]
	\caption{Friedman Setup with $n=50\,000$ and $n=100\,000$; $s$ is the batch size, $r$ is the number of rounds and $M$ is the number of internal MCMC iterations in TVS.}
	\scalebox{0.7}{
	\begin{tabular}{llllllllllll}
		\hline
		\multicolumn{12}{c}{n = 50000}                                                                                                            \\ \hline
		Method                & M    & s    & r             & Mean Time & SD Time & Mean FDR & SD FDR & Mean Power & SD Power & Mean Ham & SD Ham \\ \hline
		\multirow{16}{*}{TVS} & 500  & 500  & Till Converge & 71.12     & 8.71    & 0.00     & 0.00   & 1.00       & 0.00     & 0.00     & 0.00   \\
		& 500  & 500  & 1             & 35.49     & 3.52    & 0.00     & 0.00   & 1.00       & 0.00     & 0.00     & 0.00   \\
		& 500  & 500  & 5             & 174.43    & 16.98   & 0.00     & 0.02   & 1.00       & 0.00     & 0.02     & 0.14   \\
		& 500  & 500  & 10            & 347.30    & 35.24   & 0.00     & 0.00   & 1.00       & 0.00     & 0.00     & 0.00   \\
		& 1000 & 500  & Till Converge & 130.57    & 13.44   & 0.00     & 0.00   & 1.00       & 0.00     & 0.00     & 0.00   \\
		& 1000 & 500  & 1             & 65.60     & 6.82    & 0.00     & 0.00   & 1.00       & 0.00     & 0.00     & 0.00   \\
		& 1000 & 500  & 5             & 324.83    & 33.21   & 0.00     & 0.00   & 1.00       & 0.00     & 0.00     & 0.00   \\
		& 1000 & 500  & 10            & 647.04    & 65.02   & 0.00     & 0.00   & 1.00       & 0.00     & 0.00     & 0.00   \\
		& 500  & 1000 & Till Converge & 122.22    & 45.41   & 0.00     & 0.00   & 1.00       & 0.00     & 0.00     & 0.00   \\
		& 500  & 1000 & 1             & 32.26     & 3.27    & 0.01     & 0.03   & 1.00       & 0.00     & 0.04     & 0.20   \\
		& 500  & 1000 & 5             & 162.98    & 16.85   & 0.00     & 0.02   & 1.00       & 0.00     & 0.02     & 0.14   \\
		& 500  & 1000 & 10            & 325.19    & 32.44   & 0.00     & 0.02   & 1.00       & 0.00     & 0.02     & 0.14   \\
		& 1000 & 1000 & Till Converge & 227.68    & 73.92   & 0.00     & 0.02   & 1.00       & 0.00     & 0.02     & 0.14   \\
		& 1000 & 1000 & 1             & 60.99     & 7.37    & 0.00     & 0.02   & 1.00       & 0.03     & 0.04     & 0.20   \\
		& 1000 & 1000 & 5             & 307.25    & 31.56   & 0.00     & 0.00   & 1.00       & 0.00     & 0.00     & 0.00   \\
		& 1000 & 1000 & 10            & 612.30    & 60.46   & 0.00     & 0.02   & 1.00       & 0.00     & 0.02     & 0.14   \\ \hline
		\multicolumn{12}{c}{n = 100000}                                                                                                           \\ \hline
		\multirow{16}{*}{TVS} & 500  & 500  & Till Converge & 70.27     & 6.16    & 0.00     & 0.00   & 1.00       & 0.00     & 0.00     & 0.00   \\
		& 500  & 500  & 1             & 70.27     & 6.16    & 0.00     & 0.00   & 1.00       & 0.00     & 0.00     & 0.00   \\
		& 500  & 500  & 5             & 344.26    & 28.89   & 0.00     & 0.00   & 1.00       & 0.00     & 0.00     & 0.00   \\
		& 500  & 500  & 10            & 680.12    & 48.41   & 0.00     & 0.00   & 1.00       & 0.00     & 0.00     & 0.00   \\
		& 1000 & 500  & Till Converge & 127.92    & 9.35    & 0.00     & 0.00   & 1.00       & 0.00     & 0.00     & 0.00   \\
		& 1000 & 500  & 1             & 127.92    & 9.35    & 0.00     & 0.00   & 1.00       & 0.00     & 0.00     & 0.00   \\
		& 1000 & 500  & 5             & 635.11    & 46.23   & 0.00     & 0.00   & 1.00       & 0.00     & 0.00     & 0.00   \\
		& 1000 & 500  & 10            & 1266.57   & 94.32   & 0.00     & 0.00   & 1.00       & 0.00     & 0.00     & 0.00   \\
		& 500  & 1000 & Till Converge & 160.84    & 45.59   & 0.01     & 0.05   & 1.00       & 0.00     & 0.08     & 0.27   \\
		& 500  & 1000 & 1             & 63.31     & 4.11    & 0.01     & 0.03   & 1.00       & 0.00     & 0.04     & 0.20   \\
		& 500  & 1000 & 5             & 316.52    & 20.79   & 0.00     & 0.00   & 1.00       & 0.00     & 0.00     & 0.00   \\
		& 500  & 1000 & 10            & 633.90    & 37.48   & 0.00     & 0.02   & 1.00       & 0.00     & 0.02     & 0.14   \\
		& 1000 & 1000 & Till Converge & 272.04    & 68.80   & 0.00     & 0.00   & 1.00       & 0.00     & 0.00     & 0.00   \\
		& 1000 & 1000 & 1             & 119.85    & 7.15    & 0.00     & 0.00   & 1.00       & 0.00     & 0.00     & 0.00   \\
		& 1000 & 1000 & 5             & 599.73    & 35.98   & 0.00     & 0.02   & 1.00       & 0.00     & 0.02     & 0.14   \\
		& 1000 & 1000 & 10            & 1192.89   & 68.52   & 0.00     & 0.02   & 1.00       & 0.00     & 0.02     & 0.14   \\ \cline{1-12} 
	\end{tabular}
	}
\end{table}
\begin{table}[ht!]
	\caption{Forest Setup with $n=10\,000$, BART and DART are implemented using \texttt{Prob} (median probability model) rule or \texttt{Avg Split} (truncating the importance measure at $1$); $s$ is the batch size, $r$ is the number of rounds and $M$ is the number of internal MCMC iterations in TVS.}
	\scalebox{0.7}{
		\begin{tabular}{llllllllllll}
			\hline
			Method                & M    & s                                 & r                                         & Mean Time & SD Time & Mean FDR & SD FDR & Mean Power & SD Power & Mean Ham & SD Ham \\ \hline
			\multirow{16}{*}{TVS} & 500  & 500                               & Till Converge                             & 52.80     & 12.24   & 0.00     & 0.02   & 1.00       & 0.00     & 0.02     & 0.14   \\
			& 500  & 500                               & 1                                         & 8.33      & 1.21    & 0.00     & 0.00   & 0.96       & 0.09     & 0.18     & 0.44   \\
			& 500  & 500                               & 5                                         & 40.29     & 5.31    & 0.00     & 0.00   & 1.00       & 0.00     & 0.00     & 0.00   \\
			& 500  & 500                               & 10                                        & 79.77     & 10.26   & 0.00     & 0.00   & 1.00       & 0.00     & 0.00     & 0.00   \\
			& 1000 & 500                               & Till Converge                             & 88.68     & 11.37   & 0.00     & 0.00   & 1.00       & 0.00     & 0.00     & 0.00   \\
			& 1000 & 500                               & 1                                         & 14.79     & 1.95    & 0.00     & 0.00   & 0.98       & 0.05     & 0.08     & 0.27   \\
			& 1000 & 500                               & 5                                         & 73.30     & 9.15    & 0.00     & 0.00   & 1.00       & 0.00     & 0.00     & 0.00   \\
			& 1000 & 500                               & 10                                        & 145.98    & 18.12   & 0.00     & 0.00   & 1.00       & 0.00     & 0.00     & 0.00   \\
			& 500  & 1000                              & Till Converge                             & 91.11     & 23.10   & 0.00     & 0.00   & 1.00       & 0.00     & 0.00     & 0.00   \\
			& 500  & 1000                              & 1                                         & 6.42      & 1.01    & 0.03     & 0.09   & 0.69       & 0.19     & 1.68     & 1.00   \\
			& 500  & 1000                              & 5                                         & 33.02     & 5.05    & 0.00     & 0.00   & 1.00       & 0.00     & 0.00     & 0.00   \\
			& 500  & 1000                              & 10                                        & 66.82     & 9.69    & 0.00     & 0.02   & 1.00       & 0.00     & 0.02     & 0.14   \\
			& 1000 & 1000                              & Till Converge                             & 152.94    & 33.50   & 0.00     & 0.00   & 1.00       & 0.00     & 0.00     & 0.00   \\
			& 1000 & 1000                              & 1                                         & 11.15     & 1.61    & 0.02     & 0.05   & 0.83       & 0.15     & 0.94     & 0.77   \\
			& 1000 & 1000                              & 5                                         & 60.73     & 8.66    & 0.00     & 0.00   & 1.00       & 0.00     & 0.00     & 0.00   \\
			& 1000 & 1000                              & 10                                        & 122.70    & 16.98   & 0.00     & 0.02   & 1.00       & 0.00     & 0.02     & 0.14   \\ \hline
			& D    & \multicolumn{2}{l}{\begin{tabular}[c]{@{}l@{}}Avg Split\\ /Prob\end{tabular}} & \multicolumn{8}{l}{}                                                                \\ \hline
			\multirow{6}{*}{DART} & 10   & \multicolumn{2}{l}{Avg Split}                                                 & 509.13    & 65.73   & 0.13     & 0.13   & 0.93       & 0.10     & 1.14     & 1.14   \\
			& 10   & \multicolumn{2}{l}{Prob}                                                      & 509.13    & 65.73   & 0.14     & 0.14   & 0.93       & 0.10     & 1.22     & 1.17   \\
			& 50   & \multicolumn{2}{l}{Avg Split}                                                 & 1132.98   & 190.56  & 0.16     & 0.16   & 1.00       & 0.00     & 1.18     & 1.34   \\
			& 50   & \multicolumn{2}{l}{Prob}                                                      & 1132.98   & 190.56  & 0.32     & 0.16   & 1.00       & 0.00     & 2.78     & 1.80   \\
			& 200  & \multicolumn{2}{l}{Avg Split}                                                 & 4028.27   & 931.10  & 0.59     & 0.09   & 1.00       & 0.00     & 7.76     & 2.92   \\
			& 200  & \multicolumn{2}{l}{Prob}                                                      & 4028.27   & 931.10  & 0.68     & 0.07   & 1.00       & 0.00     & 11.18    & 3.35   \\ \hline
			\multirow{6}{*}{BART} & 10   & \multicolumn{2}{l}{Avg Split}                                                 & 1099.59   & 270.16  & 0.34     & 0.16   & 1.00       & 0.00     & 3.08     & 1.91   \\
			& 10   & \multicolumn{2}{l}{Prob}                                                      & 1099.59   & 270.16  & 0.68     & 0.05   & 1.00       & 0.00     & 11.14    & 2.60   \\
			& 50   & \multicolumn{2}{l}{Avg Split}                                                 & 1857.85   & 280.43  & 0.47     & 0.15   & 1.00       & 0.00     & 5.08     & 2.30   \\
			& 50   & \multicolumn{2}{l}{Prob}                                                      & 1857.85   & 280.43  & 0.82     & 0.03   & 1.00       & 0.00     & 23.26    & 3.86   \\
			& 200  & \multicolumn{2}{l}{Avg Split}                                                 & 5620.33   & 847.79  & 0.83     & 0.03   & 1.00       & 0.00     & 25.50    & 4.81   \\
			& 200  & \multicolumn{2}{l}{Prob}                                                      & 5620.33   & 847.79  & 0.91     & 0.01   & 1.00       & 0.00     & 53.72    & 6.32   \\ \hline
			SSLASSO               & \multicolumn{3}{l}{}                                                                 & 1.26      & 0.16    & 0.00     & 0.00   & 1.00       & 0.00     & 0.00     & 0.00   \\ \hline
			Alpha                 & \multicolumn{3}{l}{}                                                                 & 5.03      & 1.08    & 0.11     & 0.13   & 1.00       & 0.00     & 0.78     & 1.02   \\ \hline
		\end{tabular}
	}
\end{table}

\begin{table}[ht!]
	\caption{Forest Setup with $n=50\,000$ and $n=100\,000$; $s$ is the batch size, $r$ is the number of rounds and $M$ is the number of internal MCMC iterations in TVS.}
	\scalebox{0.7}{
		\begin{tabular}{llllllllllll}
			\hline
			\multicolumn{12}{c}{$n = 50\,000$}                                                                                                            \\ \hline
			Method                & M    & s    & r             & Mean Time & SD Time & Mean FDR & SD FDR & Mean Power & SD Power & Mean Ham & SD Ham \\ \hline
			\multirow{16}{*}{TVS} & 500  & 500  & Till Converge & 90.28     & 13.28   & 0.00     & 0.00   & 1.00       & 0.00     & 0.00     & 0.00   \\
			& 500  & 500  & 1             & 45.60     & 6.80    & 0.00     & 0.00   & 1.00       & 0.00     & 0.00     & 0.00   \\
			& 500  & 500  & 5             & 224.06    & 32.44   & 0.00     & 0.00   & 1.00       & 0.00     & 0.00     & 0.00   \\
			& 500  & 500  & 10            & 446.61    & 63.34   & 0.00     & 0.00   & 1.00       & 0.00     & 0.00     & 0.00   \\
			& 1000 & 500  & Till Converge & 163.70    & 22.66   & 0.00     & 0.00   & 1.00       & 0.00     & 0.00     & 0.00   \\
			& 1000 & 500  & 1             & 82.49     & 11.65   & 0.00     & 0.00   & 1.00       & 0.00     & 0.00     & 0.00   \\
			& 1000 & 500  & 5             & 408.11    & 55.52   & 0.00     & 0.00   & 1.00       & 0.00     & 0.00     & 0.00   \\
			& 1000 & 500  & 10            & 812.39    & 111.83  & 0.00     & 0.00   & 1.00       & 0.00     & 0.00     & 0.00   \\
			& 500  & 1000 & Till Converge & 117.99    & 23.19   & 0.00     & 0.00   & 1.00       & 0.00     & 0.00     & 0.00   \\
			& 500  & 1000 & 1             & 37.73     & 6.62    & 0.00     & 0.02   & 0.99       & 0.04     & 0.06     & 0.24   \\
			& 500  & 1000 & 5             & 189.66    & 30.05   & 0.00     & 0.02   & 1.00       & 0.00     & 0.02     & 0.14   \\
			& 500  & 1000 & 10            & 377.06    & 60.06   & 0.00     & 0.00   & 1.00       & 0.00     & 0.00     & 0.00   \\
			& 1000 & 1000 & Till Converge & 211.24    & 31.96   & 0.00     & 0.02   & 1.00       & 0.00     & 0.02     & 0.14   \\
			& 1000 & 1000 & 1             & 68.27     & 11.64   & 0.00     & 0.02   & 1.00       & 0.00     & 0.02     & 0.14   \\
			& 1000 & 1000 & 5             & 341.68    & 54.73   & 0.00     & 0.00   & 1.00       & 0.00     & 0.00     & 0.00   \\
			& 1000 & 1000 & 10            & 676.40    & 105.33  & 0.00     & 0.02   & 1.00       & 0.00     & 0.02     & 0.14   \\ \hline
			\multicolumn{12}{c}{$n = 100\,000$}                                                                                                           \\ \hline
			\multirow{16}{*}{TVS} & 500  & 500  & Till Converge & 81.54     & 11.43   & 0.00     & 0.00   & 1.00       & 0.00     & 0.00     & 0.00   \\
			& 500  & 500  & 1             & 81.54     & 11.43   & 0.00     & 0.00   & 1.00       & 0.00     & 0.00     & 0.00   \\
			& 500  & 500  & 5             & 398.59    & 49.39   & 0.00     & 0.00   & 1.00       & 0.00     & 0.00     & 0.00   \\
			& 500  & 500  & 10            & 796.94    & 98.24   & 0.00     & 0.00   & 1.00       & 0.00     & 0.00     & 0.00   \\
			& 1000 & 500  & Till Converge & 146.05    & 17.39   & 0.00     & 0.00   & 1.00       & 0.00     & 0.00     & 0.00   \\
			& 1000 & 500  & 1             & 146.05    & 17.39   & 0.00     & 0.00   & 1.00       & 0.00     & 0.00     & 0.00   \\
			& 1000 & 500  & 5             & 718.56    & 79.50   & 0.00     & 0.00   & 1.00       & 0.00     & 0.00     & 0.00   \\
			& 1000 & 500  & 10            & 1457.97   & 130.28  & 0.00     & 0.00   & 1.00       & 0.00     & 0.00     & 0.00   \\
			& 500  & 1000 & Till Converge & 142.28    & 20.80   & 0.00     & 0.00   & 1.00       & 0.00     & 0.00     & 0.00   \\
			& 500  & 1000 & 1             & 69.61     & 8.27    & 0.00     & 0.02   & 1.00       & 0.00     & 0.02     & 0.14   \\
			& 500  & 1000 & 5             & 348.10    & 37.10   & 0.00     & 0.00   & 1.00       & 0.00     & 0.00     & 0.00   \\
			& 500  & 1000 & 10            & 692.50    & 72.61   & 0.00     & 0.00   & 1.00       & 0.00     & 0.00     & 0.00   \\
			& 1000 & 1000 & Till Converge & 256.28    & 40.84   & 0.00     & 0.00   & 1.00       & 0.00     & 0.00     & 0.00   \\
			& 1000 & 1000 & 1             & 125.82    & 14.09   & 0.00     & 0.02   & 1.00       & 0.00     & 0.02     & 0.14   \\
			& 1000 & 1000 & 5             & 622.97    & 59.21   & 0.00     & 0.00   & 1.00       & 0.00     & 0.00     & 0.00   \\
			& 1000 & 1000 & 10            & 1251.07   & 133.65  & 0.00     & 0.00   & 1.00       & 0.00     & 0.00     & 0.00   \\ \cline{1-12} 
		\end{tabular}
	}
\end{table}
	\newpage
	\section{Additional Results for the HIV Dataset}\label{sec:supplement_hiv}
	In this section, we show additional results on the analysis of the HIV dataset. First, we present some basic statistics about the data. The entire data comes from Stanford HIV Drug Resistance Database. The raw data can be downloaded from \url{https://hivdb.stanford.edu/pages/published_analysis/genophenoPNAS2006/}. \cite{barber_controlling_2015} published cleaning codes (available at  \url{https://web.stanford.edu/group/candes/knockoffs/software/knockoffs/tutorial-4-r.html}) which we adopt.  We provide a basic overview of the dataset in Table \ref{tab:HIV_stats}. 
	
	\begin{table}[ht!]

		\centering
		\scalebox{0.8}{	\begin{tabular}{lllll}
				HIV Virus Life Cycle   & Drug Class  & Mean Log DS & Number of Mutations & Number of Samples \\ \hline
				\multirow{7}{*}{PI}    & APV         & 0.75     & 201                & 767               \\
				& ATV         & 1.59     & 147                & 328               \\
				& IDV         & 1.33     & 206                & 825               \\
				& LPV         & 1.74     & 184                & 515               \\
				& NFV         & 2.00     & 207                & 842               \\
				& RTV        & 1.72     & 205                & 793               \\
				& SQV        & 1.22     & 206                & 824               \\ \hline
				\multirow{6}{*}{NRTI}  & X3TC       & 3.10     & 283                & 629               \\
				& ABC        & 1.14     & 283                & 623               \\
				& AZT        & 1.55     & 283                & 626               \\
				& D4T        & 0.43     & 281                & 625               \\
				& DDI        & 0.43     & 283                & 628               \\
				& TDF        & 0.22     & 215                & 351               \\ \hline
				\multirow{3}{*}{NNRTI} & DLV        & 0.98     & 305                & 730               \\
				& EFV        & 1.08     & 312                & 732               \\
				& NVP        & 1.80     & 313                & 744               \\ \hline
		\end{tabular}}
		\caption{Basic summary statistics of the HIV dataset. DS refers to the decrease in susceptibility of the drug once the mutations has occurred.}\label{tab:HIV_stats}
	\end{table}
	
	In Section \ref{sec:hiv}, we illustrated TVS  on only the drug LPV. Here, we present the rest of the results. As is done in \cite{barber_controlling_2015}, we record  both the number of verified positions discovered (True Discoveries) and the number of discovered unverified positions (False Discoveries)  for each of the five methods. 
	
	\begin{table}[H]
		\caption{ PI Drugs }
		\centering
		\scalebox{0.8}{	\begin{tabular}{llllll}
				\hline
				Methods           & {\color[HTML]{000000} Knockoff} & {\color[HTML]{000000} LASSO} & {\color[HTML]{000000} DART} & {\color[HTML]{000000} SSLASSO} & {\color[HTML]{000000} TVS} \\ \hline
				\multicolumn{6}{c}{APV}                                                                                                                                                        \\ \hline
				True Discoveries  & {\color[HTML]{000000} 19}       & {\color[HTML]{000000} 20}    & {\color[HTML]{000000} 16}   & {\color[HTML]{000000} 8}       & {\color[HTML]{000000} 18}  \\
				False Discoveries & {\color[HTML]{000000} 3}        & {\color[HTML]{000000} 3}     & {\color[HTML]{000000} 0}    & {\color[HTML]{000000} 0}       & {\color[HTML]{000000} 1}   \\ \hline
				\multicolumn{6}{c}{ATV}                                                                                                                                                        \\ \hline
				True Discoveries  & {\color[HTML]{000000} 22}       & {\color[HTML]{000000} 29}    & {\color[HTML]{000000} 19}   & {\color[HTML]{000000} 6}       & {\color[HTML]{000000} 20}  \\
				False Discoveries & {\color[HTML]{000000} 8}        & {\color[HTML]{000000} 20}    & {\color[HTML]{000000} 0}    & {\color[HTML]{000000} 0}       & {\color[HTML]{000000} 0}   \\ \hline
				\multicolumn{6}{c}{IDV}                                                                                                                                                        \\ \hline
				True Discoveries  & {\color[HTML]{000000} 19}       & {\color[HTML]{000000} 28}    & {\color[HTML]{000000} 18}   & {\color[HTML]{000000} 5}       & {\color[HTML]{000000} 17}  \\
				False Discoveries & {\color[HTML]{000000} 12}       & {\color[HTML]{000000} 24}    & {\color[HTML]{000000} 3}    & {\color[HTML]{000000} 0}       & {\color[HTML]{000000} 4}   \\ \hline
				\multicolumn{6}{c}{LPV}                                                                                                                                                        \\ \hline
				True Discoveries  & {\color[HTML]{000000} 16}       & {\color[HTML]{000000} 30}    & {\color[HTML]{000000} 15}   & {\color[HTML]{000000} 4}       & {\color[HTML]{000000} 17}  \\
				False Discoveries & {\color[HTML]{000000} 1}        & {\color[HTML]{000000} 22}    & {\color[HTML]{000000} 1}    & {\color[HTML]{000000} 0}       & {\color[HTML]{000000} 1}   \\ \hline
				\multicolumn{6}{c}{NFV}                                                                                                                                                        \\ \hline
				True Discoveries  & {\color[HTML]{000000} 21}       & {\color[HTML]{000000} 23}    & {\color[HTML]{000000} 19}   & {\color[HTML]{000000} 6}       & {\color[HTML]{000000} 20}  \\
				False Discoveries & {\color[HTML]{000000} 2}        & {\color[HTML]{000000} 2}     & {\color[HTML]{000000} 2}    & {\color[HTML]{000000} 0}       & {\color[HTML]{000000} 3}   \\ \hline
				\multicolumn{6}{c}{RTV}                                                                                                                                                        \\ \hline
				True Discoveries  & {\color[HTML]{000000} 19}       & {\color[HTML]{000000} 34}    & {\color[HTML]{000000} 16}   & {\color[HTML]{000000} 4}       & {\color[HTML]{000000} 18}  \\
				False Discoveries & {\color[HTML]{000000} 8}        & {\color[HTML]{000000} 29}    & {\color[HTML]{000000} 4}    & {\color[HTML]{000000} 0}       & {\color[HTML]{000000} 3}   \\ \hline
				\multicolumn{6}{c}{SQV}                                                                                                                                                        \\ \hline
				True Discoveries  & {\color[HTML]{000000} 17}       & {\color[HTML]{000000} 22}    & {\color[HTML]{000000} 17}   & {\color[HTML]{000000} 5}       & {\color[HTML]{000000} 17}  \\
				False Discoveries & {\color[HTML]{000000} 4}        & {\color[HTML]{000000} 8}     & {\color[HTML]{000000} 1}    & {\color[HTML]{000000} 0}       & {\color[HTML]{000000} 2}   \\ \hline
		\end{tabular}}
	\end{table}
	
	\begin{table}[H]
		\caption{NRTI Drugs}
		\centering
		\scalebox{0.8}{	\begin{tabular}{llllll}
				\hline
				Methods           & {\color[HTML]{000000} Knockoff} & {\color[HTML]{000000} LASSO} & {\color[HTML]{000000} DART} & {\color[HTML]{000000} SSLASSO} & {\color[HTML]{000000} TVS} \\ \hline
				\multicolumn{6}{c}{ABC}                                                                                                                                                        \\ \hline
				True Discoveries  & {\color[HTML]{000000} 10}       & {\color[HTML]{000000} 15}    & {\color[HTML]{000000} 10}   & {\color[HTML]{000000} 11}      & {\color[HTML]{000000} 14}  \\
				False Discoveries & {\color[HTML]{000000} 1}        & {\color[HTML]{000000} 7}     & {\color[HTML]{000000} 4}    & {\color[HTML]{000000} 3}       & {\color[HTML]{000000} 6}   \\ \hline
				\multicolumn{6}{c}{AZT}                                                                                                                                                        \\ \hline
				True Discoveries  & {\color[HTML]{000000} 16}       & {\color[HTML]{000000} 23}    & {\color[HTML]{000000} 15}   & {\color[HTML]{000000} 5}       & {\color[HTML]{000000} 18}  \\
				False Discoveries & {\color[HTML]{000000} 4}        & {\color[HTML]{000000} 51}    & {\color[HTML]{000000} 5}    & {\color[HTML]{000000} 0}       & {\color[HTML]{000000} 8}   \\ \hline
				\multicolumn{6}{c}{D4T}                                                                                                                                                        \\ \hline
				True Discoveries  & {\color[HTML]{000000} 6}        & {\color[HTML]{000000} 13}    & {\color[HTML]{000000} 13}   & {\color[HTML]{000000} 12}      & {\color[HTML]{000000} 13}  \\
				False Discoveries & {\color[HTML]{000000} 1}        & {\color[HTML]{000000} 2}     & {\color[HTML]{000000} 5}    & {\color[HTML]{000000} 4}       & {\color[HTML]{000000} 6}   \\ \hline
				\multicolumn{6}{c}{DDI}                                                                                                                                                        \\ \hline
				True Discoveries  & {\color[HTML]{000000} 0}        & {\color[HTML]{000000} 23}    & {\color[HTML]{000000} 11}   & {\color[HTML]{000000} 17}      & {\color[HTML]{000000} 12}  \\
				False Discoveries & {\color[HTML]{000000} 0}        & {\color[HTML]{000000} 34}    & {\color[HTML]{000000} 6}    & {\color[HTML]{000000} 7}       & {\color[HTML]{000000} 6}   \\ \hline
				\multicolumn{6}{c}{TDF}                                                                                                                                                        \\ \hline
				True Discoveries  & {\color[HTML]{000000} 0}        & {\color[HTML]{000000} 13}    & {\color[HTML]{000000} 14}   & {\color[HTML]{000000} 7}       & {\color[HTML]{000000} 12}  \\
				False Discoveries & {\color[HTML]{000000} 0}        & {\color[HTML]{000000} 9}     & {\color[HTML]{000000} 7}    & {\color[HTML]{000000} 1}       & {\color[HTML]{000000} 6}   \\
				\hline
				\multicolumn{6}{c}{X3TC}                                                                                                                                                        \\ \hline
				True Discoveries  & {\color[HTML]{000000} 0}        & {\color[HTML]{000000} 14}    & {\color[HTML]{000000} 11}   & {\color[HTML]{000000}1}       & {\color[HTML]{000000} 11}  \\
				False Discoveries & {\color[HTML]{000000} 0}        & {\color[HTML]{000000} 8}     & {\color[HTML]{000000} 4}    & {\color[HTML]{000000}0}       & {\color[HTML]{000000} 6}   \\
				\hline
		\end{tabular}}
	\end{table} 
	
	\begin{table}[H]
		\caption{NNRTI Drugs}
		\centering
		\scalebox{0.8}{	\begin{tabular}{llllll}
				\hline
				Methods           & {\color[HTML]{000000} Knockoff} & {\color[HTML]{000000} LASSO} & {\color[HTML]{000000} DART} & {\color[HTML]{000000} SSLASSO} & {\color[HTML]{000000} TVS} \\ \hline
				\multicolumn{6}{c}{DLV}                                                                                                                                                        \\ \hline
				True Discoveries  & {\color[HTML]{000000} 10}       & {\color[HTML]{000000} 11}    & {\color[HTML]{000000} 9}    & {\color[HTML]{000000} 3}       & {\color[HTML]{000000} 9}   \\
				False Discoveries & {\color[HTML]{000000} 14}       & {\color[HTML]{000000} 31}    & {\color[HTML]{000000} 6}    & {\color[HTML]{000000} 1}       & {\color[HTML]{000000} 9}   \\ \hline
				\multicolumn{6}{c}{EFV}                                                                                                                                                        \\ \hline
				True Discoveries  & {\color[HTML]{000000} 11}       & {\color[HTML]{000000} 14}    & {\color[HTML]{000000} 10}   & {\color[HTML]{000000} 5}       & {\color[HTML]{000000} 11}  \\
				False Discoveries & {\color[HTML]{000000} 11}       & {\color[HTML]{000000} 83}    & {\color[HTML]{000000} 5}    & {\color[HTML]{000000} 0}       & {\color[HTML]{000000} 6}   \\ \hline
				\multicolumn{6}{c}{NVP}                                                                                                                                                        \\ \hline
				True Discoveries  & {\color[HTML]{000000} 7}        & {\color[HTML]{000000} 14}    & {\color[HTML]{000000} 8}    & {\color[HTML]{000000} 6}       & {\color[HTML]{000000} 9}   \\
				False Discoveries & {\color[HTML]{000000} 10}       & {\color[HTML]{000000} 87}    & {\color[HTML]{000000} 6}    & {\color[HTML]{000000} 0}       & {\color[HTML]{000000} 11}  \\ \hline
		\end{tabular}}
	\end{table}
	
	\newpage
	
	\section{Details ont the Marketing Data}
	\label{sec:additional_marketing_data}
	{\bf Summary of Predictor Variables} 
	\begin{table}[H]
		\small
		\scalebox{0.75}{\begin{tabular}{|l|l|l|l|l|l|l|}
				\hline
				Name                                                                 & Indicator & Mean   & Sd     & Max      & Min      & Skewness \\ \hline
				mail indicator                                                       & 1         & 0.50   & NA     & NA       & NA       & NA       \\ \hline
				largest sale amount                                                  & 0         & 169.02 & 286.77 & 7999.99  & -3999.99 & 4.63     \\ \hline
				count of product categories that make up 20\% or more of total sales & 0         & 2.15   & 0.37   & 4.00     & 2.00     & 2.10     \\ \hline
				number of months since first esp or first esp return                 & 0         & 30.14  & 16.05  & 60.00    & 2.00     & 0.17     \\ \hline
				number of months since most recent esp purchase or return            & 0         & 23.00  & 14.22  & 60.00    & 2.00     & 0.60     \\ \hline
				day of first purchase: weekend day                                   & 1         & 0.37   & NA     & NA       & NA       & NA       \\ \hline
				day of first purchase: weekday day                                   & 1         & 0.63   & NA     & NA       & NA       & NA       \\ \hline
				total number of sales in previous 12 months                          & 0         & 5.09   & 4.76   & 189.00   & 2.00     & 5.79     \\ \hline
				total number of sales in previous 24 months                          & 0         & 5.95   & 6.16   & 270.00   & 2.00     & 6.23     \\ \hline
				total number of sales in previous 36 months                          & 0         & 6.53   & 7.10   & 344.00   & 2.00     & 6.44     \\ \hline
				total number of large ticket items in previous 12 months             & 0         & 2.37   & 0.94   & 21.00    & 2.00     & 6.95     \\ \hline
				total number of large ticket items in previous 24 months             & 0         & 2.50   & 1.08   & 32.00    & 2.00     & 5.79     \\ \hline
				total number of large ticket items in previous 36 months             & 0         & 2.58   & 1.24   & 46.00    & 2.00     & 6.94     \\ \hline
				total number of large ticket items in previous 60 months             & 0         & 2.80   & 1.47   & 50.00    & -1.00    & 5.37     \\ \hline
				total number of medium ticket items in previous 12 months            & 0         & 3.06   & 1.92   & 46.00    & 2.00     & 4.62     \\ \hline
				total number of medium ticket items in previous 24 months            & 0         & 3.43   & 2.49   & 81.00    & 2.00     & 5.10     \\ \hline
				total number of medium ticket items in previous 36 months            & 0         & 3.71   & 2.90   & 124.00   & 2.00     & 5.48     \\ \hline
				total number of medium ticket items in previous 60 months            & 0         & 4.36   & 3.68   & 198.00   & 2.00     & 6.39     \\ \hline
				total number of small ticket items in previous 12 months             & 0         & 4.49   & 7.83   & 507.00   & 2.00     & 34.98    \\ \hline
				total number of small ticket items in previous 24 months             & 0         & 5.14   & 10.18  & 667.00   & 2.00     & 29.59    \\ \hline
				total number of small ticket items in previous 36 months             & 0         & 5.54   & 10.73  & 667.00   & 2.00     & 25.12    \\ \hline
				total number of small ticket items in previous 60 months             & 0         & 6.55   & 12.73  & 673.00   & 2.00     & 20.36    \\ \hline
				total sales amount in previous 12 months                             & 0         & 524.90 & 766.12 & 18380.40 & -437.80  & 4.31     \\ \hline
				total sales amount in previous 24 months                             & 0         & 650.24 & 935.70 & 33664.26 & -250.01  & 4.36     \\ \hline
				count of unique categories in previous 12 months                     & 0         & 2.94   & 1.28   & 12.00    & 2.00     & 1.78     \\ \hline
				count of unique categories in previous 24 months                     & 0         & 3.15   & 1.47   & 12.00    & 2.00     & 1.62     \\ \hline
				count of unique class numbers in previous 12 months                  & 0         & 4.13   & 2.84   & 55.00    & 2.00     & 3.00     \\ \hline
				count of unique class numbers in previous 24 months                  & 0         & 4.74   & 3.58   & 60.00    & 2.00     & 3.00     \\ \hline
				percent gift cards category sales of total sales                     & 0         & 0.09   & 0.13   & 1.03     & 0.00     & 3.22     \\ \hline
				percent home ins category sales of total sales                       & 0         & 0.02   & 0.05   & 0.42     & 0.00     & 4.48     \\ \hline
				percent imaging category sales of total sales                        & 0         & 0.29   & 0.28   & 1.00     & -0.20    & 1.08     \\ \hline
				percent mobile category sales of total sales                         & 0         & 0.35   & 0.26   & 1.00     & -0.03    & 0.60     \\ \hline
				percent music  category sales of total sales                         & 0         & 0.18   & 0.21   & 1.00     & 0.00     & 1.72     \\ \hline
				percent other  category sales of total sales                         & 0         & 0.36   & 0.33   & 0.99     & 0.01     & 0.63     \\ \hline
				percent pc hardware category sales of total sales                    & 0         & 0.50   & 0.30   & 1.49     & -2.88    & -0.02    \\ \hline
				percent pst category sales of total sales                            & 0         & 0.17   & 0.19   & 1.11     & -0.49    & 1.89     \\ \hline
				percent tv category sales of total sales                             & 0         & 0.40   & 0.29   & 1.00     & -0.03    & 0.38     \\ \hline
				percent vcr category sales of total sales                            & 0         & 0.29   & 0.25   & 3.13     & -0.16    & 1.04     \\ \hline
				percent wireless category sales of total sales                       & 0         & 0.24   & 0.24   & 1.45     & -0.06    & 1.26     \\ \hline
				percent audio category sales of total sales                          & 0         & 0.24   & 0.24   & 1.12     & -0.08    & 1.17     \\ \hline
				percent dss category sales of total sales                            & 0         & 0.32   & 0.28   & 1.00     & 0.00     & 0.94     \\ \hline
				largest return amount                                                & 0         & 176.57 & 273.43 & 4999.99  & -2699.99 & 4.14     \\ \hline
				number of months since oldest return                                 & 0         & 26.20  & 15.33  & 56.00    & 2.00     & 0.28     \\ \hline
				number of months since most recent return                            & 0         & 20.61  & 13.93  & 56.00    & 2.00     & 0.67     \\ \hline
				number of distinct merchandise classes returned                      & 0         & 2.35   & 0.78   & 11.00    & 2.00     & 3.61     \\ \hline
				total return amount in previous 12 months                            & 0         & 301.28 & 530.07 & 24926.85 & 0.01     & 10.03    \\ \hline
				total return amount in previous 24 months                            & 0         & 326.14 & 584.04 & 31826.61 & 0.01     & 12.74    \\ \hline
				total number of items returned in previous 12 months                 & 0         & 3.16   & 2.87   & 100.00   & 2.00     & 17.05    \\ \hline
				total number of items returned in previous 24 months                 & 0         & 3.40   & 5.50   & 527.00   & 2.00     & 63.78    \\ \hline
				number of months shopped once in previous 12 months                  & 0         & 2.59   & 1.07   & 12.00    & 2.00     & 2.89     \\ \hline
				number of months shopped once in previous 24 months                  & 0         & 3.01   & 1.67   & 24.00    & 2.00     & 3.19     \\ \hline
				count of unique purchase trips in previous 12 months                 & 0         & 2.95   & 1.97   & 81.00    & 2.00     & 8.23     \\ \hline
				count of unique purchase trips in previous 24 months                 & 0         & 3.41   & 2.74   & 126.00   & 2.00     & 8.76     \\ \hline
				total number of items purchased in previous 12 months                & 0         & 5.23   & 7.24   & 513.00   & 2.00     & 28.99    \\ \hline
				total number of items purchased in previous 24 months                & 0         & 6.17   & 9.55   & 672.00   & 2.00     & 24.33    \\ \hline
				total number of weekday items in previous 12 months                  & 0         & 4.65   & 6.68   & 506.00   & 2.00     & 35.47    \\ \hline
		\end{tabular}}
	\end{table}
	
	\newpage
	
	\begin{table}[H]
			\small
	\scalebox{0.75}{
		\begin{tabular}{|l|l|l|l|l|l|l|}
			\hline
			Name                                                                  & Indicator & Mean   & Sd     & Max      & Min     & Skewness \\ \hline
			total number of weekend items in previous 12 months                   & 0         & 4.15   & 5.18   & 403.00   & 2.00    & 40.58    \\ \hline
			total number of weekend items in previous 24 months                   & 0         & 4.62   & 6.76   & 504.00   & 2.00    & 34.99    \\ \hline
			total christmas sales amount in previous 12 months                    & 0         & 379.85 & 560.04 & 9877.93  & -660.00 & 4.33     \\ \hline
			total christmas sales amount in previous 24 months                    & 0         & 430.10 & 611.41 & 10500.09 & -660.00 & 4.05     \\ \hline
			total christmas items in previous 12 months                           & 0         & 4.13   & 6.63   & 506.00   & 2.00    & 44.49    \\ \hline
			total christmas items in previous 24 months                           & 0         & 4.46   & 7.03   & 506.00   & 2.00    & 41.25    \\ \hline
			total amount of back to school sales in previous 12 months            & 0         & 396.37 & 619.21 & 10159.37 & -150.00 & 4.23     \\ \hline
			total amount of back to school sales in previous 24 months            & 0         & 416.13 & 623.76 & 10159.37 & -485.02 & 3.93     \\ \hline
			total amount of graduation sales in previous 12 months                & 0         & 376.62 & 564.66 & 13670.70 & -200.00 & 5.34     \\ \hline
			total amount of graduation sales in previous 24 months                & 0         & 405.84 & 585.14 & 13678.65 & -252.00 & 4.33     \\ \hline
			total spring sales amount in previous 12 months                       & 0         & 381.35 & 567.06 & 13190.30 & -308.90 & 4.63     \\ \hline
			total spring sales amount in previous 24 months                       & 0         & 421.00 & 612.91 & 13190.30 & -340.00 & 4.31     \\ \hline
			total summer sales amount in previous 12 months                       & 0         & 407.91 & 623.00 & 13205.62 & -300.00 & 4.51     \\ \hline
			total summer sales amount in previous 24 months                       & 0         & 439.29 & 643.85 & 17671.37 & -300.00 & 4.23     \\ \hline
			total autumn sales amount in previous 12 months                       & 0         & 401.56 & 616.92 & 10650.45 & -437.80 & 4.12     \\ \hline
			total autumn sales amount in previous 24 months                       & 0         & 447.43 & 663.93 & 11819.31 & -372.00 & 4.05     \\ \hline
			total winter sales amount in previous 24 months                       & 0         & 453.60 & 645.40 & 13921.45 & -189.99 & 4.02     \\ \hline
			total spring items in previous 12 months                              & 0         & 4.02   & 4.72   & 325.00   & 2.00    & 30.25    \\ \hline
			total spring items in previous 24 months                              & 0         & 4.35   & 5.95   & 400.00   & 2.00    & 28.17    \\ \hline
			total summer items in previous 12 months                              & 0         & 4.44   & 4.79   & 362.00   & 2.00    & 32.61    \\ \hline
			total summer items in previous 24 months                              & 0         & 4.69   & 7.63   & 504.00   & 2.00    & 38.59    \\ \hline
			total autumn items in previous 12 months                              & 0         & 4.20   & 7.47   & 501.00   & 2.00    & 41.44    \\ \hline
			total autumn items in previous 24 months                              & 0         & 4.46   & 7.51   & 501.00   & 2.00    & 42.06    \\ \hline
			total winter items in previous 12 months                              & 0         & 4.12   & 5.76   & 506.00   & 2.00    & 48.15    \\ \hline
			total winter items in previous 24 months                              & 0         & 4.54   & 6.97   & 506.00   & 2.00    & 38.14    \\ \hline
			total count of back to school items in previous 12 months             & 0         & 4.29   & 5.38   & 362.00   & 2.00    & 38.67    \\ \hline
			total count of back to school items in previous 24 months             & 0         & 4.41   & 7.12   & 502.00   & 2.00    & 36.16    \\ \hline
			total count of graduation items in previous 12 months                 & 0         & 4.15   & 3.83   & 190.00   & 2.00    & 14.61    \\ \hline
			total count of graduation items in previous 24 months                 & 0         & 4.41   & 6.34   & 504.00   & 2.00    & 37.55    \\ \hline
			total number of net instore esps in previous 12 months                & 0         & 2.60   & 1.17   & 17.00    & 2.00    & 3.74     \\ \hline
			total number of net instore esps in previous 24 months                & 0         & 2.81   & 1.45   & 35.00    & 2.00    & 4.48     \\ \hline
			total number of net instore esps lifetime                             & 0         & 3.14   & 1.98   & 43.00    & 2.00    & 4.09     \\ \hline
			avg term of all esps in previous 12 months                            & 0         & 25.32  & 14.48  & 120.00   & 0.48    & 0.89     \\ \hline
			total number of returned instore esps in previous 12 months           & 0         & 2.55   & 1.19   & 14.00    & 2.00    & 4.00     \\ \hline
			total number of returned instore esps in previous 24 months           & 0         & 2.61   & 1.38   & 23.00    & 2.00    & 5.36     \\ \hline
			total number of returned instore esps lifetime                        & 0         & 2.71   & 1.52   & 25.00    & 2.00    & 5.19     \\ \hline
			total items purchased during back to school gift guide 2002 promotion & 0         & 4.26   & 9.39   & 170.00   & -1.00   & 15.76    \\ \hline
			total items purchased during bond 2002 promotion                      & 0         & 3.97   & 9.15   & 307.00   & -2.00   & 29.86    \\ \hline
			total items purchased during expo 2001 promotion                      & 0         & 3.68   & 3.95   & 76.00    & -2.00   & 11.07    \\ \hline
			total items purchased during holiday gift guide 2001 promotion        & 0         & 2.94   & 1.97   & 20.00    & -2.00   & 3.43     \\ \hline
			total items purchased during holiday gift guide 2002 promotion        & 0         & 3.31   & 2.42   & 26.00    & -1.00   & 3.36     \\ \hline
			total items purchased during holiday mailer 2001 promotion            & 0         & 3.59   & 2.22   & 18.00    & -3.00   & 2.09     \\ \hline
			total items purchased during holiday mailer 2002 promotion            & 0         & 3.81   & 2.84   & 44.00    & -3.00   & 4.20     \\ \hline
			promo\_nov period:  total sales                                       & 0         & 333.94 & 674.07 & 11633.21 & -111.09 & 5.97     \\ \hline
			total \$ spent during bond 2002 promotion                             & 0         & 338.26 & 526.43 & 6821.51  & -134.00 & 4.29     \\ \hline
			total \$ spent during expo 2001 promotion                             & 0         & 366.73 & 570.37 & 6491.80  & -372.00 & 3.94     \\ \hline
			total \$ spent during holiday mailer 2002 promotion                   & 0         & 344.76 & 493.71 & 5639.97  & -165.00 & 3.85     \\ \hline
			total \$ spent during holiday mailer promotions                       & 0         & 366.64 & 498.45 & 5818.81  & -165.00 & 3.52     \\ \hline
			mailed in holiday 2001 mailer                                         & 1         & 0.18   & NA     & NA       & NA      & NA       \\ \hline
			mailed in holiday 2002 mailer                                         & 1         & 0.20   & NA     & NA       & NA      & NA       \\ \hline
			indicator of  holiday gift guide 2002 promotion response              & 1         & 0.01   & NA     & NA       & NA      & NA       \\ \hline
			indicator of back to school gift guide 2002 promotion response        & 1         & 0.01   & NA     & NA       & NA      & NA       \\ \hline
			indicator of bond 2002 promotion response                             & 1         & 0.01   & NA     & NA       & NA      & NA       \\ \hline
			indicator of expo 2001 promotion response                             & 1         & 0.01   & NA     & NA       & NA      & NA       \\ \hline
			indicator of holiday gift guide 2001 promotion response               & 1         & 0.00   & NA     & NA       & NA      & NA       \\ \hline
			indicator of holiday mailer 2001 promotion response                   & 1         & 0.01   & NA     & NA       & NA      & NA       \\ \hline
			indicator of holiday mailer 2002 promotion response                   & 1         & 0.01   & NA     & NA       & NA      & NA       \\ \hline
		\end{tabular}
	}
	\end{table}
	\newpage
	\bibliographystyle{chicago}
	\bibliography{Thompson1}
	\newpage